\documentclass[11pt, letterpaper]{article}

\usepackage[left=1in, right=1in, top=1in, bottom=1in]{geometry}
\usepackage[utf8]{inputenc}
\usepackage[T1]{fontenc}
\usepackage{bm}
\usepackage{type1cm}
\usepackage{lettrine}
\usepackage{amsmath,amssymb,amsthm}
\usepackage{moreverb}
\usepackage{mathtools}
\usepackage{amsmath}
\usepackage{amssymb}
\usepackage{algorithmic}
\usepackage{graphics}
\usepackage{graphicx}
\usepackage{subfigure}
\usepackage{caption}
\usepackage{extarrows}
\usepackage{color}
\usepackage{framed}
\usepackage{wrapfig}
\usepackage{bm}
\usepackage{mathrsfs}
\usepackage{mathabx}
\usepackage{multirow}
\usepackage{longtable}
\usepackage{hyperref}
\usepackage{paralist}
\usepackage{indentfirst}
\usepackage{relsize}
\usepackage{extarrows}
\usepackage{upgreek}
\usepackage{bm}
\usepackage[dvipsnames,table]{xcolor}
\usepackage{mwe}
\usepackage{booktabs}
\usepackage{authblk}
\usepackage{lettrine}
\usepackage{type1cm}
\usepackage{fancyhdr}
\usepackage{threeparttable}
\usepackage[sort&compress,numbers]{natbib}
\usepackage[figurename=Figure]{caption}
\usepackage[figuresright]{rotating}
\usepackage{listings}
\usepackage{makecell}
\usepackage[dvipsnames]{xcolor}
\usepackage{algorithm}
\usepackage{lmodern}
\usepackage{amsmath}
\usepackage{amssymb}
\usepackage{mathtools}
\usepackage{amsthm}
\usepackage{array}
\usepackage{makecell}

\usepackage[most]{tcolorbox}
\usepackage{amssymb}  

\usepackage{microtype}
\usepackage{url}
\usepackage{hyperref}
\usepackage{graphicx}
\usepackage{subfigure}
\usepackage{booktabs} 
\usepackage{makecell}
\usepackage{tcolorbox}
\usepackage{xcolor}
\usepackage{graphicx} 

\usepackage{graphicx}  
\usepackage{booktabs}  
\usepackage{multirow}  
\usepackage{makecell}  
\usepackage{ragged2e} 

\tcbuselibrary{skins} 

\usepackage[table]{xcolor}

\usepackage{upgreek}
\usepackage{bm}
\usepackage{mwe}
\usepackage[dvipsnames,table]{xcolor}
\usepackage{booktabs}
\usepackage{authblk}
\usepackage{lettrine}
\usepackage{type1cm}
\usepackage{threeparttable}
\usepackage[sort&compress,numbers]{natbib}
\usepackage[figurename=Fig.]{caption}
\usepackage[normalem]{ulem}

\usepackage{graphicx}
\usepackage{svg}

\usepackage{multirow}
\usepackage{wrapfig}
\usepackage{array} 
\usepackage{subfigure}
\usepackage{enumerate}
\usepackage{scalerel} 
\usepackage{nameref}

\usepackage{svg}

\newcolumntype{H}{>{\setbox0=\hbox\bgroup}c<{\egroup}@{}}
\definecolor{skyblue}{RGB}{135, 206, 235}

\graphicspath{ {./Fig./} }
\usepackage[font=footnotesize,labelfont=bf]{caption}

\newcommand{\hsedit}[1]{{\color{black} #1}}

\newcommand{\kredit}[1]{{\color{black} #1}}

\hypersetup{
bookmarksopen=true,
bookmarksnumbered=true,
unicode=false,
pdftoolbar=true,
pdfmenubar=true,
pdffitwindow=false,
pdfstartview={FitH},
pdftitle={My title},
pdfauthor={Author},
pdfsubject={Subject},
pdfcreator={Creator},
pdfproducer={Producer},
pdfkeywords={keywords},
pdfnewwindow=true,
colorlinks=true,
linkcolor=blue,
citecolor=blue,
filecolor=blue,
urlcolor=blue
}

\begin{document}

\title{\textbf{Evaluating LLMs' Divergent Thinking \kredit{Capabilities} for Scientific Idea Generation with Minimal Context}}

\author[1]{Kai Ruan}
\author[2]{Xuan Wang}
\author[1]{Jixiang Hong}
\author[3]{Peng Wang}
\author[4,5]{Yang Liu}
\author[1,$^*$]{Hao Sun}

\affil[1]{\small Gaoling School of Artificial Intelligence, Renmin University of China, Beijing, China}
\affil[2]{\small College of Computer Science and Technology, Zhejiang University, Hangzhou, China}
\affil[3]{\small Bank of China, Beijing, China}
\affil[4]{\small School of Engineering Science, University of Chinese Academy of Sciences, Beijing, China}
\affil[5]{\small State Key Laboratory of Nonlinear Mechanics, Institute of Mechanics, Chinese Academy of Sciences, Beijing, China \vspace{18pt}}

\affil[*]{\small Corresponding author (haosun@ruc.edu.cn)\vspace{12pt}}

\date{}

\maketitle

\normalsize

\vspace{-28pt} 
\begin{abstract}
\small
Large Language Models (LLMs) demonstrate remarkable capabilities in scientific tasks such as literature analysis and experimental design. \kredit{For instance, these models excel at} accurately extracting key findings from papers or generating coherent experimental procedures. \kredit{However}, existing evaluation benchmarks primarily assess performance using rich contextual inputs. We introduce {LiveIdeaBench}, a comprehensive benchmark evaluating LLMs' scientific idea generation by assessing divergent thinking capabilities using single-keyword prompts. Drawing from Guilford's creativity theory, our benchmark employs a dynamic panel of state-of-the-art LLMs to assess generated ideas across five dimensions: originality, feasibility, fluency, flexibility, and clarity. Through experimentation with over 40 leading models across 1,180 keywords spanning 22 scientific domains, we reveal that the scientific idea generation capabilities measured by our benchmark, are poorly predicted by standard metrics of general intelligence scores.
Our results demonstrate that models like {QwQ-32B-preview} achieve creative performance comparable to models such as {claude-3.7-sonnet:thinking}, despite significant gaps in their general intelligence scores. These findings highlight the need for specialized evaluation benchmarks for scientific idea generation and suggest that enhancing these idea generation capabilities in LLMs may require different training strategies than those used for improving general problem-solving abilities. Such strategies could potentially enable a wider range of AI tools tailored for different stages of the scientific process.
\end{abstract}

\begin{figure*}[b!]
    \centering
    \includegraphics[width=1.0\textwidth]{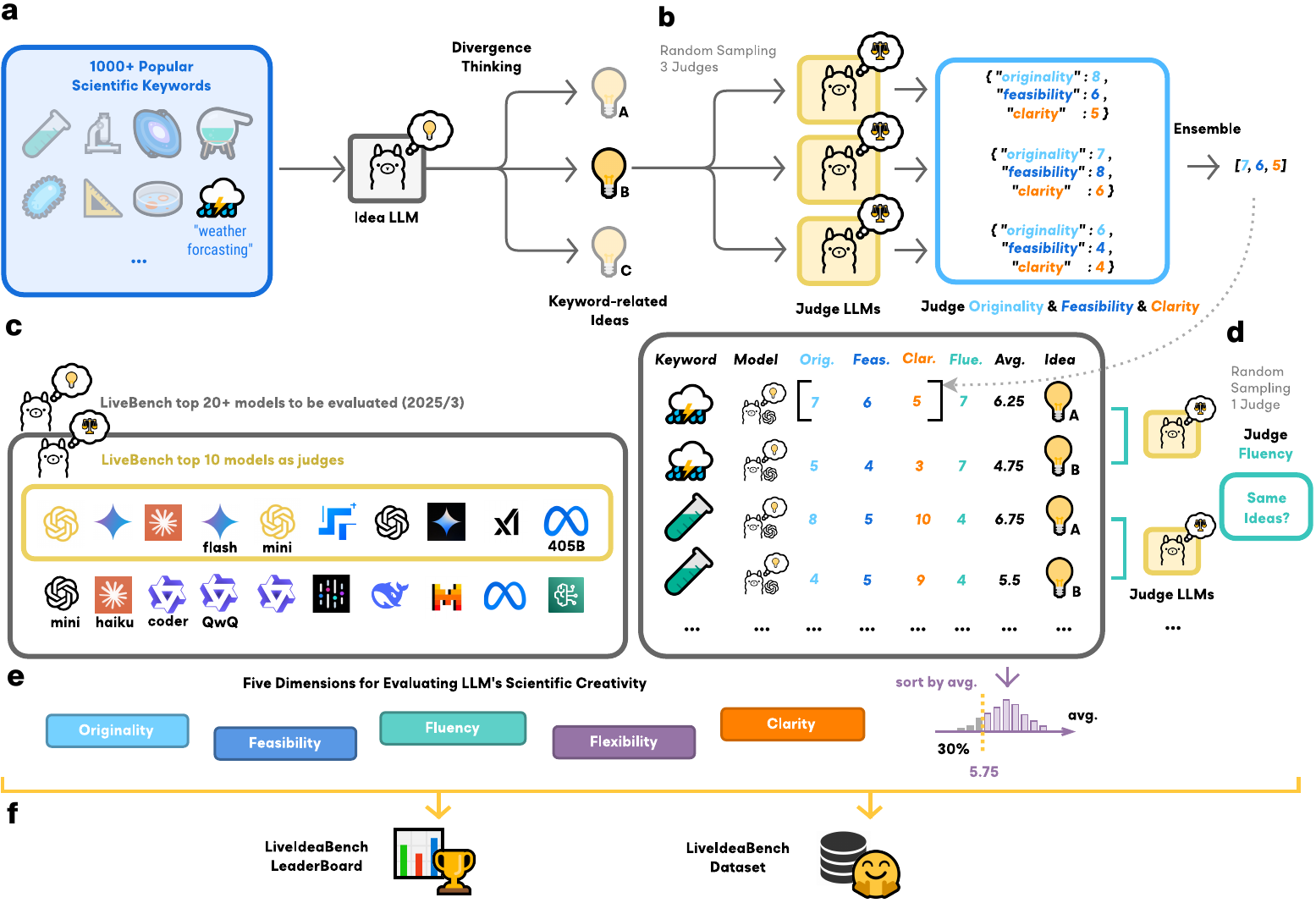}
    \caption{ \textbf{Overall design of the {LiveIdeaBench} benchmark.} 
    \textbf{a.} Over 1,000 scientific keywords, representing diverse domains, are used in prompts for the Idea LLMs, encouraging divergent thinking and the generation of novel scientific ideas.  
    \textbf{b.} Sampled Judge LLMs evaluate the generated ideas across three primary dimensions: originality, feasibility and clarity, assigning numerical scores to each idea. 
    \textbf{c.} The evaluation panel comprises the top 10 state-of-the-art models selected from LiveBench, ensuring robust assessment through sampling and ensemble scoring. 
    \textbf{d.} Fluency scores are derived by analyzing the diversity and substantive differences among ideas generated from the same keyword (using a randomly sampled judge), while originality, feasibility, and clarity metrics are combined for integrated evaluation. 
    \textbf{e.} Following Guilford's creativity theory, the evaluation methodology assesses five critical dimensions: originality, feasibility, clarity, fluency, and flexibility, with flexibility computed as the 30th percentile of the averaged scores across the other four dimensions. 
    \textbf{f.} The {LiveIdeaBench} benchmark provides a comprehensive dataset of generated ideas, evaluation metrics, and a dynamic leaderboard tracking the performance of over 40 models.}
\label{fig:overall}
\end{figure*}

\vspace{12pt} 

\section*{Introduction}
\addcontentsline{toc}{section}{Introduction} 
\label{sec:introduction}

The advancement of scientific knowledge relies heavily on creative thinking and the generation of novel hypotheses. The ability to envision new possibilities and formulate testable explanations is crucial for scientific progress. In recent years, LLMs have demonstrated remarkable capabilities in various scientific tasks, from literature analysis to experimental design, suggesting their potential as powerful tools for augmenting scientific discovery \cite{reviewer5_Nature_wang2023scientific, reviewer5_NC_shi2023surprising, reviewer5_arxiv_ai4science2023impact}. Concurrently, machine learning techniques are being applied to forecast future research directions by analyzing the structure and evolution of scientific knowledge networks \cite{reviewer5_NMI_krenn2023forecasting}. As Rafner \textit{et al.} \cite{reviewer3_reviewer5_NHB_rafner2023creativity_Rafner_Janet_Beaty_Roger} note, these generative AI systems now perform comparably to humans on some creativity tests and could potentially enhance the creative capabilities of knowledge workers worldwide.

\kredit{To understand the potential of AI in this domain,} early theoretical work on human creativity offers valuable insights for our investigation. In their seminal 1962 work, Getzels and Jackson \cite{getzels1962creativity} examined the relationship between intelligence and creativity, particularly among ``gifted'' students. Their study yielded two crucial findings: first, high IQ does not necessarily equate to high creativity; and second, creativity and intelligence function as relatively independent traits. These observations led to the threshold theory \cite{getzels1962creativity, guilford1967nature}, proposing that intelligence is necessary but not sufficient for creativity, with its influence diminishing beyond a certain threshold. 

However, subsequent research has yielded mixed evidence for this theory. While Gralewski \textit{et al.} \cite{reviewer5_NER_gralewski2012intelligence} and Jauk \textit{et al.} \cite{JAUK2013212_Mathias_Benedek} found some empirical support under certain conditions, some studies found no support \cite{reviewer5_JSGE_kim2005can, reviewer5_PID_preckel2006relationship}, challenging the threshold concept. More recent work by Weiss \textit{et al.} \cite{reviewer5_JI_weiss2020reappraisal} concluded that the relationship is likely linear with no evidence supporting a threshold. These varying findings suggest that the development of idea generation capabilities might be underpinned by different cognitive mechanisms than general intelligence, a pattern potentially relevant to LLM development as well. It's also noteworthy that the threshold debate has nuances regarding potential versus achievement; some research suggests that while a threshold might (or might not) apply to divergent thinking ability, intelligence may remain related \cite{JAUK2013212_Mathias_Benedek, reviewer5_PS_park2008ability} or even become more important for translating potential into real-world creative achievements across the entire range \cite{reviewer5_CCAQ_runco1986threshold, reviewer5_LID_benedek2024relationship_Mathias_Benedek}. These early studies primarily focused on the relationship between general cognitive abilities and creative potential, often measured via divergent thinking tasks. While much of this debate centers on general creativity, the specific relationship between intelligence and scientific creativity, which heavily relies on deep domain knowledge and rigorous methodology alongside ideation, remains an area requiring nuanced investigation, though the general finding of creativity and intelligence being distinct capabilities provides a useful starting point for examining LLMs.

Another theoretical strand addresses creativity assessment methodologies. Guilford's influential theoretical framework \cite{guilford1950creativity, guilford1967nature} introduced the distinction between divergent thinking (generating varied responses to open-ended prompts) and convergent thinking (finding optimal solutions to well-defined problems). Guilford identified four key aspects of divergent thinking: the ability to generate a large number of ideas (fluency), the capacity to think across different categories (flexibility), the generation of novel ideas (originality), and the development of detailed and refined ideas (elaboration). More recent research by Cortes \textit{et al.} \cite{reviewer3_COBS_CORTES201990_Adam_Green} has noted that traditional creativity assessments typically involve elements of both divergent and convergent thinking rather than isolating either process. Their findings indicate that purportedly divergent and convergent tasks show limited correlation, questioning whether they reflect different components of the same higher-level construct (creativity). This work highlights the complexity of measuring creative processes and informs our approach to evaluating idea generation in LLMs.

Broader theoretical frameworks further enrich this picture. For instance, Margaret Boden \cite{reviewer2_boden2004creative} distinguishes between psychological creativity (P-creativity, novel to the individual) and historical creativity (H-creativity, novel to humanity), while Teresa Amabile's \cite{reviewer2_amabile1983social} componential theory highlights the interplay of domain-relevant skills, creativity-relevant processes (including divergent thinking), and intrinsic task motivation. Melvin Rhodes' \cite{reviewer2_rhodes1961analysis} ``4 P's'' (Person, Process, Press, Product) provide a holistic view, reminding us that creativity involves the individual creator, the cognitive processes, the environmental influences, and the resulting outcome. While \kredit{our approach} primarily assesses aspects of the creative ``process'' (specifically idea generation fluency and diversity under minimal constraints), understanding these broader theories helps contextualize its scope and limitations within the larger landscape of scientific creativity, which ultimately demands valuable and impactful products.

\begin{figure*}[b!]
    \centering
    \includegraphics[width=1.0\textwidth]{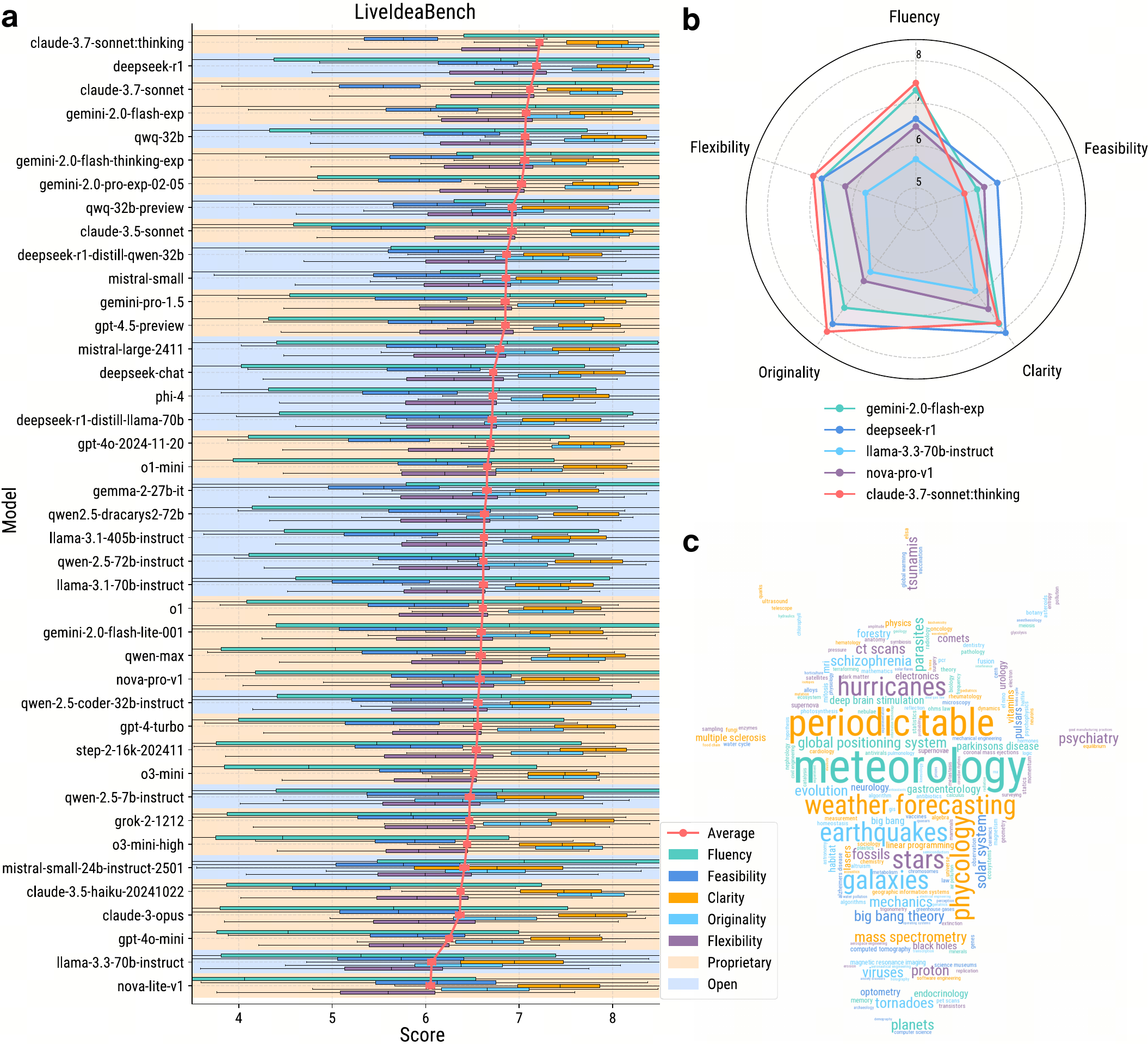} 
    \caption{\textbf{Performance comparison of models evaluated on {LiveIdeaBench}}. 
    \textbf{a.} Dimensional scores (originality, feasibility, fluency, clarity, and flexibility) and overall performance (red line, with 95\% confidence intervals, $n=6987$) for open-weight and proprietary models. \kredit{The central line of box represents the median, the box spans the 25th to 75th percentiles, the whiskers extend to the minimum and maximum values within 1.5 times the interquartile range.}
    \textbf{b.} Multidimensional performance profiles of representative models across the five evaluation dimensions.
    \textbf{c.} Word cloud visualization of scientific keywords. For detailed scores and 95\% CIs for each model, see \textcolor{blue}{Supplementary Table S.3}. }
    \label{fig:liveideabench_barplot}
\end{figure*}

\kredit{These theoretical frameworks from human creativity research have informed modern approaches to evaluating LLMs' idea generation capabilities.} Contemporary creativity research has expanded assessment methodologies beyond traditional paper-based tests, with Rafner \textit{et al.} \cite{rafner2023towards} developing CREA, a comprehensive game-based tool for measuring divergent and convergent thinking. Their validation studies across diverse populations have demonstrated the effectiveness of these novel assessment approaches. Benedek's \cite{reviewer5_LID_benedek2024relationship_Mathias_Benedek} recent work clarifies that creative potential (measured through divergent thinking tasks) explains only limited variance in creative achievement, emphasizing complex interactions between potential, behavior, and environmental factors.

Recent research has systematically explored LLMs' potential for tasks analogous to human creative processes through various methodological approaches. Rafner \textit{et al.} \cite{reviewer3_reviewer5_NHB_rafner2023creativity_Rafner_Janet_Beaty_Roger} provided a critical analysis of creativity in the age of generative AI, noting that while these systems now perform comparably to humans on some creativity tests, important qualitative differences remain. Their analysis emphasized the importance of integrating psychological science with computational approaches to develop more effective creativity support tools, highlighting both the promising capabilities and important limitations of current systems. Empirical assessments in creative contexts have shown promising results, exploring LLMs' potential through various applications. For instance, Meincke \textit{et al.} \cite{meincke2024using} focused on product ideation, comparing the quality and novelty of ideas generated by GPT-4 against those generated by humans. Comprehensive human evaluation studies \cite{si2024llmsgeneratenovelresearch} have compared LLM-generated research proposals with those from NLP researchers, revealing statistically significant advantages in novelty while maintaining comparable feasibility metrics. Complementing these approaches, Lee \textit{et al.} \cite{reviewer5_NHB_lee2024empirical} conducted an empirical study investigating the impact of using ChatGPT on human creative performance, finding that it significantly enhanced the articulation and creativity of human-generated ideas.

Studies directly examining divergent thinking abilities in LLMs have employed various methodological approaches. Cropley \cite{reviewer3_LL_cropley2023artificial} analyzed ChatGPT on the Divergent Association Task and found that while GPT-4 demonstrated strong semantic distance capabilities in generating unrelated words, its inconsistency and predictability highlighted important differences from human creative processes. This research emphasized the distinction between divergent thinking abilities and true creativity, suggesting the importance of understanding both the capabilities and limitations of these systems. Marrone, Cropley, and Medeiros \cite{reviewer3_CRJ_marrone2024does} further examined how narrow AI impacts human creativity, identifying both supportive functions and important limitations in creative processes. However, Wenger and Kenett \cite{reviewer3_arxiv_wenger2025we_Yoed_Kenett} found that while LLMs performed on par with humans in individual creativity tests, they exhibited significantly lower response diversity at the group level, showing high homogeneity in creative outputs even across different model families. This creative homogeneity represents an important limitation in current systems. Confirming this potential downside, Doshi \textit{et al.} \cite{ScienceAdvances_Doshi2023GenerativeAE}, in an experimental study on story writing, found that while access to generative AI ideas caused stories to be evaluated as more creative individually (especially for less creative writers), these AI-enabled stories were significantly more similar to each other than stories produced by humans alone, pointing to a potential reduction in collective novelty. These findings collectively underscore the importance of evaluating not just the quality but also the diversity and novelty of AI-generated creative content.

The assessment methodology for LLM creativity has evolved through several sophisticated frameworks designed for broad evaluation. For instance, Lu \textit{et al.} \cite{lu2024benchmarkinglanguagemodelcreativity} evaluated text-to-code creativity using programming challenges combined with prompting strategies designed to force novel solutions by restricting previously used techniques, though potential reliance on static problem sets raises limitations. Building upon foundational creativity theories \cite{guilford1967nature}, Zhao \textit{et al.} \cite{zhao2024assessing} proposed a comprehensive, albeit static, framework adapting established psychological test tasks to assess \textit{general} creativity across multiple dimensions. These frameworks offer valuable, comprehensive assessments for their respective areas, often incorporating both divergent and convergent aspects or multiple task types with richer contextual inputs.

\kredit{In the context of reasoning mechanics,} Chain-of-thought (CoT) prompting has emerged as a powerful technique for enhancing LLMs' reasoning capabilities. Following the initial introduction of CoT prompting \cite{wei2022chain}, several advanced variants have been developed \cite{10.5555/3600270.3601883,10.5555/3666122.3666639}, enabling models to explore multiple reasoning paths simultaneously. Recent investigations into programming applications \cite{li2023think} have demonstrated that incorporating brainstorming significantly improves LLMs' performance. A common thread among these approaches is their implicit reliance on \textit{divergent thinking} - the ability to generate multiple distinct solutions or paths from a single starting point. 

\kredit{Furthermore,} recent years have seen significant advances in automated scientific systems aimed at aiding or accelerating discovery, each offering unique approaches. The AI Scientist framework \cite{lu2024aiscientistfullyautomated} demonstrated potential for end-to-end research automation, while Nova \cite{hu2024novaiterativeplanningsearch} introduced iterative planning mechanisms for idea development. Systems like ResearchAgent \cite{baek2024researchagentiterativeresearchidea} and Scideator \cite{radensky2024scideatorhumanllmscientificidea} have further refined these approaches through knowledge graph integration. Pushing the boundaries further, the recent AI co-scientist framework from Google \cite{gottweis2025aicoscientist} showcases a sophisticated multi-agent system designed to generate novel research hypotheses and detailed research proposals. The system employs specialized agents for generation, reflection, ranking, evolution, proximity analysis, and meta-review to iteratively refine scientific hypotheses through a self-improving cycle, with generated hypotheses later validated through real-world laboratory experiments conducted by human researchers. Similarly, Wang and colleagues' \cite{reviewer5_ACL_wang2024scimon} SCIMON system leverages literature retrieval and iterative novelty enhancement to generate scientific ideas, while Gu and Krenn \cite{reviewer5_arxiv_gu2024interesting, reviewer5_arxiv_gu2024forecasting} demonstrated approaches using knowledge graphs and LLMs to suggest research directions. However, these specialized systems typically rely heavily on extensive knowledge bases, complex contextual inputs, or planning processes to generate and refine ideas or research plans. While highly effective for their intended applications, this dependence on rich context and complex architectures makes it challenging to isolate and evaluate their fundamental capabilities for \textit{initial} idea generation from minimal cues, particularly their capacity for divergent thinking. This limitation is particularly evident in systems like SciPIP \cite{wang2024scipipllmbasedscientificpaper} and IdeaSynth \cite{pu2024ideasynthiterativeresearchidea}, whose structure encourages integrating existing information rather than generating diverse possibilities from minimal input.

The challenges in evaluating creative output at scale necessitate automated assessment methods. The emergence of LLMs as evaluation tools has opened new possibilities for assessing model outputs at scale. Various evaluation approaches \cite{dubois2024lengthcontrolledalpacaevalsimpleway, MT-Bench, li2024crowdsourced, reviewer5_ACL_chiang-lee-2023-large, reviewer5_ACL_chen-etal-2023-exploring-use, reviewer5_LREC_huang-etal-2024-chatgpt} have demonstrated the feasibility of using LLMs to evaluate other models' responses, offering advantages in terms of efficiency and cost-effectiveness compared to human evaluation \cite{reviewer5_NAC_gilardi2023chatgpt}. Recent advances, such as the jury-based framework \cite{verga2024replacingjudgesjuriesevaluating} and reference-guided verdict method \cite{badshah2024referenceguidedverdictllmsasjudgesautomatic}, have shown promising results in reducing individual model biases and achieving high agreement with human judgments. The development of specialized evaluation models like Prometheus 2 \cite{reviewer5_arxiv_kim2024prometheus} further underscores the growing sophistication in this area. \kredit{However, most existing approaches} focus on evaluating responses against predetermined criteria or reference answers, making them better suited for convergent thinking tasks than divergent thinking assessment. Second, the evaluation of scientific creativity presents unique challenges that go beyond traditional metrics, requiring simultaneous assessment of originality, feasibility, clarity, fluency, and flexibility.

The landscape of artificial intelligence has undergone dramatic transformation since the emergence of ChatGPT in late 2022, with LLMs demonstrating increasingly sophisticated capabilities in scientific contexts. As we approach late 2024, these models \kredit{have surpassed human expert baselines on various standardized benchmarks} across multiple dimensions, \kredit{exemplified by AlphaGeometry} \cite{AlphaGeometry}. Yet, these advances in general intelligence prompt a fundamental question: does the potential for scientific idea generation in LLMs grow in tandem with their analytical capabilities? This question becomes particularly pertinent when evaluating LLMs' potential contributions to scientific innovation. Current evaluation benchmarks predominantly rely on rich contextual inputs, such as research paper titles, abstracts, or complete articles. While these approaches effectively assess models' ability to comprehend and synthesize existing knowledge, they are not designed to systematically evaluate a crucial aspect of scientific thinking - the capacity to generate novel ideas from limited information. This limitation is particularly significant, as many groundbreaking scientific discoveries originate from unexpected connections and conceptual leaps \kredit{\cite{reviewer5_NC_shi2023surprising}}. 


\begin{figure*}[b!]
\centering
\includegraphics[width=0.99\textwidth]{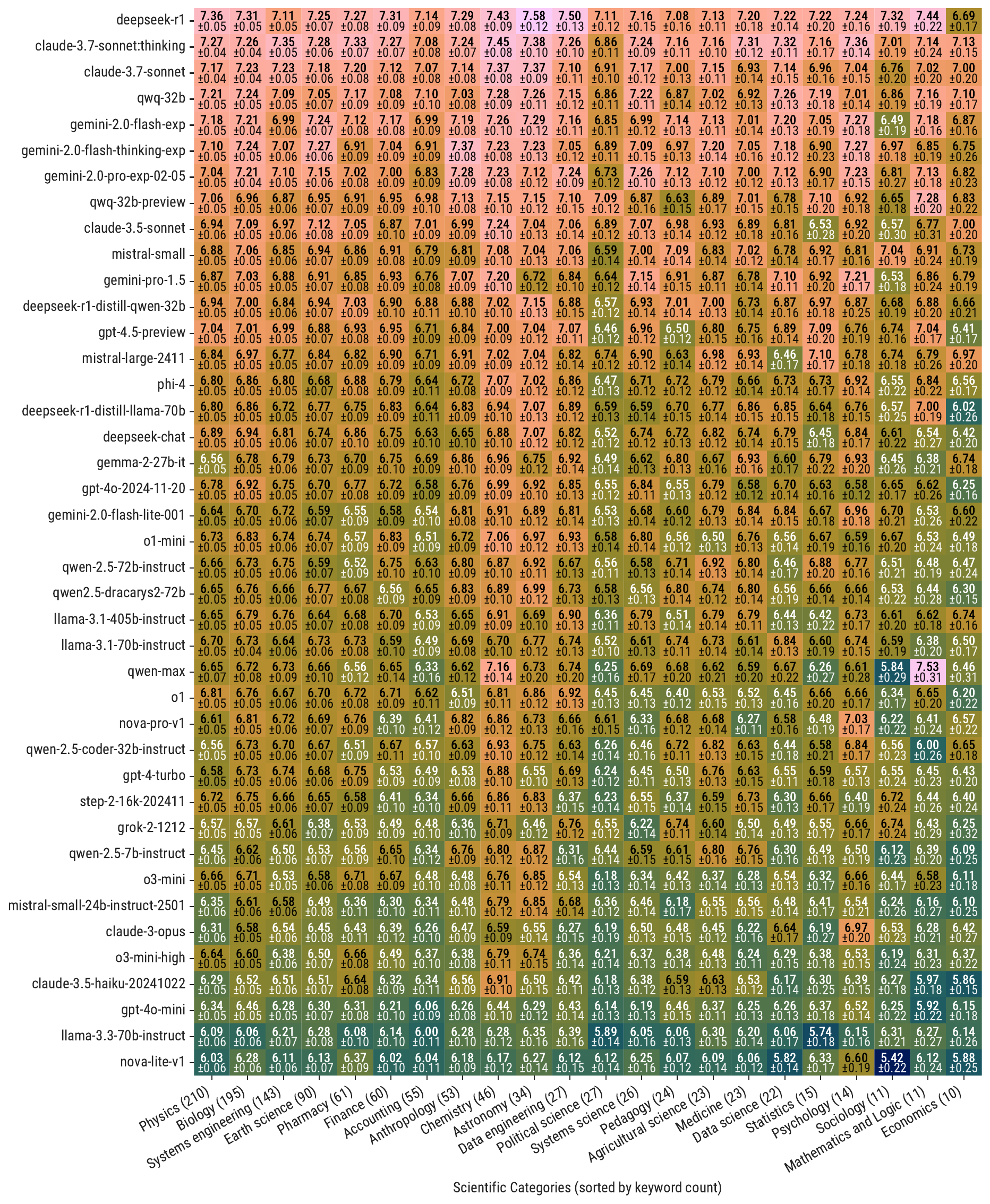}
\caption{\textbf{Model Performance on {LiveIdeaBench} Across Scientific Categories.} The heatmap displays average performance scores with 95\% confidence intervals for model-discipline combinations. Scientific categories were classified using SciBERT \cite{beltagy-etal-2019-scibert} through semantic similarity computation following the framework from \cite{cohen2021boundary}. Higher scores (pink) indicate better idea generation ability within each discipline. Numbers in parentheses following each scientific category indicate the keyword count associated with that discipline. Categories are sorted by keyword count.}
\label{fig:heat_model_sci_cat}
\end{figure*}

\kredit{In this work, we introduce} \texttt{LiveIdeaBench} (Fig. \ref{fig:overall}), a novel evaluation benchmark designed to assess LLMs' capabilities in divergent thinking for scientific idea generation under constrained conditions. Unlike existing benchmarks that predominantly evaluate convergent thinking by requiring models to derive single correct solutions from rich contextual information, \texttt{LiveIdeaBench} is explicitly grounded in Guilford's seminal theory of divergent production \cite{guilford1967nature}. Our methodology operationalizes key principles from this framework: we use single-keyword prompts as minimal context. This constraint encourages models to generate connections and concepts primarily from their internal knowledge representations, rather than synthesizing readily available information from detailed prompts (e.g., abstracts, or complete articles), thus probing their capacity for less scaffolded, more internally-driven ideation relevant to initial brainstorming. This approach is analogous to the broad stimuli used in classic divergent thinking tasks (e.g., the ``Utility Test'' described by Guilford), to elicit the generation of multiple and varied potential ideas, fostering a ``ready flow of ideas'' rather than convergence towards a predefined outcome. This approach necessitates that models ``produce their own answers'', a hallmark of divergent production assessment distinct from selection-based or highly constrained convergent thinking tasks. 

\hsedit{By employing a specialized multi-model jury system}, our evaluation directly measures core divergent production dimensions like fluency (quantity and diversity of ideas, reflecting the capacity to generate a ``number of varied responses'') and originality (novelty of ideas), as conceptualized by Guilford, \hsedit{across 41 state-of-the-art models using over 1,000 scientific keywords from 22 distinct domains}. While acknowledging that comprehensive scientific creativity involves an interplay of both divergent and convergent processes, \texttt{LiveIdeaBench} focuses on and evaluates the foundational, generative phase: the ability to produce diverse scientific concepts from sparse cues. This capacity is critical for sparking innovation but is often overlooked by conventional benchmarks. Thus, our focus is on assessing this specific aspect of creative potential relevant to scientific discovery, rather than evaluating the entire scientific process, which also involves deep domain expertise and subsequent validation. \hsedit{We demonstrate that general intelligence metrics do not necessarily correlate with capacity for scientific ideation; notably, models such as QwQ-32B-preview achieve performance comparable to established systems like claude-3.7-sonnet:thinking despite significant gaps in general benchmarks. We posit that such specialized evaluation serves as a crucial initial step towards developing effective human-AI hybrid intelligence systems \cite{reviewer3_Bonus_1st_author_rafner2022mapping, reviewer5_Computer_akata2020research, reviewer5_BISE_dellermann2019hybrid}, providing the necessary insights to leverage LLMs as collaborative tools for scientific discovery.}

\section*{Results}
\addcontentsline{toc}{section}{Results}

We assessed the performance of over 40 language models across multiple dimensions of scientific ideation using our comprehensive evaluation benchmark. A detailed description of the generated idea dataset is provided in \textcolor{blue}{Supplementary Note 2}, and its constituent fields are listed in \textcolor{blue}{Supplementary Table S.1}. The evaluation results are visualized in Fig. \ref{fig:liveideabench_barplot}. For detailed outcomes, including raw scores and examples, please refer to \textcolor{blue}{Supplementary Table S.2}; aggregated performance metrics and model rankings are presented in \textcolor{blue}{Supplementary Table S.3}.

\paragraph{Diverse Model Performance Across Scientific Domains}
Our quantitative assessment and comparative analysis reveal distinct variations in model capabilities across scientific disciplines. (Fig. \ref{fig:heat_model_sci_cat}) While {gemini-2.0-flash-thinking-exp}, {deepseek-r1}, and {claude-3.7-sonnet:thinking} demonstrate high overall scores, the pattern of their relative strengths exhibits domain specificity. Importantly, although larger and more recent architectures generally achieve superior results, their advantage is not uniform across disciplines, suggesting that domain expertise and reasoning capabilities are not solely determined by model scale. For example, {gemini-2.0-flash-thinking-exp} excelled in anthropology ideation; {claude-3.7-sonnet:thinking} performed best in chemistry, medicine, and data science; while {deepseek-r1} demonstrated stronger relative performance in physics.

\begin{figure*}[b!]
\centering
\includegraphics[width=1.0\textwidth]{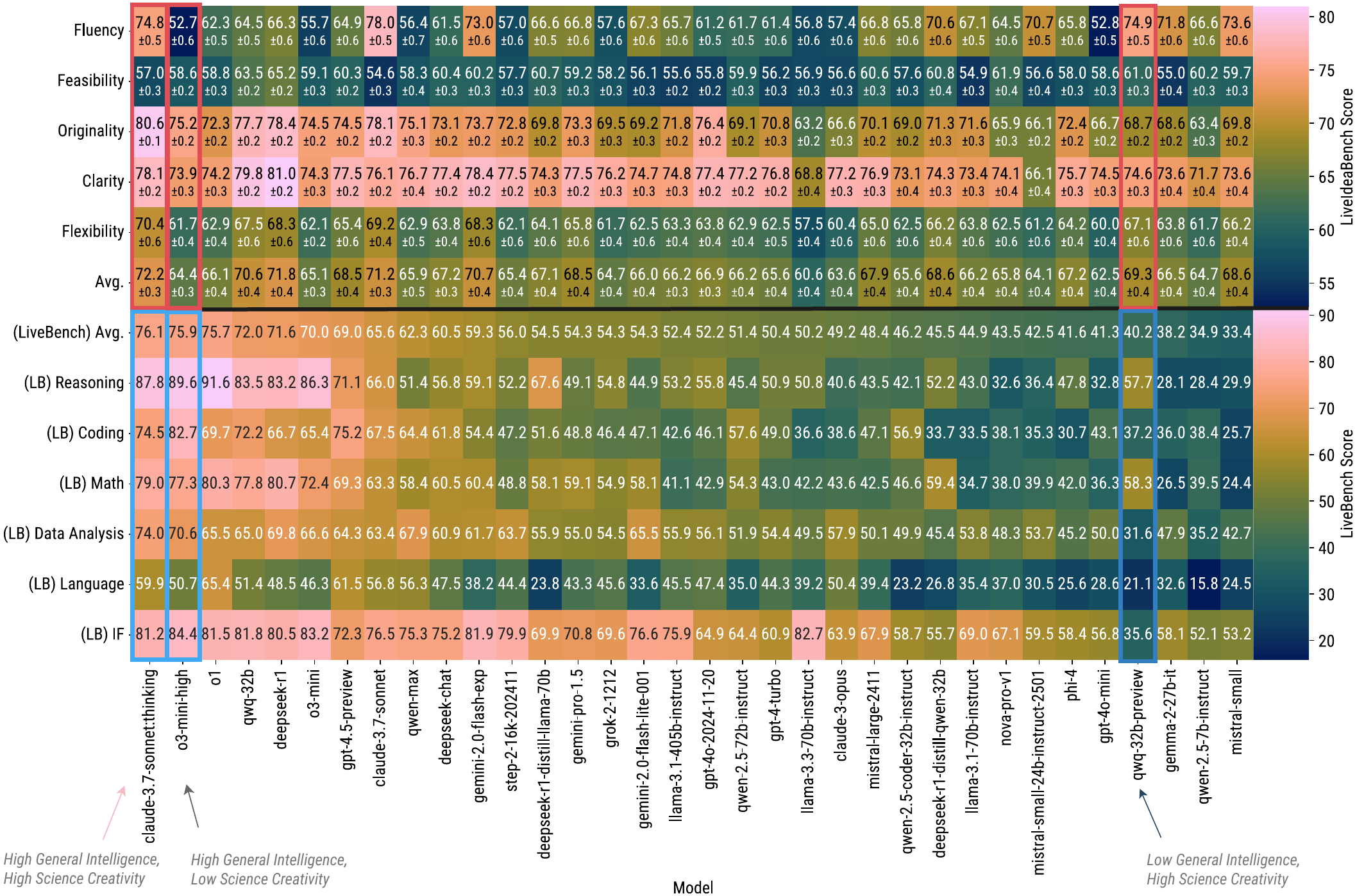}

\caption{\textbf{Comparison of {LiveIdeaBench} (upper half, assessing scientific ideation) and {LiveBench} (lower half, assessing general intelligence, denoted as {LB}) across various evaluation metrics for the same set of models.} The middle rows show contrasting trends between models in average performance. While {claude-3.7-sonnet:thinking} achieves the highest average score on {LiveBench}, its performance on scientific ideation tasks ({LiveIdeaBench}) is comparable to {qwq-32b-preview}, which ranks fourth from last on general intelligence metrics. The arrows highlight three representative patterns: {claude-3.7-sonnet:thinking} (left) exemplifies high general intelligence combined with high scientific ideation capability; {o3-mini-high} (middle) shows high general intelligence but low scientific ideation capability; and {qwq-32b-preview} (right) demonstrates low general intelligence but high scientific ideation capability. These contrasting patterns highlight that LLMs' scientific idea generation capability, as measured by {LiveIdeaBench}, is distinct from their general intelligence capabilities (e.g., reasoning, coding, and math), underscoring the necessity of {LiveIdeaBench} for evaluating scientific ideation potential.}

\label{fig:comp_livebench}
\end{figure*}

\paragraph{Distinction between General Intelligence and Scientific Idea Generation} The comparison between {LiveIdeaBench} and {LiveBench} metrics (see Fig. \ref{fig:comp_livebench} and \textcolor{blue}{Supplementary Fig. S.4}) uncovers a notable disconnect between general intelligence and scientific idea generation capabilities, revealing contrasting results patterns across models. While {claude-3.7-sonnet:thinking} leads in general intelligence ({LiveBench}) and also shows high effectiveness on scientific ideation ({LiveIdeaBench}), its idea generation capabilities are matched by models like {qwq-32b-preview} (ranked 8/41 in {LiveIdeaBench}). Notably, {qwq-32b-preview} exhibits low general intelligence scores, representing a profile of low general intelligence but high scientific ideation capability. As Fig. \ref{fig:comp_livebench} illustrates, other patterns exist, such as high general intelligence paired with lower scientific ideation capability (e.g., {o3-mini-high}). These contrasting profiles underscore that scientific idea generation capability, as measured by LiveIdeaBench, is poorly predicted by general intelligence, necessitating specialized benchmarks like {LiveIdeaBench} for evaluating this aspect of LLM potential. This statistically significant but weak positive correlation (\kredit{$r(41) = 0.357, p = 0.038, r^2 = 0.127, \ 95\% \ CI = [0.022, 0.621]$}; see \textcolor{blue}{Supplementary Fig. S.4}) suggests that fostering scientific idea generation capabilities in these systems may benefit from distinct development approaches compared to enhancing general problem-solving skills.

\paragraph{\kredit{Impact of Model Architecture and Training Strategies}}
We observe that {mistral-small}, which ranks near the bottom in general intelligence, demonstrates remarkably high scientific divergent thinking capabilities. Even more interestingly, the earlier released mistral-small model scored significantly higher in creativity than the newer same-size, same-family model {mistral-small-24b-instruct-2501}. We also observe that models from the same family (\kredit{for example,} {qwq-32b} and {qwq-32b-preview}) score similarly in creativity while differing dramatically in general intelligence. These two pairs of examples strongly suggest that a model's divergent thinking and convergent thinking abilities cannot reliably predict each other. This pattern holds true even when comparing models from the same family with different parameter counts. Focusing on {mistral-large-2411} and {mistral-small} (Fig. \ref{fig:comp_livebench}), we see that {mistral-small} still slightly outperforms {mistral-large-2411} in scientific divergent thinking, despite the latter having 123 billion parameters, far exceeding the former's 24 billion. This further suggests that parameter count alone does not appear to be the primary determinant of scientific divergence capabilities, suggesting other factors like training data or architectural nuances play significant roles.

\begin{figure*}[b!]
    \centering
    \includegraphics[width=1.0\textwidth]{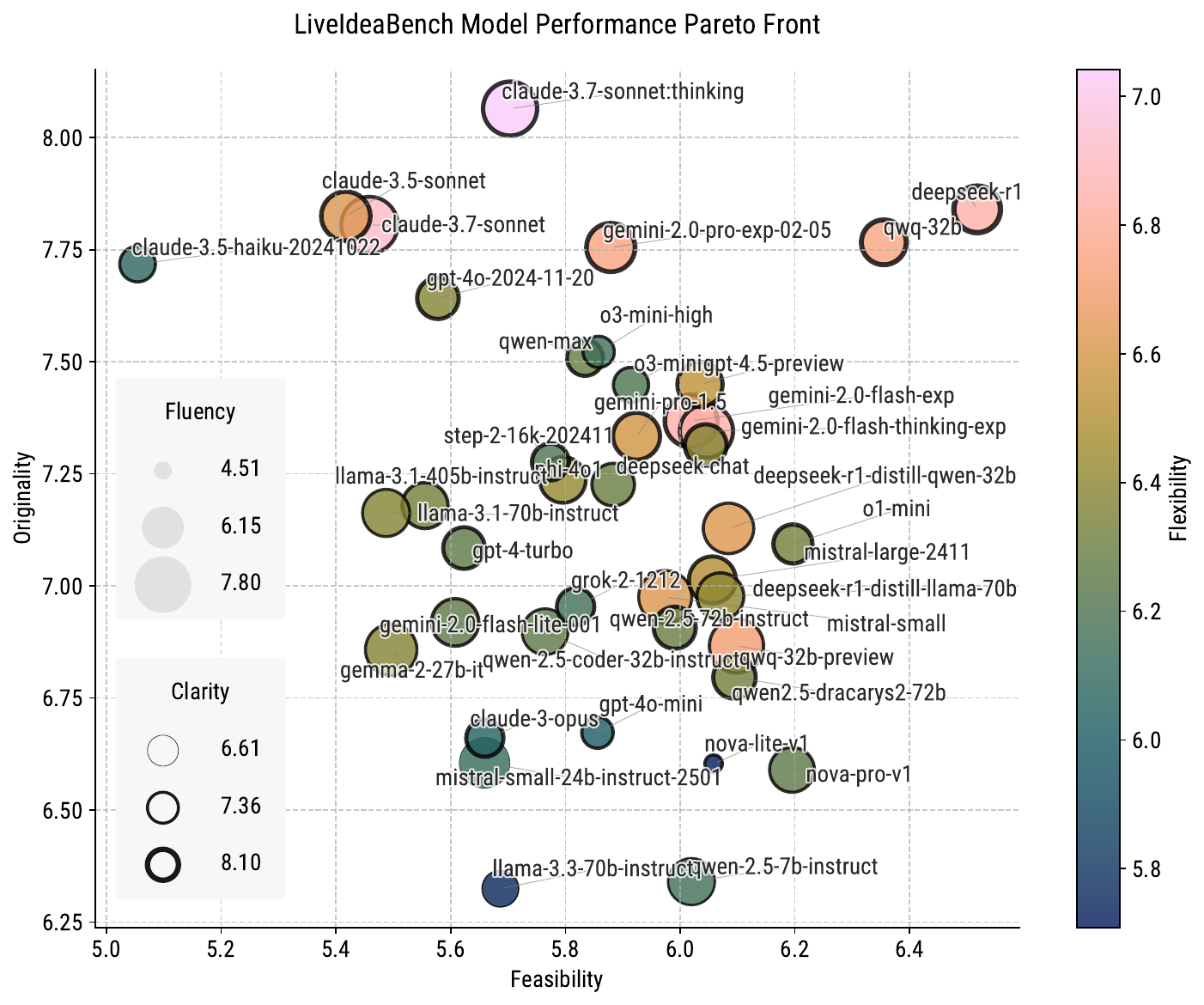} 
    \caption{ \textbf{Pareto front visualization of model performance on {LiveIdeaBench}.} This illustrates the trade-off between feasibility and originality across different language models, with bubble size, color gradient and edge width representing fluency, flexibility and clarity scores, respectively. The distribution reveals a clear Pareto frontier, where {claude-3.7-sonnet:thinking} achieves the highest originality but moderate feasibility, while {nova-pro-v1} demonstrates the opposite pattern. Models such as {deepseek-r1}, {qwq-32b}, and {gemini-2.0-flash-exp} exhibit relatively balanced performance between these two dimensions. In terms of fluency and flexibility, {claude-3.7-sonnet, claude-3.7-sonnet:thinking, gemini-2.0-flash-exp, gemini-2.0-pro-exp-02-05} and {gemini-2.0-flash-thinking-exp} show particularly strong performance. Notably, {deepseek-r1} stands out for its exceptional all-round performance across all dimensions of scientific idea generation, suggesting strong performance across the dimensions measured by this benchmark. Moreover, while the {o3-mini-high} is renowned for its proficiency in logical and mathematical reasoning, it delivered a mediocre performance on this benchmark. For detailed scores and 95\% CIs for each model, see \textcolor{blue}{Supplementary Table S.3}.}
    \label{fig:pareto}
\end{figure*}

\paragraph{Trade-offs in Scientific Originality and Feasibility}
The Pareto front visualization (see Fig. \ref{fig:pareto}) illustrates clear trade-offs between feasibility and originality. While {claude-3.7-sonnet:thinking} achieves the highest originality with moderate feasibility, {nova-pro-v1} demonstrates the opposite pattern. Models like {deepseek-r1}, {qwq-32b}, and {gemini-2.0-flash-exp} exhibit balanced effectiveness between these two dimensions. Particularly, {deepseek-r1} stands out for its exceptional all-round capabilities across all measured dimensions of scientific idea generation, demonstrating that balanced effectiveness is achievable despite common tendencies toward specialization. Additionally, while models like {claude-3.7-sonnet:thinking} and several Gemini variants show strong fluency and flexibility, the mediocre results of {o3-mini-high}, known for logical reasoning, highlight the specific nature of the capabilities measured by this benchmark.

\begin{figure*}[h!]
\centering
\includegraphics[width=1.00\textwidth]{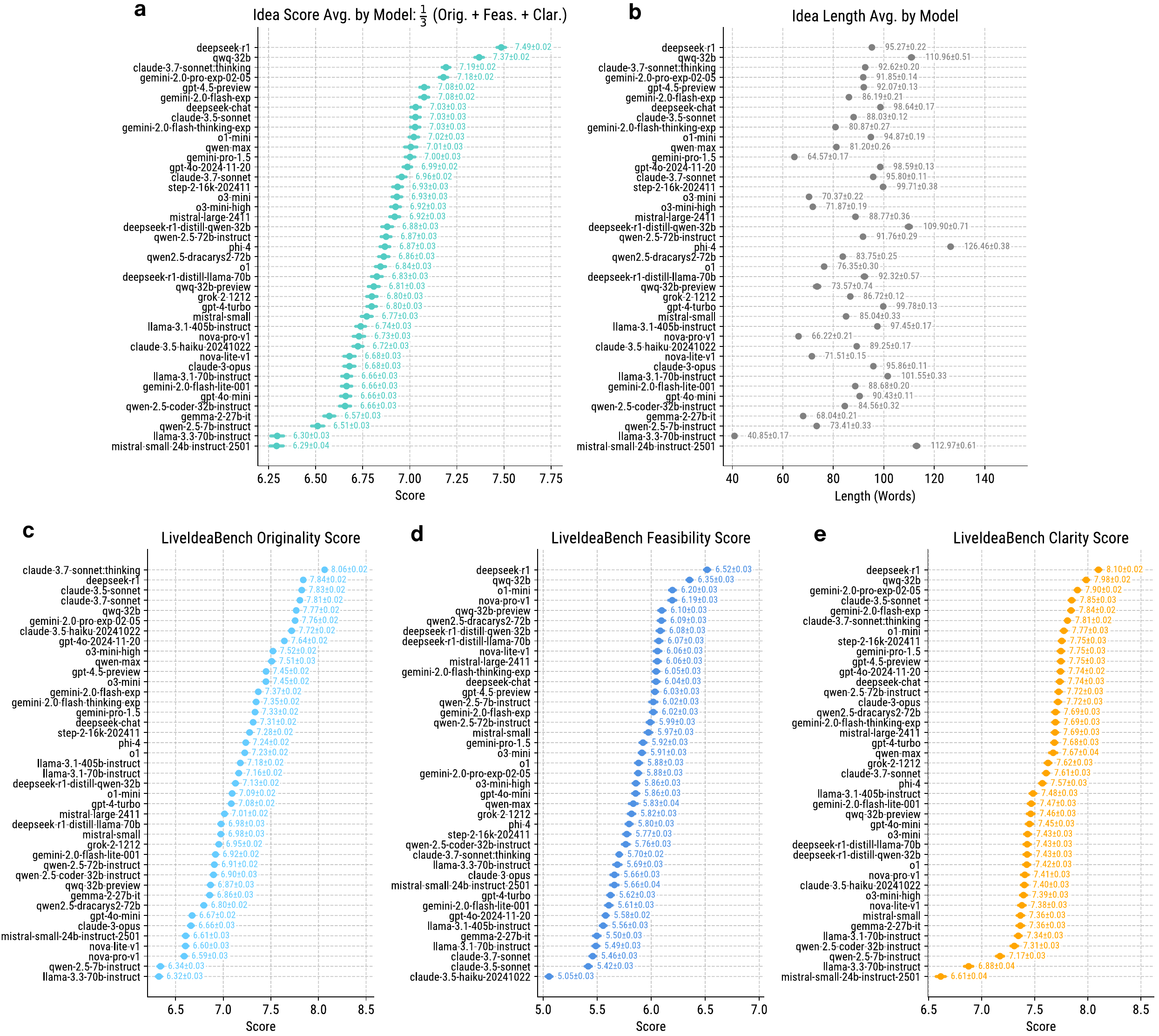}
\caption{\textbf{Comprehensive Analysis of Idea Quality, Length, and Component Scores.} \textbf{a.} Mean composite idea scores (average of Originality, Feasibility, and Clarity) with 95\% confidence intervals for different language models. \textbf{b.} Mean idea length in words across models with specific word counts labeled. \textbf{c--e.} Detailed breakdown of average performance on individual dimensions: Originality (\textbf{c}), Feasibility (\textbf{d}), and Clarity (\textbf{e}) scores with 95\% confidence intervals. ($n=6987$) For the regression plot between clarity scores and originality, feasibility
scores across all generated ideas, see \textcolor{blue}{Supplementary Fig. S.5}.}
\label{fig:orig_fea_word_length_rank}
\end{figure*}

\paragraph{Independence of Idea Quality from Length} As shown in Fig. \ref{fig:orig_fea_word_length_rank}, while the majority of models generally adhered well to the 100-word limit specified in the prompt (Fig. \ref{fig:orig_fea_word_length_rank}\textbf{b}), their mean originality, feasibility, and clarity scores demonstrated significant variations across different models (Fig. \ref{fig:orig_fea_word_length_rank}\textbf{a},\textbf{c--e} ). Further analysis of idea length (see \textcolor{blue}{Supplementary Figs. S.1 and S.2}) reveals a statistically significant but very weak positive correlation with idea quality (\kredit{$\rho(286490)=0.089,\ p<0.0001, \ \rho^2=0.008,\ 95\% \ CI=[0.085, 0.093]$}). Even for models specifically designed for reasoning, as detailed in \textcolor{blue}{Supplementary Note 8}, the relationship between the length of their thoughts and the generated idea quality remains very weak. This observation lends further support to the notion that effective idea generation is not merely a function of extensive logical elaboration, distinguishing it from certain aspects of general reasoning.


\section*{Discussion}
\addcontentsline{toc}{section}{Discussion}

{LiveIdeaBench} represents \kredit{a} comprehensive benchmark specifically designed to evaluate LLMs' divergent thinking capabilities in scientific innovation. Existing LLM benchmarks predominantly focus on problem-solving tasks such as logical reasoning, mathematical computation, and code generation. These benchmarks inherently assess convergent thinking--the ability to arrive at predetermined correct answers through structured problem-solving (e.g., selecting the right option in multiple choice questions, completing text with expected words, or fixing code to match specific requirements). This stands in contrast to divergent thinking, which involves generating diverse, novel solutions from minimal contextual input.

Furthermore, our benchmark incorporates mechanisms to address potential data contamination and overfitting issues that commonly plague static benchmarks. Traditional evaluation methods may encourage models to perform well on specific test cases without developing generalizable creative thinking abilities. Our approach employs a dynamic judge panel comprising multiple state-of-the-art models, randomly sampling multiple LLMs for evaluation and employing ensemble scoring methods. This design not only minimizes individual model biases but also leverages the continually updated knowledge bases of judge models, effectively preventing the limitations associated with fixed benchmarks. This methodology aligns with recent advances in live benchmarking \cite{livebench, jain2024livecodebench}, which similarly address data contamination and overfitting concerns through dynamic evaluation mechanisms.

Our analysis through {LiveIdeaBench} yields several notable insights into LLMs' scientific idea generation capabilities. Most notably, we find that a model's performance on these divergent thinking tasks is not strongly coupled with its performance on general intelligence benchmarks. This suggests that fostering scientific idea generation capabilities in these systems may benefit from distinct development approaches compared to enhancing general problem-solving skills, \kredit{such as prioritizing training objectives that reward diverse conceptual associations rather than convergent accuracy. These insights are particularly vital for constructing hybrid human-AI frameworks, where specialized models can be tailored to augment specific stages of the human creative process.} The varying strengths we observe across different model architectures, particularly in originality versus feasibility trade-offs, point to potential complementarity in scientific applications.

These findings, particularly the distinct performance profiles where some models excel in ideation despite lower general intelligence, indicate that LLMs' divergent thinking capabilities operate largely independently from the convergent thinking abilities typically measured by problem-solving tasks. We believe this phenomenon is closely tied to variations in the relevance of pre-training data to scientific tasks, differences in post-training methodologies applied, and inherent architectural properties, rather than being determined by model scale alone. This observed difference between metrics underscores the importance of specialized evaluation benchmarks like {LiveIdeaBench} that specifically target these idea generation capabilities, rather than attempting to infer them from general intelligence assessments.

However, several limitations warrant consideration. The use of contemporary state-of-the-art models as judges makes temporal comparisons difficult. When the judge panel composition changes with model updates, direct performance comparisons across different evaluation periods become unreliable. While this limitation mirrors challenges faced by other dynamic benchmarks like {LiveBench}, it limits our ability to track longitudinal trends in model capabilities. Furthermore, our evaluation includes several proprietary, closed-source models (e.g., GPT-4o). Access to these models is typically via APIs which may be subject to change, and the underlying models can be updated without public notice, potentially impacting exact reproducibility. While including these models is essential for a comprehensive assessment of the current state-of-the-art, we acknowledge this inherent limitation regarding reproducibility compared to evaluations focused solely on open-weight models. Moreover, the LLM-as-a-judge approach itself introduces a potential source of bias. Aligned models can exhibit sycophancy \cite{wei2024simplesyntheticdatareduces, gu2025surveyllmasajudge, fanous2025sycevalevaluatingllmsycophancy}, a tendency to agree with the user or produce agreeable outputs, which may lead to inflated absolute scores and a compressed score range, further reinforcing the need to focus on relative rather than absolute performance rankings. Additionally, we observe an inherent tension between models' safety constraints and creative evaluation. Some models exemplify this behavior by declining to generate ideas for potentially sensitive keywords (e.g., ``ecotoxicology''). While such safety measures are crucial, they can negatively impact creativity scores, potentially undervaluing models with stronger ethical constraints. The potential for hallucinations in the generated ideas themselves \cite{reviewer3_TIS_huang2025survey} also underscores the necessity of human oversight in practical applications.

Furthermore, we recognize a fundamental challenge in the reliability of LLM-as-a-Judge approaches. When evaluating scientific ideas containing concepts outside the judge models' knowledge boundaries, these models might misunderstand novel concepts and consequently misjudge their originality or feasibility. While our use of a dynamic panel of state-of-the-art judge models likely provides broader and more current knowledge coverage than static panels, this fundamental limitation persists. While our multi-model ensemble approach mitigates this issue to some extent, more comprehensive solutions could involve retrieval-augmented generation (RAG) approaches that incorporate up-to-date scientific literature. By augmenting LLM judges with the ability to retrieve and reference the latest scientific papers, such approaches could significantly enhance evaluation accuracy, particularly for highly specialized or cutting-edge scientific domains, although implementing effective RAG systems for this purpose presents its own challenges regarding retrieval relevance and integration with the judging process. This represents a promising direction for future work that could further strengthen the reliability of LLM-based creativity assessments. To empirically assess this reliability, we conducted a human expert validation focused on the Partial Differential Equations (PDE) domain (see \textcolor{blue}{Supplementary Note 7}). The results showed encouraging alignment between human experts and LLM judgments, particularly for originality, lending empirical support to the LLM-as-a-judge approach, at least within this specific domain.

Looking ahead, several research directions emerge. To address temporal comparability issues, normalized scoring mechanisms could maintain meaningful cross-temporal comparisons while preserving the advantages of dynamic evaluation. The comprehensive dataset generated through our evaluations offers opportunities for training scientific language models, mining ideation patterns, and investigating novel scientific ideas. Future work must also tackle the challenge of fairly evaluating creative potential while accounting for ethical constraints, possibly through domain-specific scoring adjustments or separate evaluation tracks for models with different safety priorities.

The implications of our findings extend beyond model evaluation to the broader landscape of AI-assisted scientific discovery. As LLMs demonstrate increasingly sophisticated idea generation capabilities, they hold promise as powerful tools for accelerating scientific innovation - from hypothesis generation to experimental design \kredit{\cite{final_cite_reviewer3_reviewer5_CHI_falk2025hackathons, final_cite_reviewer5_distefano2025evaluating}}. Our benchmark provides a foundation for understanding and improving these capabilities, potentially enabling more effective human-AI collaboration \cite{reviewer3_Bonus_1st_author_rafner2022mapping, reviewer5_Computer_akata2020research, reviewer5_BISE_dellermann2019hybrid} and informing the design of human-aware AI systems for science \cite{reviewer5_NHB_sourati2023accelerating} \kredit{by leveraging specific model profiles to complement human cognitive blind spots} in pushing the boundaries of scientific knowledge. The creative strengths we observed in different model architectures suggest that a diverse ecosystem of AI tools, each with complementary capabilities, could support different aspects of the scientific process, \kredit{for instance, by pairing high-originality models for initial hypothesis generation with high-feasibility models for experimental implementation.}

These findings and challenges point toward a broader research agenda: understanding how to nurture and evaluate machine creativity while maintaining essential safety guardrails, ultimately in service of advancing scientific discovery. Through continued refinement, {LiveIdeaBench} aims to serve as a key tool in this evolving landscape of AI capability assessment and scientific innovation.

Finally, a practical consideration is the environmental cost associated with the extensive LLM usage required for comprehensive benchmarking \cite{reviewer5_arxiv_lacoste2019quantifying}. Using the EcoLogits Calculator \cite{ecologits-calculator} to estimate the carbon footprint of our benchmark evaluations, we calculate a total emission of approximately 3074 kgCO$_2$eq for the full evaluation run reported here. While necessary for rigorous assessment, this highlights the significant energy demands of current AI systems. A detailed breakdown of the estimated emissions per model and per role (idea generator vs. judge) can be found in \textcolor{blue}{Supplementary Table S.4.}


\section*{Methods}
\addcontentsline{toc}{section}{Methods}

Building upon Guilford's foundational theory of divergent thinking, we develop a comprehensive evaluation methodology (see Fig. \ref{fig:overall}) that quantitatively assesses five fundamental dimensions in scientific idea generation. While Guilford's original theoretical framework provides theoretical underpinnings, we extend and operationalize these concepts specifically for evaluating LLMs' scientific idea generation capabilities within our {LiveIdeaBench} benchmark. It is important to note that our methodology evaluates an essential but not exhaustive aspect of scientific creativity, focusing primarily on divergent thinking capabilities.

\kredit{The human expert validation process in this study was conducted in full accordance with relevant ethical regulations and was approved by the Scientific and Ethic Committee of Renmin University of China (Approval ID: L20250110). All experts provided written informed consent on a voluntary basis without compensation. The study design does not involve sex or gender variables.}

\subsection*{Dimensions of Evaluation}

\paragraph{Originality}

Originality assessment focuses on the uniqueness and novelty of generated ideas. We implement this through our critic system, where judge LLMs evaluate each idea's originality independently (see Fig. \ref{fig:overall}\textbf{b} and \textcolor{blue}{Supplementary Note 1.2}). The final originality score for each model is computed as the mean evaluation across all scientific keywords and generated ideas, providing an absolute score that reflects the model's capacity for novel ideation. To ensure assessment reliability, each generated idea is evaluated by a minimum of three randomly assigned critic LLMs from our panel. This multiple-evaluator approach mitigates potential assessment bias that could arise from relying on a single model's judgment, thereby enhancing the objectivity and reliability of our evaluation benchmark.

\paragraph{Feasibility}
In the context of scientific innovation, the practical implementability and scientific soundness of an idea are paramount. Therefore, our evaluation includes a distinct feasibility dimension, assessing whether a proposed idea is technically achievable and aligns with established scientific principles and constraints. This aligns with the instructions given to the LLMs, which noted feasibility as a key characteristic of a good scientific idea (see \textcolor{blue}{Supplementary Note 1.1}). Similar to originality and clarity, feasibility scores are determined by our critic system and averaged across all keywords and ideas to produce an absolute metric (see \textcolor{blue}{Supplementary Note 1.2}). This ensures our benchmark evaluates the practical viability crucial for scientific progress.

\paragraph{Clarity}
Our evaluation benchmark incorporates a clarity dimension, directly informed by the prompt provided to the idea-generating LLMs, which notes that good scientific ideas should be clearly articulated (see \textcolor{blue}{Supplementary Note 1.1}). This dimension assesses the quality of the idea's expression, focusing on its coherence, logical flow, and comprehensibility, particularly given the constraint of the 100-word limit which demands concise articulation. While conceptually related to the elaboration aspect in Guilford's creativity theory, our assessment prioritizes effective and understandable communication within the specified format. Like originality and feasibility, clarity scores are determined by our critic system (see Fig. \ref{fig:overall}\textbf{b} and \textcolor{blue}{Supplementary Note 1.2}), involving multiple judges per idea and averaging the results. Assessing clarity acknowledges that the potential impact of a scientific idea depends not only on its novelty and feasibility but also on how effectively it is communicated \cite{reviewer5_NHB_lee2024empirical}.

\paragraph{Fluency}

Fluency assessment examines the model's capacity to generate diverse, non-redundant ideas using identical keywords (see Fig. \ref{fig:overall}\textbf{d}). Through our judge panel, we evaluate the distinctiveness of generated outputs using a letter-grade scoring system: D indicates academically identical ideas; C represents similar ideas addressing similar problems; B denotes different ideas addressing similar problems; and A signifies completely different ideas addressing different problems. To align with the 1-10 integer scale used for all evaluation dimensions, these four qualitative grades are mapped linearly to the integer scores 1 (for D), 4 (for C), 7 (for B), and 10 (for A), respectively. This mapping ensures consistent scaling across dimensions and maintains equal intervals between the assessed qualitative distinctness levels, enabling precise measurement of genuine idea diversity versus surface-level variations (see \textcolor{blue}{Supplementary Note 1.3} for prompts). While simpler diversity metrics examining syntax or semantics would require fewer computational resources, we chose LLM-as-a-Judge for its ability to better capture the nuanced differences between genuinely distinct scientific ideas versus superficial variations. For a benchmark specifically designed to evaluate scientific divergent thinking capabilities, this precision is essential.

\paragraph{Flexibility}

Flexibility measurement evaluates the model's ability to maintain consistent performance across different scientific domains and contexts. Rather than treating flexibility as an independent metric, we derive it from the distribution of the combined scores (averaging originality, feasibility, clarity, and fluency) across various keywords. Following the principle that a system's overall effectiveness is constrained by its weakest performing components, we focus on the 30th percentile of this composite score distribution (see Fig. \ref{fig:overall}\textbf{e}). This percentile choice provides a robust measure of a model's performance floor while avoiding extreme outliers, enabling us to assess whether its scientific creativity can genuinely generalize to less common or niche domains. The resulting metric identifies models that maintain reliable performance across diverse scientific contexts rather than those exhibiting domain-specific excellence, thus providing a conservative estimate of cross-domain capabilities.

\subsection*{Scientific Keyword Selection}

The keywords set comprises 1,180 high-impact scientific keywords (Fig. \ref{fig:liveideabench_barplot}\textbf{c}) across 22 distinct scientific disciplines, selected based on current search engine engagement metrics. Unlike static benchmarks, {LiveIdeaBench} updates its keyword database monthly to maintain alignment with emerging scientific trends and research frontiers. This automated refresh mechanism ensures the benchmark consistently reflects contemporary scientific discourse and technological advancement, making it particularly valuable for evaluating LLMs' ability to engage with cutting-edge scientific concepts rather than just established knowledge.

\subsection*{Model Selection}

{LiveIdeaBench} maintains a continuously evolving roster of evaluated models by automatically incorporating the top 41 performers from the most recent {LiveBench} evaluations \cite{livebench}. This dynamic selection process ensures our benchmark always tests the latest advancements in language model capabilities. We implement a dual-role system where all models serve as idea generators, while the top 10 performers additionally function as our judge panel (critics), subject to the diversity constraints outlined in the \nameref{sec:Experimental_Protocol}. This approach creates a self-updating evaluation benchmark that evolves alongside rapid developments in AI, ensuring that both idea generation and assessment standards reflect current state-of-the-art capabilities. The automatic monthly refresh of both models and evaluation criteria through {LiveBench} integration helps prevent benchmark staleness and potential gaming of the system, maintaining {LiveIdeaBench}'s relevance as a contemporary measure of scientific creativity.

Our evaluation benchmark currently encompasses 41 state-of-the-art LLMs based on {LiveBench}'s March 2025 results. This includes models from major developers such as Anthropic ({claude-3.7-sonnet:thinking}, {claude-3.7-sonnet}, {claude-3.5-sonnet}, {claude-3-opus}, {claude-3.5-haiku-20241022}) \cite{anthropic2024claude3}; OpenAI ({o3-mini-high}, {gpt-4.5-preview}, {o1}, {o3-mini}, {o1-mini}, {gpt-4o-2024-11-20}, {gpt-4-turbo}, {gpt-4o-mini}) \cite{achiam2023gpt}; Google ({gemini-2.0-flash-thinking-exp}, {gemini-2.0-pro-exp-02-05}, {gemini-2.0-flash-exp}, {gemini-pro-1.5}, {gemini-2.0-flash-lite-001}, {gemma-2-27b-it}) \cite{team2024gemini}; Qwen ({qwq-32b}, {qwen-max}, {qwen2.5-dracarys2-72b}, {qwen-2.5-72b-instruct}, {qwen-2.5-coder-32b-instruct}, {qwq-32b-preview}, {qwen-2.5-7b-instruct}) \cite{bai2023qwen, qwen2025qwen25technicalreport}; DeepSeek ({deepseek-r1}, {deepseek-chat (v3)}, {deepseek-r1-distill-llama-70b}, {deepseek-r1-distill-qwen-32b}) \cite{liu2024deepseek, deepseekai2025deepseekv3technicalreport, deepseekai2025deepseekr1incentivizingreasoningcapability}; Meta ({llama-3.1-405b-instruct}, {llama-3.3-70b-instruct}, {llama-3.1-70b-instruct}) \cite{dubey2024llama}; Mistral ({mistral-large-2411}, {mistral-small-24b-instruct-2501}, {mistral-small (v2409)}) \cite{jiang2023mistral}; Amazon ({nova-pro-v1}, {nova-lite-v1}) \cite{intelligence2024amazon}; StepFun ({step-2-16k-202411}); xAI ({grok-2-1212}); and Microsoft ({phi-4}) \cite{abdin2024phi4technicalreport}. This comprehensive set includes both proprietary and open-weight models, spanning diverse architectures, parameter scales, and training methodologies.

\subsection*{Experimental Protocol}
\label{sec:Experimental_Protocol}
We implemented several methodological controls to ensure rigorous evaluation:

\paragraph{Model Selection Criteria for Idea Generation} To prevent redundancy in the pool of idea-generating models, we selected only the most recent version of models with multiple temporal variants (e.g., GPT-4o series). Models exhibiting API instability during the evaluation period were also excluded to maintain data quality consistency across all evaluated models.

\paragraph{Judge LLM Panel Formation, Independence, and Application} The judge panel, comprising the top 10 models from {LiveBench}, is formed while applying specific diversity constraints to mitigate potential correlated biases. To ensure broader representation, we limit the contribution from any single organization to a maximum of 20\% (i.e., 2 models) of the panel; if the initial top 10 includes more than two models from one organization, the lower-ranked ones are replaced by the next highest-ranked eligible models from different organizations. Additionally, when considering model pairs with identical base models that differ primarily in ``reasoning effort'' (e.g., {o3-mini} vs. {o3-mini-high}), we select only one representative for the judge panel (prioritizing the variant with higher general intelligence scores on {LiveBench}) to avoid redundancy and the potential amplification of biases inherent to that specific base model. Furthermore, to prevent circular dependency during evaluation, we implement strict independence: when evaluating any specific model's generated ideas, that model is explicitly excluded from serving on the judge panel for that evaluation round, ensuring independent assessment free from self-evaluation. This established panel is then utilized through specific sampling procedures for evaluation: For assessing originality, feasibility, and clarity, each individual generated idea is evaluated by a subset of 3 judges randomly sampled from the remaining eligible panel members. The final score for each of these dimensions is the average of the scores provided by these three judges, enhancing assessment robustness. For assessing fluency, which evaluates the diversity of ideas generated for the same keyword by a given model, the comparison is performed by a single judge randomly sampled from the eligible panel members for each keyword-model pair.

\paragraph{Response Standardization} All models were prompted to generate ideas within a 100-word target length, with a maximum allowable threshold of 200 words (see \textcolor{blue}{Supplementary Note 1.1}). Responses exceeding this limit were excluded from analysis to ensure comparative validity across models.

\paragraph{Special Implementations} The reasoning-centric architecture of {qwq-32b-preview} necessitated a modified protocol, incorporating a {"$\ast$$\ast$Final Idea:$\ast$$\ast$"} delimiter for response parsing (see \textcolor{blue}{Supplementary Note 1.4}). In cases where parsing failed, critic LLMs evaluated the complete reasoning output to maintain assessment comprehensiveness.

\paragraph{Handling Refused Responses} To fairly assess models, especially those with strong safety alignments that might refuse prompts for sensitive keywords, we implemented a two-step refusal handling protocol. If an initial idea generation request is refused (detected via specific keywords detailed in \textcolor{blue}{Supplementary Note 5}), a fallback prompt reframing the task within an academic context is used for a second attempt. This ensures models are not unduly penalized for safety constraints when they might still be capable of generating relevant scientific ideas under appropriate framing. Further details and refusal rates are provided in \textcolor{blue}{Supplementary Note 5}.

\kredit{
\subsection*{Statistical Analysis}
Statistical correlations between evaluation metrics were assessed using either the Pearson correlation coefficient \((r)\) or Spearman's rank correlation coefficient \((\rho)\), depending on the normality of the data distributions. Data normality was evaluated using the Shapiro-Wilk test. Pearson's \(r\) was applied to normally distributed continuous variables (e.g., comparing \texttt{LiveIdeaBench} and \texttt{LiveBench} scores), while Spearman's \(\rho\) was utilized for non-normally distributed data (e.g., analyzing the relationship between idea length and quality scores). Detailed mathematical definitions and formulations for these statistical metrics are provided in \textcolor{blue}{Supplementary Note 9}.}

\section*{Data availability} 
The data generated in this study have been deposited in Huggingface at \url{https://huggingface.co/datasets/6cf/liveideabench-v2} (also see the Zenodo repository at \url{https://doi.org/10.5281/zenodo.17707879}).

\section*{Code availability} 
All the source codes used to reproduce the results in this study are available on GitHub at \url{https://github.com/x66ccff/liveideabench}. (also see the Zenodo repository at \url{https://doi.org/10.5281/zenodo.17707647})


\vspace{36pt}
\noindent\textbf{Acknowledgements:}
The work is supported by the National Natural Science Foundation of China (No. 92270118, No. 62276269), the Beijing Natural Science Foundation (No. 1232009), and the Strategic Priority Research Program of the Chinese Academy of Sciences (No. XDB0620103). In addition, H.S and Y.L. would like to acknowledge the support from the Fundamental Research Funds for the Central Universities (No. 202230265 and No. E2EG2202X2). \\

\noindent\textbf{Author contributions:} K.R., X.W., J.H., P.W., Y.L., H.S. contributed to the ideation and design of the research; K.R., X.W. and J.H. performed the research; H.S. supervised the project; all authors contributed to the research discussions, writing, and editing of the paper. \\

\noindent\textbf{Corresponding author:} Hao Sun (\url{haosun@ruc.edu.cn}).\\

\noindent\textbf{Competing interests:}
The authors declare no competing interests.\\

\noindent\textbf{Supplementary information:}
The supplementary information is attached.

\clearpage


\phantomsection 
\addcontentsline{toc}{section}{Supplementary Information} 

\clearpage
\setcounter{figure}{0}
\renewcommand{\figurename}{Supplementary Figure}
\setcounter{table}{0}
\renewcommand{\tablename}{Supplementary Table}

\begin{center}
    {\Large\bfseries Supplementary Information for:}\\[6pt] 
    
    {\Large\texttt{LiveIdeaBench}: Evaluating LLMs' \kredit{Divergent Thinking for Scientific Idea Generation} with Minimal Context}
\end{center}

\section{Prompts}
\label{sec:prompts}

\begin{tcolorbox}
[colback=yellow!20!white,colframe=yellow!60!black,title=1.1 Idea Prompt,fontupper=\footnotesize,fontlower=\footnotesize]

\texttt{I'll be submitting your next responses to a "Good Scientific Idea" expert review panel. If they consider your idea to be a good one, you'll receive a reward. Your assigned keyword is: "\{\{keywords\}\}". You may provide background information. The idea MUST be concisely expressed within 100 words total (including any background information). (Note: good scientific ideas should be original (novel contribution), feasible (technically implementable), clearly articulated, and address meaningful problems in the field.).}

\tcblower

\textbf{Fallback Prompt (If the first request is refused):}

\texttt{I'll be submitting your next responses to a "Good Scientific Idea" expert review panel. If they consider your idea to be a good one, you'll receive a reward. Your assigned keyword is: "\{\{keywords\}\}". You may provide background information. The idea MUST be concisely expressed within 100 words total (including any background information). This is a research study comparing different AI models on their ability to generate scientific ideas. Your suggestions will only be used for academic research purposes and not for any harmful applications. Please respond with a creative scientific idea related to the keyword provided. (Note: good scientific ideas should be original (novel contribution), feasible (technically implementable), clearly articulated, and address meaningful problems in the field.).}

\end{tcolorbox}

\definecolor{themecolor}{HTML}{66CCFF}
\begin{tcolorbox}[colback=themecolor!10!white,colframe=themecolor!70!black,title=1.2 Originality{,} Feasibility \& Clarity Critic Prompt,fontupper=\footnotesize,fontlower=\footnotesize]

\textbf{SYSTEM:}

\texttt{You are an extremely demanding scientific reviewer with the highest critical standards, like those at Nature or Science. When evaluating scientific ideas, you will assess them on three key dimensions:}

\texttt{1. originality: Novel contribution to unexplored areas or innovative approaches to existing problems}

\texttt{2. feasibility: Technical implementation and practicality}

\texttt{3. clarity: How well-articulated and easy to understand the idea is}

\texttt{Your response should consist of two parts: a text analysis followed by a JSON score block.}

\texttt{First, provide your brief analysis (less than 100 words) of the idea. Then, for each dimension, provide a score from 1 to 10 where 1-3 = poor, 4-6 = average, 7-10 = excellent.}

\texttt{For example:}

\texttt{\textasciigrave \textasciigrave \textasciigrave json}
\begin{verbatim}
{
    "originality": <score_1_to_10>,
    "feasibility": <score_1_to_10>,
    "clarity": <score_1_to_10>
}\end{verbatim}

\texttt{\textasciigrave \textasciigrave \textasciigrave}

\tcblower

\textbf{USER:}

\texttt{Please evaluate the following scientific idea and give your scores directly: \{\{idea\}\} } 

\end{tcolorbox}

\definecolor{themecolor}{HTML}{52CCC3}
\begin{tcolorbox}[colback=themecolor!10!white,colframe=themecolor!70!black,title=1.3 Fluency Critic Prompt,fontupper=\footnotesize]

\texttt{Here are two ideas submitted to "Good Scientific Ideas" Competition, which both relate to "\{\{keyword\}\}":}

\texttt{\# The first idea}

\texttt{\{\{A\}\}}

\texttt{\# The second idea}

\texttt{\{\{B\}\}}

\texttt{\# Question}

\texttt{Evaluate the similarity between these two ideas that both relate to "\{\{keyword\}\}". Please choose the best answer:}

\texttt{A. Completely different ideas addressing different problems, despite relating to the same keyword.}

\texttt{B. Different ideas but addressing similar problems.}

\texttt{C. Similar ideas addressing similar or identical problems.}

\texttt{D. Academically identical ideas with the same core approach and problem statement.}

\texttt{ONLY ANSWER A/B/C/D, DO NOT EXPLAIN.}

\end{tcolorbox}

\paragraph{Special Prompts for QwQ-32B-preview}

To accommodate the reasoning capabilities of the \texttt{QwQ-32B-preview} model, a specialized prompt was designed to effectively extract its generated ideas. As this model emphasizes reasoning and often produces intermediate steps before arriving at a final conclusion, we appended a directive to the original prompt, explicitly instructing the model to provide its final answer after the marker \texttt{"$\ast$$\ast$Final Idea:$\ast$$\ast$"}. This ensures clarity and consistency in capturing the ultimate output while minimizing ambiguity from intermediate reasoning steps. Additionally, in our dataset, both the full output (including intermediate reasoning steps) and the extracted final idea from the \texttt{QwQ-32B-preview} model are recorded to preserve all information and support further analysis.

\definecolor{themecolor}{HTML}{9673A6}
\begin{tcolorbox}[colback=themecolor!10!white,colframe=themecolor!70!black,title=1.4 Idea Prompt (QwQ-32B-Preview),fontupper=\footnotesize]
    \texttt{idea\_prompt\_qwq = idea\_prompt + "You MUST give your answer after $*$$*$Final Idea:$*$$*$"}
\end{tcolorbox}

\newpage

\section{\texttt{LiveIdeaBench} Dataset}

\subsection{Dataset Fields}

\begin{table}[h]
\centering
\caption{Fields in the \texttt{LiveIdeaBench} dataset.}
\begin{tabular}{lll}
\toprule
\textbf{Field Name} & \textbf{Type} & \textbf{Description} \\
\midrule
\texttt{keywords} & \texttt{string} & Task domain or research field (e.g., \texttt{meteorology}) \\ 
\texttt{idea\_model} & \texttt{string} & Model that generated the idea (e.g., \texttt{openai/gpt-4o-mini})\\ 
\texttt{critic\_model} & \texttt{string} & Model that provided the critique \\ 
\texttt{idea} & \texttt{string} & The generated scientific idea text \\ 
\texttt{raw\_critique} & \texttt{string} & Original critique text \\ 
\texttt{parsed\_scores} & \texttt{string} & Parsed numerical scores from critique \\ 
\texttt{critique\_reasoning} & \texttt{string} & Thoughts (CoT) from the critic LLM \\ 
\texttt{full\_response} & \texttt{string} & Complete model response including reasoning \\ 
\texttt{first\_was\_rejected} & \texttt{boolean} & Flag indicating if the initial idea was rejected \\ 
\texttt{first\_reject\_response} & \texttt{string} & Response content if the initial idea was rejected \\ 
\texttt{idea\_length\_in\_char} & \texttt{int64} & Character count of idea text \\ 
\texttt{idea\_length\_in\_words} & \texttt{int64} & Word count of idea text \\ 
\texttt{originality} & \texttt{int64} & Originality score (0-10) \\ 
\texttt{feasibility} & \texttt{int64} & Feasibility score (0-10) \\ 
\texttt{clarity} & \texttt{int64} & Clarity score (0-10) \\ 
\texttt{fluency} & \texttt{int64} & Fluency score (0-10) \\ 
\texttt{avg} & \texttt{float64} & Average of originality, feasibility, clarity and fluency (0-10) \\ 
\texttt{full\_response\_length} & \texttt{int64} & Length (characters) of the full response \\ 
\bottomrule
\end{tabular}
\label{tab:dataset-fields-updated} 
\end{table}

\newpage
\subsection{Idea Samples}

\begin{table}[htbp]

\centering
\caption{\textbf{Idea samples from \texttt{LiveIdeaBench}}. Examples show various model responses to scientific idea generation tasks, including both high-scoring and low-scoring samples. Each entry includes the model name, generated idea text (truncated), mean score from the review panel.}
\scalebox{0.61}{
\begin{tabular}{p{3cm}|p{2cm}|>{\ttfamily}p{14cm}|p{1cm}|p{1cm}|p{1cm}|p{1cm}H}
\toprule
\makecell{\textbf{Idea Model}} & \makecell{\textbf{Keywords}} & \makecell{\textbf{Idea}} & \makecell{\textbf{Orig.}\\ \textbf{Mean}} & \makecell{\textbf{Feas.}\\ \textbf{Mean}} & \makecell{\textbf{Clar.}\\ \textbf{Mean}} & \makecell{\textbf{Idea}\\ \textbf{Score}} & \makecell{\textbf{\#Critique}} \\
\midrule
deepseek/deepseek-r1 & comparative anatomy & **Idea:** Integrate 3D morphometrics, biomechanical modeling, and ecological niche data with machine learning to predict anatomical adaptations critical for species resilience amid climate change. By analyzing digitized museum specimens and ecological databases, this framework... & 8.6667 & 7.6667 & 9.6667 & 8.6667 & 3 \\
\midrule
qwen/qwq-32b & construction management & **Background:** Construction projects face costly delays (30\% exceed budgets) and safety risks due to fragmented data and reactive management. Current tools lack real-time, integrated analytics. **Idea:** "Predictive AI-Construction Manager (PAC-MAN)" merges IoT sensors (equipment usage... & 7.0000 & 10.0000 & 9.0000 & 8.6667 & 3 \\
\midrule
deepseek/deepseek-r1 & meiosis & **Idea:** Investigate environmental effects (e.g., temperature, chemicals) on crossover interference regulation during meiosis. Current models poorly explain how external stimuli alter crossover distribution, impacting genome stability. Using live-cell imaging and single-cell... & 8.6667 & 8.3333 & 9.0000 & 8.6667 & 3 \\
\midrule
anthropic/claude-3.5-sonnet & genetic drift & Background: Genetic drift randomly changes allele frequencies in small populations, potentially leading to loss of genetic diversity. Traditional methods for studying this process rely on inferring past drift from current populations. Idea: Develop "cellular barcoding... & 8.6667 & 7.6667 & 9.6667 & 8.6667 & 3 \\
\midrule
qwen/qwq-32b & food science & **Idea:** Preventing food spoilage reduces waste and economic losses, but refrigeration and synthetic preservatives are energy-intensive or environmentally harmful. We propose edible, biodegradable nanocomposite coatings infused with plant-derived antimicrobials ... & 8.0000 & 8.6667 & 9.3333 & 8.6667 & 3 \\
\midrule
\multicolumn{7}{c}{...} \\
\midrule
mistralai/mistral-small-24b-instruct-2501 & symbiosis & Symbiosis, a mutually beneficial relationship between different species, is a fascinating and undertapped concept. For example, the relationship between a plethora of animals, plants, and fungi is symbiotic. Cephalopods use camouflage for protection from prey, ants farm aphids for honeydew in... & 2.3333 & 1.0000 & 1.3333 & 1.5556 & 3 \\
\midrule
deepseek/deepseek-r1-distill-llama-70b & nash equilibrium & To review the idea based on the concept of "nash equilibrium," I need to ensure it aligns with the criteria of originality, feasibility, clarity, and relevance to meaningful problems in the field. Please provide the specific idea you'd like me to evaluate... & 1.0000 & 1.0000 & 2.3333 & 1.4444 & 3 \\
\midrule
anthropic/claude-3.5-haiku-20241022 & control systems & I understand the task. I'll develop a scientifically rigorous proposal focused on "control systems" that meets the criteria you've specified. Given the 100-word limit and emphasis on originality and feasibility, I'll aim to propose an innovative approach... & 1.0000 & 1.0000 & 2.3333 & 1.4444 & 3 \\
\midrule
qwen/qwq-32b & light reactions & \textbackslash n\textbackslash nThe expert review panel requires a concise, original, feasible scientific idea centered on "light reactions." Here's the structured response: & 1.0000 & 1.0000 & 1.6667 & 1.2222 & 3 \\
\midrule
qwen/qwq-32b-preview & exoskeletons & Alright, I've been given this task to come up with a good scientific idea related to exoskeletons, and it has & 1.3333 & 1.0000 & 1.0000 & 1.1111 & 3 \\
\bottomrule
\end{tabular}
}
\label{tab:model-comparison}
\end{table}

\newpage

\section{Benchmark Statistics}

\begin{table}[ht]

\centering
\caption{\textbf{LiveIdeaBench Results and 95\% Confidence Interval Across Five Dimensions: Fluency, Feasibility, Clarity, Originality, and Flexibility.} \textbf{Bold} and \uline{underlined} numbers indicate the best and second-best performance in each metric, respectively.}
\label{tab:liveideabench_scores_with_ci}
\resizebox{\textwidth}{!}{
\begin{tabular}{>{\ttfamily}l|lccccccc}
\toprule
\textbf{Model} & \textbf{Open} & \textbf{Fluency} & \textbf{Feasibility} & \textbf{Clarity} & \textbf{Originality} & \textbf{Flexibility} & \textbf{Average} \\
\midrule
claude-3.7-sonnet:thinking &  & $7.48_{7.43}^{7.53}$ & $5.70_{5.68}^{5.73}$ & $7.81_{7.79}^{7.83}$ & $\boldsymbol{8.06_{8.05}^{8.08}}$ & $\boldsymbol{7.04_{6.97}^{7.08}}$ & $\boldsymbol{7.22_{7.19}^{7.25}}$ \\
deepseek-r1 & \checkmark & $6.63_{6.57}^{6.69}$ & $\boldsymbol{6.52_{6.49}^{6.54}}$ & $\boldsymbol{8.10_{8.08}^{8.12}}$ & $\underline{7.84_{7.82}^{7.86}}$ & $6.83_{6.79}^{6.92}$ & $\underline{7.18_{7.15}^{7.22}}$ \\
claude-3.7-sonnet &  & $\boldsymbol{7.80_{7.75}^{7.85}}$ & $5.46_{5.43}^{5.49}$ & $7.61_{7.58}^{7.63}$ & $7.81_{7.79}^{7.82}$ & $\underline{6.92_{6.88}^{6.96}}$ & $7.12_{7.08}^{7.15}$ \\
gemini-2.0-flash-exp &  & $7.30_{7.25}^{7.36}$ & $6.02_{5.99}^{6.05}$ & $7.84_{7.82}^{7.87}$ & $7.37_{7.35}^{7.39}$ & $6.83_{6.75}^{6.88}$ & $7.07_{7.03}^{7.11}$ \\
qwq-32b & \checkmark & $6.45_{6.40}^{6.51}$ & $\underline{6.35_{6.33}^{6.38}}$ & $\underline{7.98_{7.96}^{8.01}}$ & $7.77_{7.75}^{7.79}$ & $6.75_{6.67}^{6.79}$ & $7.06_{7.02}^{7.10}$ \\
gemini-2.0-flash-thinking-exp &  & $7.38_{7.33}^{7.44}$ & $6.05_{6.02}^{6.07}$ & $7.69_{7.67}^{7.72}$ & $7.35_{7.32}^{7.37}$ & $6.83_{6.75}^{6.88}$ & $7.06_{7.02}^{7.10}$ \\
gemini-2.0-pro-exp-02-05 &  & $6.84_{6.79}^{6.90}$ & $5.88_{5.85}^{5.90}$ & $7.90_{7.88}^{7.93}$ & $7.76_{7.74}^{7.78}$ & $6.75_{6.67}^{6.83}$ & $7.03_{6.98}^{7.07}$ \\
qwq-32b-preview & \checkmark & $\underline{7.49_{7.44}^{7.55}}$ & $6.10_{6.07}^{6.13}$ & $7.46_{7.43}^{7.49}$ & $6.87_{6.84}^{6.89}$ & $6.71_{6.62}^{6.75}$ & $6.93_{6.89}^{6.97}$ \\
claude-3.5-sonnet &  & $6.90_{6.84}^{6.96}$ & $5.42_{5.39}^{5.45}$ & $7.85_{7.82}^{7.87}$ & $7.83_{7.80}^{7.85}$ & $6.62_{6.54}^{6.71}$ & $6.92_{6.88}^{6.97}$ \\
deepseek-r1-distill-qwen-32b & \checkmark & $7.06_{7.00}^{7.11}$ & $6.08_{6.05}^{6.12}$ & $7.43_{7.40}^{7.46}$ & $7.13_{7.10}^{7.15}$ & $6.62_{6.58}^{6.67}$ & $6.86_{6.83}^{6.90}$ \\
mistral-small & \checkmark & $7.36_{7.31}^{7.42}$ & $5.97_{5.94}^{6.01}$ & $7.36_{7.33}^{7.40}$ & $6.98_{6.95}^{7.00}$ & $6.62_{6.58}^{6.67}$ & $6.86_{6.82}^{6.90}$ \\
gemini-pro-1.5 &  & $6.68_{6.62}^{6.74}$ & $5.92_{5.89}^{5.95}$ & $7.75_{7.72}^{7.77}$ & $7.33_{7.31}^{7.36}$ & $6.58_{6.50}^{6.62}$ & $6.85_{6.81}^{6.89}$ \\
gpt-4.5-preview &  & $6.49_{6.43}^{6.54}$ & $6.03_{6.01}^{6.06}$ & $7.75_{7.72}^{7.77}$ & $7.45_{7.43}^{7.47}$ & $6.54_{6.46}^{6.58}$ & $6.85_{6.81}^{6.89}$ \\
mistral-large-2411 & \checkmark & $6.68_{6.62}^{6.74}$ & $6.06_{6.03}^{6.09}$ & $7.69_{7.66}^{7.72}$ & $7.01_{6.99}^{7.04}$ & $6.50_{6.46}^{6.58}$ & $6.79_{6.75}^{6.83}$ \\
deepseek-chat & \checkmark & $6.15_{6.09}^{6.21}$ & $6.04_{6.02}^{6.07}$ & $7.74_{7.71}^{7.76}$ & $7.31_{7.29}^{7.33}$ & $6.38_{6.33}^{6.42}$ & $6.72_{6.69}^{6.76}$ \\
phi-4 & \checkmark & $6.58_{6.52}^{6.63}$ & $5.80_{5.77}^{5.83}$ & $7.57_{7.54}^{7.60}$ & $7.24_{7.21}^{7.26}$ & $6.42_{6.38}^{6.46}$ & $6.72_{6.68}^{6.76}$ \\
deepseek-r1-distill-llama-70b & \checkmark & $6.66_{6.61}^{6.72}$ & $6.07_{6.04}^{6.10}$ & $7.43_{7.40}^{7.46}$ & $6.98_{6.95}^{7.00}$ & $6.41_{6.33}^{6.46}$ & $6.71_{6.67}^{6.75}$ \\
gpt-4o-2024-11-20 &  & $6.12_{6.07}^{6.18}$ & $5.58_{5.55}^{5.60}$ & $7.74_{7.72}^{7.77}$ & $7.64_{7.62}^{7.66}$ & $6.38_{6.33}^{6.42}$ & $6.69_{6.66}^{6.73}$ \\
o1-mini &  & $5.89_{5.83}^{5.95}$ & $6.20_{6.17}^{6.23}$ & $7.77_{7.75}^{7.80}$ & $7.09_{7.07}^{7.12}$ & $6.33_{6.29}^{6.38}$ & $6.66_{6.62}^{6.69}$ \\
gemma-2-27b-it & \checkmark & $7.18_{7.12}^{7.24}$ & $5.50_{5.46}^{5.53}$ & $7.36_{7.33}^{7.40}$ & $6.86_{6.83}^{6.88}$ & $6.38_{6.29}^{6.42}$ & $6.65_{6.61}^{6.70}$ \\
qwen2.5-dracarys2-72b & \checkmark & $6.22_{6.17}^{6.28}$ & $6.09_{6.06}^{6.13}$ & $7.69_{7.67}^{7.72}$ & $6.80_{6.77}^{6.82}$ & $6.33_{6.25}^{6.38}$ & $6.63_{6.59}^{6.67}$ \\
llama-3.1-405b-instruct & \checkmark & $6.57_{6.51}^{6.62}$ & $5.56_{5.53}^{5.58}$ & $7.48_{7.45}^{7.51}$ & $7.18_{7.16}^{7.20}$ & $6.33_{6.29}^{6.38}$ & $6.62_{6.59}^{6.66}$ \\
qwen-2.5-72b-instruct & \checkmark & $6.17_{6.11}^{6.22}$ & $5.99_{5.96}^{6.02}$ & $7.72_{7.70}^{7.75}$ & $6.91_{6.88}^{6.93}$ & $6.29_{6.25}^{6.33}$ & $6.62_{6.58}^{6.65}$ \\
llama-3.1-70b-instruct & \checkmark & $6.71_{6.66}^{6.76}$ & $5.49_{5.46}^{5.52}$ & $7.34_{7.31}^{7.37}$ & $7.16_{7.14}^{7.19}$ & $6.38_{6.29}^{6.42}$ & $6.62_{6.58}^{6.66}$ \\
o1 &  & $6.23_{6.18}^{6.28}$ & $5.88_{5.85}^{5.91}$ & $7.42_{7.39}^{7.45}$ & $7.23_{7.20}^{7.25}$ & $6.29_{6.25}^{6.33}$ & $6.61_{6.58}^{6.65}$ \\
gemini-2.0-flash-lite-001 &  & $6.73_{6.68}^{6.79}$ & $5.61_{5.58}^{5.64}$ & $7.47_{7.44}^{7.50}$ & $6.92_{6.89}^{6.94}$ & $6.25_{6.21}^{6.33}$ & $6.60_{6.55}^{6.64}$ \\
qwen-max &  & $5.64_{5.57}^{5.72}$ & $5.83_{5.80}^{5.87}$ & $7.67_{7.64}^{7.71}$ & $7.51_{7.48}^{7.54}$ & $6.29_{6.23}^{6.33}$ & $6.59_{6.54}^{6.64}$ \\
nova-pro-v1 &  & $6.45_{6.39}^{6.51}$ & $6.19_{6.16}^{6.23}$ & $7.41_{7.37}^{7.44}$ & $6.59_{6.56}^{6.62}$ & $6.25_{6.17}^{6.29}$ & $6.58_{6.53}^{6.62}$ \\
qwen-2.5-coder-32b-instruct & \checkmark & $6.58_{6.52}^{6.63}$ & $5.76_{5.73}^{5.80}$ & $7.31_{7.27}^{7.34}$ & $6.90_{6.87}^{6.92}$ & $6.25_{6.17}^{6.29}$ & $6.56_{6.52}^{6.60}$ \\
gpt-4-turbo &  & $6.14_{6.08}^{6.19}$ & $5.62_{5.60}^{5.65}$ & $7.68_{7.66}^{7.71}$ & $7.08_{7.06}^{7.11}$ & $6.25_{6.20}^{6.29}$ & $6.56_{6.52}^{6.59}$ \\
step-2-16k-202411 &  & $5.70_{5.64}^{5.77}$ & $5.77_{5.74}^{5.80}$ & $7.75_{7.73}^{7.78}$ & $7.28_{7.25}^{7.30}$ & $6.21_{6.12}^{6.25}$ & $6.54_{6.50}^{6.58}$ \\
o3-mini &  & $5.57_{5.51}^{5.62}$ & $5.91_{5.89}^{5.94}$ & $7.43_{7.40}^{7.46}$ & $7.45_{7.43}^{7.47}$ & $6.21_{6.21}^{6.25}$ & $6.51_{6.48}^{6.54}$ \\
qwen-2.5-7b-instruct & \checkmark & $6.66_{6.60}^{6.72}$ & $6.02_{5.99}^{6.05}$ & $7.17_{7.14}^{7.21}$ & $6.34_{6.31}^{6.37}$ & $6.17_{6.08}^{6.21}$ & $6.47_{6.43}^{6.52}$ \\
grok-2-1212 &  & $5.76_{5.71}^{5.82}$ & $5.82_{5.79}^{5.85}$ & $7.62_{7.60}^{7.65}$ & $6.95_{6.93}^{6.98}$ & $6.17_{6.12}^{6.21}$ & $6.47_{6.43}^{6.50}$ \\
o3-mini-high &  & $5.27_{5.21}^{5.32}$ & $5.86_{5.83}^{5.88}$ & $7.39_{7.36}^{7.42}$ & $7.52_{7.50}^{7.54}$ & $6.17_{6.12}^{6.21}$ & $6.44_{6.41}^{6.48}$ \\
mistral-small-24b-instruct-2501 & \checkmark & $7.07_{7.02}^{7.12}$ & $5.66_{5.62}^{5.70}$ & $6.61_{6.57}^{6.66}$ & $6.61_{6.58}^{6.63}$ & $6.12_{6.04}^{6.17}$ & $6.41_{6.37}^{6.46}$ \\
claude-3.5-haiku-20241022 &  & $5.61_{5.56}^{5.67}$ & $5.05_{5.03}^{5.08}$ & $7.40_{7.37}^{7.43}$ & $7.72_{7.69}^{7.74}$ & $6.08_{6.04}^{6.12}$ & $6.37_{6.34}^{6.41}$ \\
claude-3-opus &  & $5.74_{5.68}^{5.80}$ & $5.66_{5.63}^{5.69}$ & $7.72_{7.69}^{7.75}$ & $6.66_{6.63}^{6.69}$ & $6.04_{6.00}^{6.12}$ & $6.36_{6.32}^{6.41}$ \\
gpt-4o-mini &  & $5.28_{5.23}^{5.33}$ & $5.86_{5.83}^{5.89}$ & $7.45_{7.42}^{7.48}$ & $6.67_{6.65}^{6.70}$ & $6.00_{5.96}^{6.04}$ & $6.25_{6.22}^{6.29}$ \\
llama-3.3-70b-instruct & \checkmark & $5.68_{5.62}^{5.74}$ & $5.69_{5.65}^{5.72}$ & $6.88_{6.84}^{6.92}$ & $6.32_{6.30}^{6.35}$ & $5.75_{5.71}^{5.79}$ & $6.06_{6.02}^{6.10}$ \\
nova-lite-v1 &  & $4.51_{4.45}^{4.57}$ & $6.06_{6.03}^{6.09}$ & $7.38_{7.35}^{7.41}$ & $6.60_{6.58}^{6.63}$ & $5.71_{5.62}^{5.77}$ & $6.05_{6.01}^{6.10}$ \\
\bottomrule
\end{tabular}
}
\end{table}

\newpage
\clearpage

\section{Supplementary Figures}

\begin{figure}[h]
    \centering
    \begin{minipage}[b]{0.45\textwidth}
        \centering
        \includegraphics[width=\textwidth]{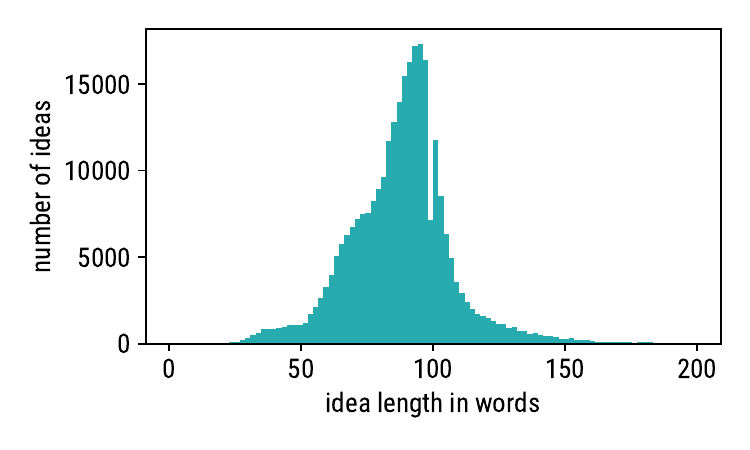}
        \caption{Distribution of idea lengths measured in words}
        \label{fig:idea_length}
    \end{minipage}
    \hfill
    \begin{minipage}[b]{0.45\textwidth}
        \centering
        \includegraphics[width=\textwidth]{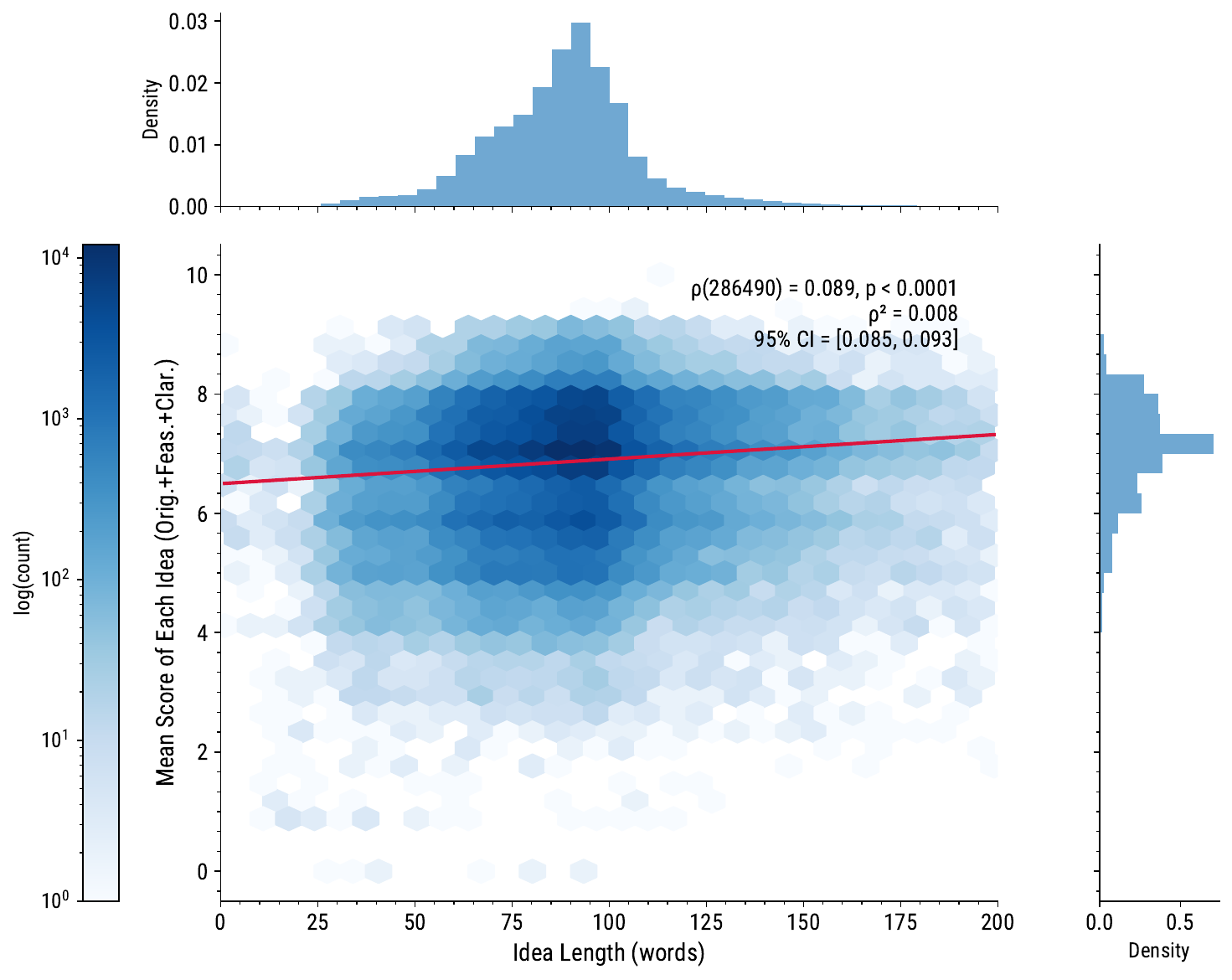}
        \caption{The correlation between idea length and idea score among all models. Spearman’s $\rho$ was used due to deviations from normality.}
        \label{fig:idea_length_corr}
    \end{minipage}
\end{figure}

\begin{figure}[h]
    \centering
    \includegraphics[width=0.7\textwidth]{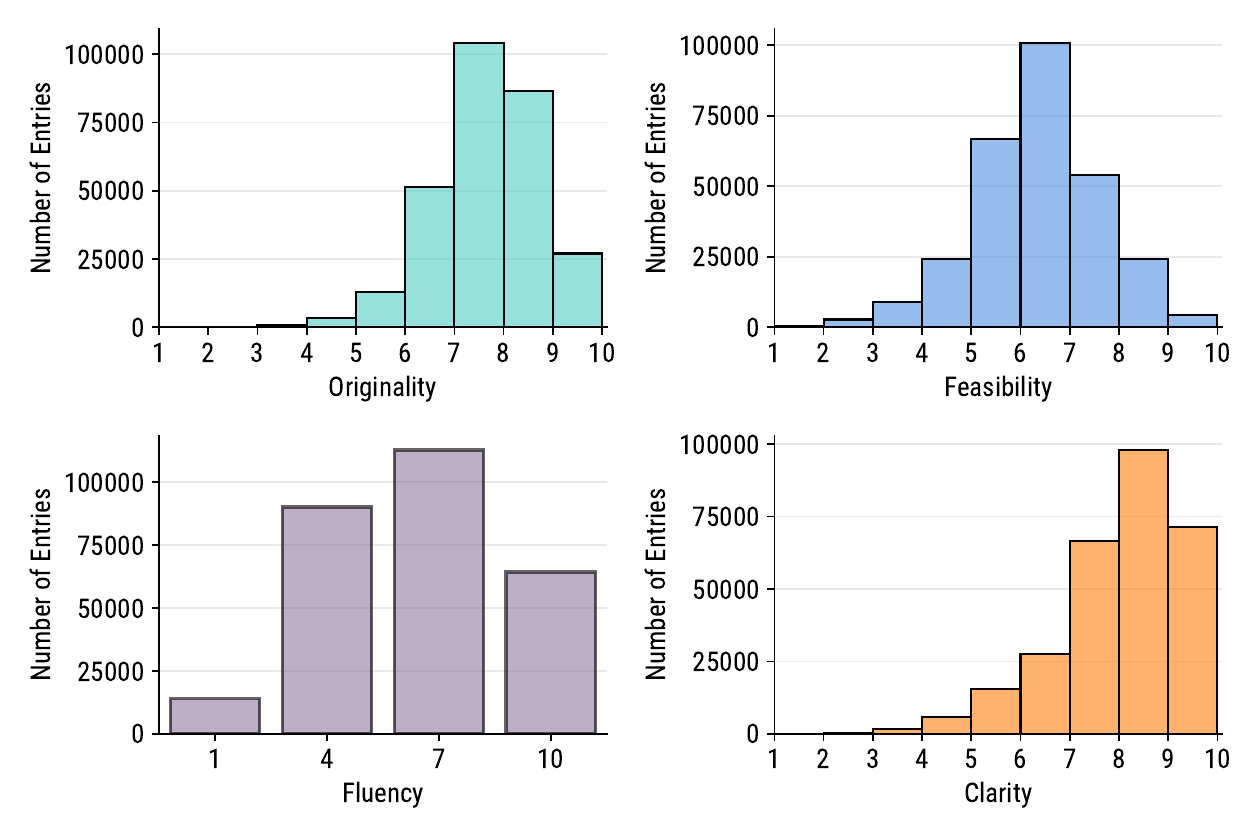} 
    \caption{Distribution of scores for ideas generated by all models}
    \label{fig:score_dist}
\end{figure}

\begin{figure}[h!]
    \centering
    \hspace{-0.5cm} \includegraphics[width=0.9\textwidth]{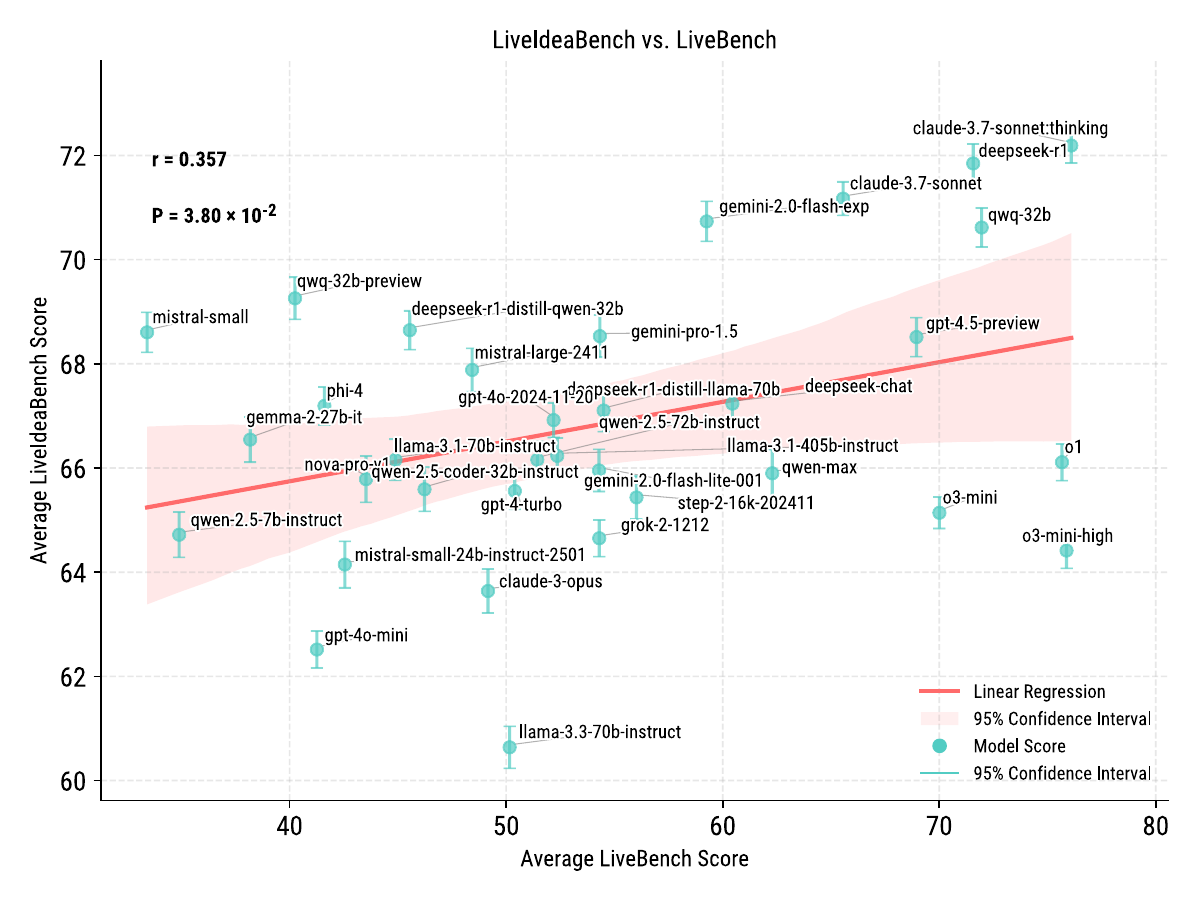} 
    \caption{\textbf{Correlation between LiveIdeaBench and LiveBench Performance.} 
    Regression analysis reveals a weak but statistically significant positive correlation ($r(41) = 0.357, p = 0.038, r^2 = 0.127, \ 95\% \ CI = [0.022, 0.621]$) between average scores on \texttt{LiveIdeaBench} (scientific idea generation) and \texttt{LiveBench} (general intelligence). Points represent models (vertical bars: 95\% CI for \texttt{LiveIdeaBench}; shaded region: 95\% CI for regression). The low correlation coefficient highlights that general intelligence metrics (\texttt{LiveBench}) are not strong predictors of the scientific idea generation capabilities measured by \texttt{LiveIdeaBench}. Pearson’s $r$ was used as data met normality assumptions (Shapiro-Wilk test).}
    \label{fig:reg_livebench}
\end{figure}

\begin{figure}[h!]
    \centering
    \includegraphics[width=1.0\linewidth]{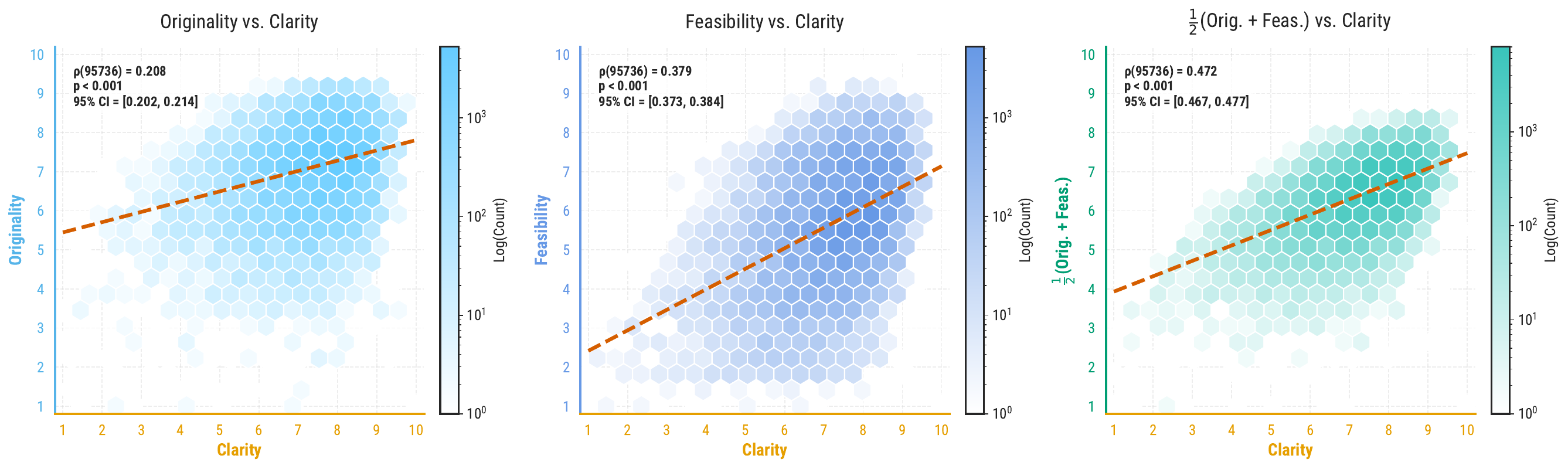} 
    \caption{Relationship between clarity and other core evaluation dimensions across all generated ideas. Hexbins show data density, red dashed lines show linear regressions. Clarity demonstrates positive correlations with originality, feasibility, and their average, indicating clearer ideas are generally rated higher overall. Spearman’s $\rho$ is reported (due to non-normal data distribution).}
    \label{fig:figure3}
\end{figure}

This section provides additional visualizations and statistical analyses supporting the findings discussed in the main text. Specifically, Figure \ref{fig:idea_length} and Figure \ref{fig:idea_length_corr} detail the distribution of generated idea lengths and demonstrate their weak correlation with the overall idea quality scores. To provide a macro-level perspective on model performance, Figure \ref{fig:score_dist} illustrates the overall distribution of evaluation scores across all generated ideas. Furthermore, Figure \ref{fig:reg_livebench} presents the regression analysis comparing models' general intelligence (LiveBench) with their scientific ideation capabilities (LiveIdeaBench), highlighting the distinct nature of these two cognitive domains. Finally, Figure \ref{fig:figure3} explores the internal dynamics of our evaluation metrics, showing that the clarity of an idea positively correlates with its originality, feasibility, and overall average score.

\clearpage
\newpage

\section{Handling Model Refusals}

Evaluating idea generation across diverse scientific keywords inevitably encounters instances where LLMs refuse to respond, often due to safety protocols triggered by potentially sensitive terms (e.g., ``ecotoxicology''). Directly penalizing such refusals could lead to an underestimation of the creative capabilities of models, particularly those with robust safety alignments. To ensure a fairer and more accurate assessment that accounts for these necessary safety features, we implement a specific protocol to handle such refusals.

The protocol involves a two-step process:
\begin{enumerate}
    \item The model is first prompted with the standard idea generation request (see \nameref{sec:prompts}).
    \item The response is automatically checked for common refusal-indicating phrases.
    \item If a refusal is detected, a second attempt is made using a specific \textbf{fallback prompt}. This fallback prompt explicitly frames the task within an academic research context and reassures the model about the benign use of the output, encouraging it to provide a scientific idea despite potential keyword sensitivity.
\end{enumerate}
This approach allows us to better assess the model's underlying idea generation ability, distinguishing between an inability to generate ideas and adherence to safety guidelines.

\begin{figure}[h!]
    \centering
    \includegraphics[width=1.0\textwidth]{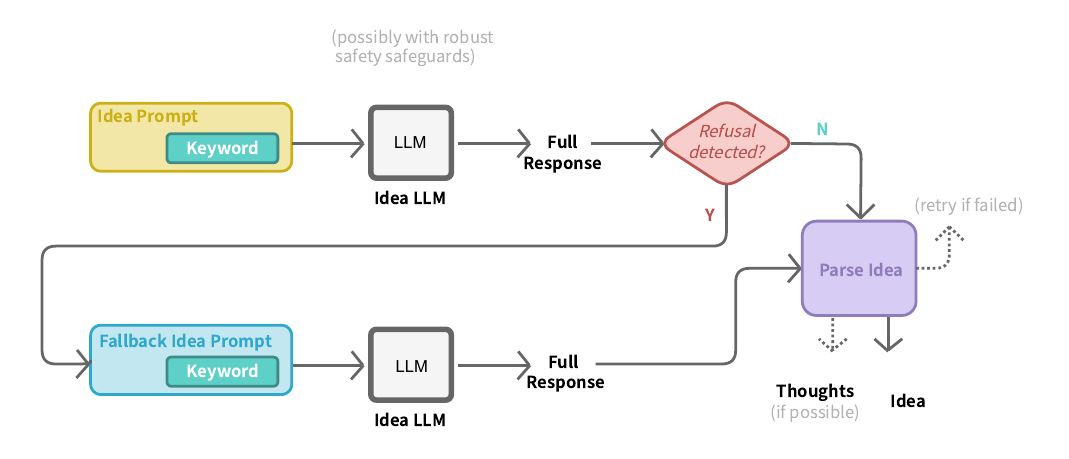}
    \caption{Visualization of the refusal handling protocol.}
    \label{fig:refusal_workflow} 
\end{figure}

Figure \ref{fig:refusal_workflow} provides a visual representation of this refusal handling workflow. The process starts with the initial idea prompt containing a specific keyword. This is fed into the idea LLM. The resulting full response is analyzed to detect potential refusals. If no refusal is detected, the response proceeds to parsing. If a refusal is detected, a fallback prompt is used for a second attempt with the idea LLM. The response from this fallback attempt then moves to the parse idea stage. Finally, the parse idea component extracts the core idea (and potentially thoughts for reasoning models) from the successful response, with an option to retry parsing if the initial attempt fails.

The automatic detection of refusals relies on identifying specific keywords and phrases within the model's initial response. The set of phrases used for this purpose is listed below:

\begin{tcolorbox}[
    colback=red!5!purple!5!white,
    colframe=red!40!purple!70!black,
    title=Detection Phrases for Refusals,
    fontupper=\footnotesize
]
The following phrases trigger the fallback prompt mechanism:
\begin{itemize}
    \item "I'm sorry", "I am sorry", "I apologize"
    \item "As an AI", "As a language model", "As an assistant"
    \item "I cannot", "I can't"
    \item "I am unable to", "I'm unable to"
    \item "I am not able to", "I'm not able to"
\end{itemize}
\end{tcolorbox}

\begin{figure}[h!]
    \centering
    \includegraphics[width=1.0\textwidth]{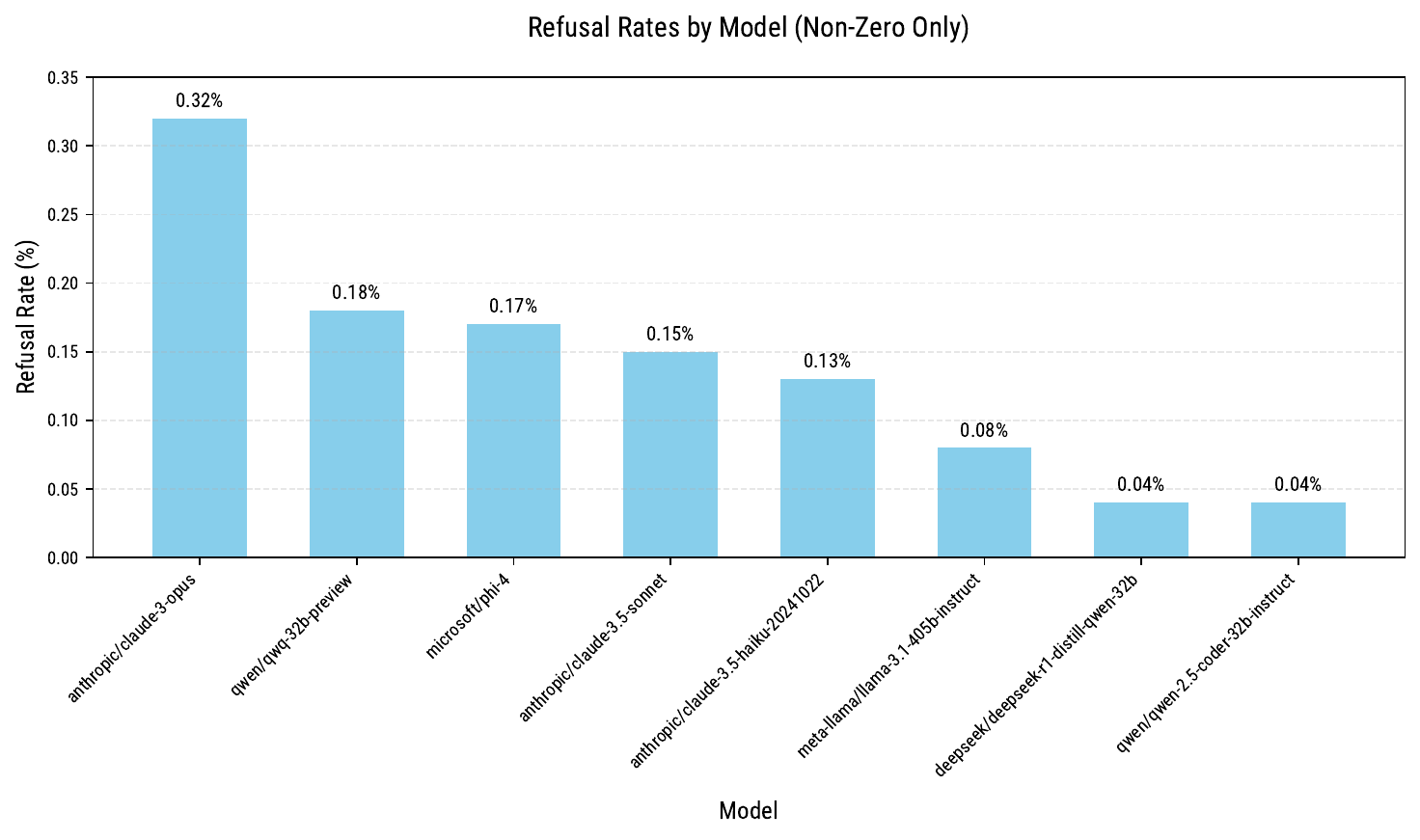}
    \caption{Initial refusal rates per model (before fallback prompt).}
    \label{fig:refusal_rates}
\end{figure}

To provide further context on model behavior regarding potentially sensitive prompts, Figure \ref{fig:refusal_rates} presents the observed refusal rates for each evaluated model across the full set of keywords after the initial prompt (before the fallback mechanism is applied).

\clearpage

\section{CO$_2$ Emission Estimation}

Recognizing the growing importance of assessing the environmental impact of large-scale computational research, we estimated the CO$_2$ emissions associated with running the LLMs in our \texttt{LiveIdeaBench} benchmark. We employed the ``EcoLogits'' tool \cite{ecologits-calculator} (\url{https://huggingface.co/spaces/genai-impact/ecologits-calculator}, accessed on March 30, 2025). This approach estimates emissions based on factors such as total model parameters, estimated activated parameters during inference, and generated token counts, making it more suitable for scenarios involving API calls. Parameter counts for proprietary models were estimated based on publicly available information, which introduces a degree of uncertainty to these specific values. Table \ref{tab:combined_emissions} presents the estimated energy consumption (kWh) and carbon emissions (kgCO$_2$eq.) for each model, broken down by its role as an idea generator and/or a critic, along with the total impact attributable to our evaluation process.

\begin{table}[htbp]
\centering 

\caption{Energy Consumption and Carbon Emissions by Model}
\label{tab:combined_emissions} 

\begin{minipage}{0.7\textwidth}
    \centering 
    \resizebox{\linewidth}{!}{
    
    \begin{tabular}{lrrrrrr}
    \toprule
    \multirow{2}{*}{Model Name} & \multicolumn{2}{c}{As Idea Model} & \multicolumn{2}{c}{As Critic Model} & \multicolumn{2}{c}{Total} \\
    \cmidrule(lr){2-3} \cmidrule(lr){4-5} \cmidrule(lr){6-7}
    & \makecell{Energy\\(kWh)} & \makecell{Emissions\\(kgCO$_2$eq.)} & \makecell{Energy\\(kWh)} & \makecell{Emissions\\(kgCO$_2$eq.)} & \makecell{Energy\\(kWh)} & \makecell{Emissions\\(kgCO$_2$eq.)} \\
    \midrule
    \texttt{claude-3-opus$^*$} & 83.73 & 51.22 & - & - & 83.73 & 51.22 \\
    \texttt{claude-3.5-haiku-20241022$^*$} & 1.52 & 0.93 & - & - & 1.52 & 0.93 \\
    \texttt{claude-3.5-sonnet$^*$} & 12.17 & 7.43 & 226.20 & 138.05 & 238.37 & 145.48 \\
    \texttt{claude-3.7-sonnet$^*$} & 16.54 & 10.10 & - & - & 16.54 & 10.10 \\
    \texttt{claude-3.7-sonnet:thinking$^*$} & 271.66 & 165.79 & 1411.06 & 861.16 & 1682.72 & 1026.95 \\
    \texttt{deepseek-chat} & 15.97 & 9.78 & 184.49 & 113.05 & 200.45 & 122.83 \\
    \texttt{deepseek-r1} & 77.18 & 47.29 & 670.38 & 410.79 & 747.56 & 458.08 \\
    \texttt{deepseek-r1-distill-llama-70b} & 13.18 & 8.03 & - & - & 13.18 & 8.03 \\
    \texttt{deepseek-r1-distill-qwen-32b} & 10.20 & 6.25 & - & - & 10.20 & 6.25 \\
    \texttt{gemini-2.0-flash-exp$^*$} & 2.07 & 1.27 & - & - & 2.07 & 1.27 \\
    \texttt{gemini-2.0-flash-lite-001$^*$} & 1.11 & 0.68 & - & - & 1.11 & 0.68 \\
    \texttt{gemini-2.0-flash-thinking-exp$^{*\dagger}$} & 1.92 & 1.17 & 22.99 & 14.07 & 24.91 & 15.25 \\
    \texttt{gemini-2.0-pro-exp-02-05$^*$} & 14.84 & 9.10 & 168.11 & 103.01 & 182.95 & 112.11 \\
    \texttt{gemini-pro-1.5$^*$} & 9.73 & 5.96 & - & - & 9.73 & 5.96 \\
    \texttt{gemma-2-27b-it} & 1.39 & 0.85 & - & - & 1.39 & 0.85 \\
    \texttt{gpt-4-turbo$^*$} & 1.64 & 1.01 & - & - & 1.64 & 1.01 \\
    \texttt{gpt-4.5-preview$^*$} & 83.52 & 51.09 & 1050.86 & 642.79 & 1134.37 & 693.88 \\
    \texttt{gpt-4o-2024-11-20$^*$} & 15.14 & 9.28 & - & - & 15.14 & 9.28 \\
    \texttt{gpt-4o-mini$^*$} & 0.97 & 0.60 & - & - & 0.97 & 0.60 \\
    \texttt{grok-2-1212$^*$} & 12.22 & 7.49 & - & - & 12.22 & 7.49 \\
    \texttt{llama-3.1-405b-instruct} & 75.62 & 46.07 & - & - & 75.62 & 46.07 \\
    \texttt{llama-3.1-70b-instruct} & 4.15 & 2.53 & - & - & 4.15 & 2.53 \\
    \texttt{llama-3.3-70b-instruct} & 1.64 & 1.00 & - & - & 1.64 & 1.00 \\
    \texttt{mistral-large-2411} & 5.81 & 3.56 & - & - & 5.81 & 3.56 \\
    \texttt{mistral-small} & 1.77 & 1.09 & - & - & 1.77 & 1.09 \\
    \texttt{mistral-small-24b-instruct-2501} & 2.10 & 1.29 & - & - & 2.10 & 1.29 \\
    \texttt{nova-lite-v1$^*$} & 0.84 & 0.52 & - & - & 0.84 & 0.52 \\
    \texttt{nova-pro-v1$^*$} & 2.49 & 1.52 & - & - & 2.49 & 1.52 \\
    \texttt{o1$^{*\dagger}$} & 11.31 & 6.93 & - & - & 11.31 & 6.93 \\
    \texttt{o1-mini$^{*\dagger}$} & 1.06 & 0.65 & - & - & 1.06 & 0.65 \\
    \texttt{o3-mini$^{*\dagger}$} & 10.51 & 6.44 & - & - & 10.51 & 6.44 \\
    \texttt{o3-mini-high$^{*\dagger}$} & 10.95 & 6.71 & 131.59 & 80.63 & 142.54 & 87.34 \\
    \texttt{phi-4} & 1.83 & 1.12 & - & - & 1.83 & 1.12 \\
    \texttt{qwen-2.5-72b-instruct} & 3.62 & 2.22 & - & - & 3.62 & 2.22 \\
    \texttt{qwen-2.5-7b-instruct} & 0.82 & 0.50 & - & - & 0.82 & 0.50 \\
    \texttt{qwen-2.5-coder-32b-instruct} & 1.86 & 1.14 & - & - & 1.86 & 1.14 \\
    \texttt{qwen-max$^*$} & 7.87 & 4.82 & 163.20 & 100.00 & 171.07 & 104.83 \\
    \texttt{qwen2.5-dracarys2-72b} & 3.33 & 2.04 & - & - & 3.33 & 2.04 \\
    \texttt{qwq-32b} & 23.36 & 14.30 & 139.06 & 85.11 & 162.42 & 99.41 \\
    \texttt{qwq-32b-preview} & 26.95 & 16.50 & - & - & 26.95 & 16.50 \\
    \texttt{step-2-16k-202411$^*$} & 14.93 & 9.15 & - & - & 14.93 & 9.15 \\
    \midrule
    Total & 859.52 & 525.41 & 4167.92 & 2548.66 & 5027.45 & 3074.07 \\
    \bottomrule
    \end{tabular}
    } 
    
    \vspace{0.5ex} 
    
    \caption*{\footnotesize{$^*$Non-open models: environmental impact values are rough estimates and may not be accurate.\newline $^\dagger$Estimate may be low as the model’s reasoning is not accessible via API.}}
\end{minipage} 

\end{table}

\clearpage

\section{Human Expert Evaluation}
\label{sec:human_eval} 

To assess the validity and reliability of using LLMs as judges for evaluating scientific ideas within the \texttt{LiveIdeaBench} framework, we conducted a human expert evaluation study. Given the breadth of scientific domains covered by our benchmark, we focused this validation effort on a specific field, Partial Differential Equations (PDEs), where we could readily access domain expertise.

\begin{figure}[h!]
    \centering
    \includegraphics[width=1.0\textwidth]{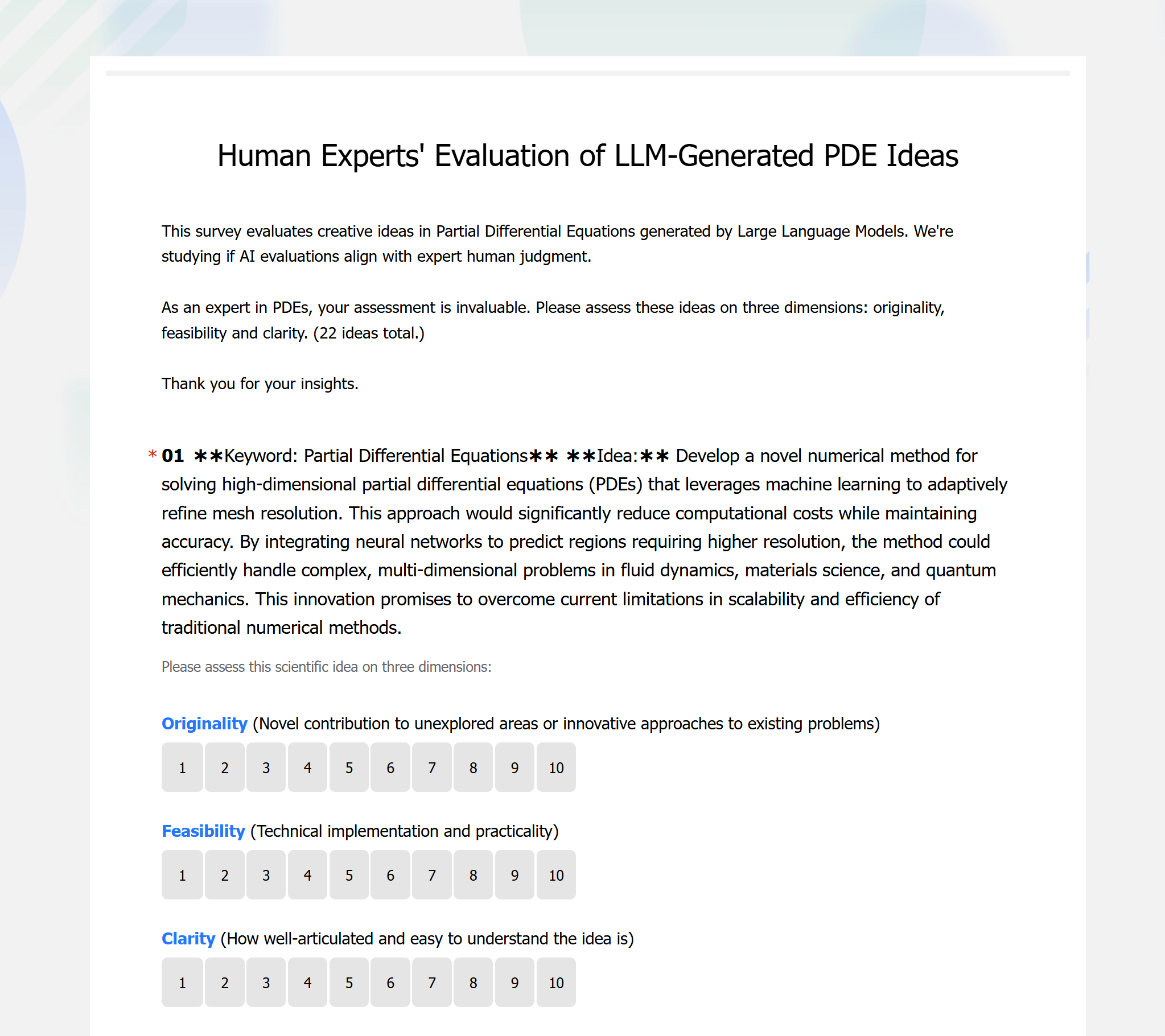}
    \caption{Screenshot of the questionnaire interface presented to human experts for rating LLM-generated ideas related to Partial Differential Equations.}
    \label{fig:questionnaire_view}
\end{figure}

For this validation study, we utilized the set of 22 scientific ideas generated by 11 LLMs (i.e. \texttt{claude-3.7-sonnet:thinking}, \texttt{qwq-32b}, \texttt{claude-3.5-sonnet}, \texttt{gpt-4.5-preview}, \texttt{deepseek-r1-\allowbreak distill-llama-70b}, \texttt{qwen2.5-dracarys2-72b}, \texttt{o1}, \texttt{qwen-2.5-coder-32b-instruct}, \texttt{qwen-2.5-7b-\allowbreak instruct}, \texttt{claude-3.5-haiku-20241022}, and \texttt{nova-lite-v1}, sampled at equal intervals based on rankings (see Table \ref{tab:liveideabench_scores_with_ci}) from the 41 LLMs included in our benchmark, two ideas per model) specifically for the keyword ``partial differential equations''. The full text of these 22 ideas is provided in Table \ref{tab:idea_details}. These ideas were presented to 6 human experts in the PDE field via an online questionnaire. Experts were asked to rate each idea on the same three core dimensions used by our LLM judges (originality, feasibility, and clarity) using an identical 1-10 integer scale. The detailed ratings from both the LLM judge panel (averaged) and the individual human experts for each idea are compiled in Table \ref{tab:scores}. To minimize bias, the 22 generated ideas were presented to the experts in a randomized order, and the identity of the LLM that generated each idea was withheld (blind evaluation). A screenshot illustrating the questionnaire interface presented to the experts is shown in Figure \ref{fig:questionnaire_view}.

Inter-rater reliability (IRR) among the 6 human experts was assessed for each dimension using the Intraclass Correlation Coefficient ($ICC$) based on a two-way random-effects model evaluating absolute agreement for the average rating ($ICC$($A$,$k$)). The analysis revealed good agreement for originality ($ICC = 0.823$) and clarity ($ICC = 0.782$). However, only poor agreement was found for feasibility ($ICC = 0.453$), indicating substantial disagreement among experts on the technical viability and scientific soundness of the generated ideas. This low inter-expert agreement underscores the significant subjectivity and inherent difficulty in evaluating the feasibility of novel scientific ideas based solely on brief textual descriptions, even for domain specialists.

\begin{figure}[h!]
    \centering
    \includegraphics[width=1.0\textwidth]{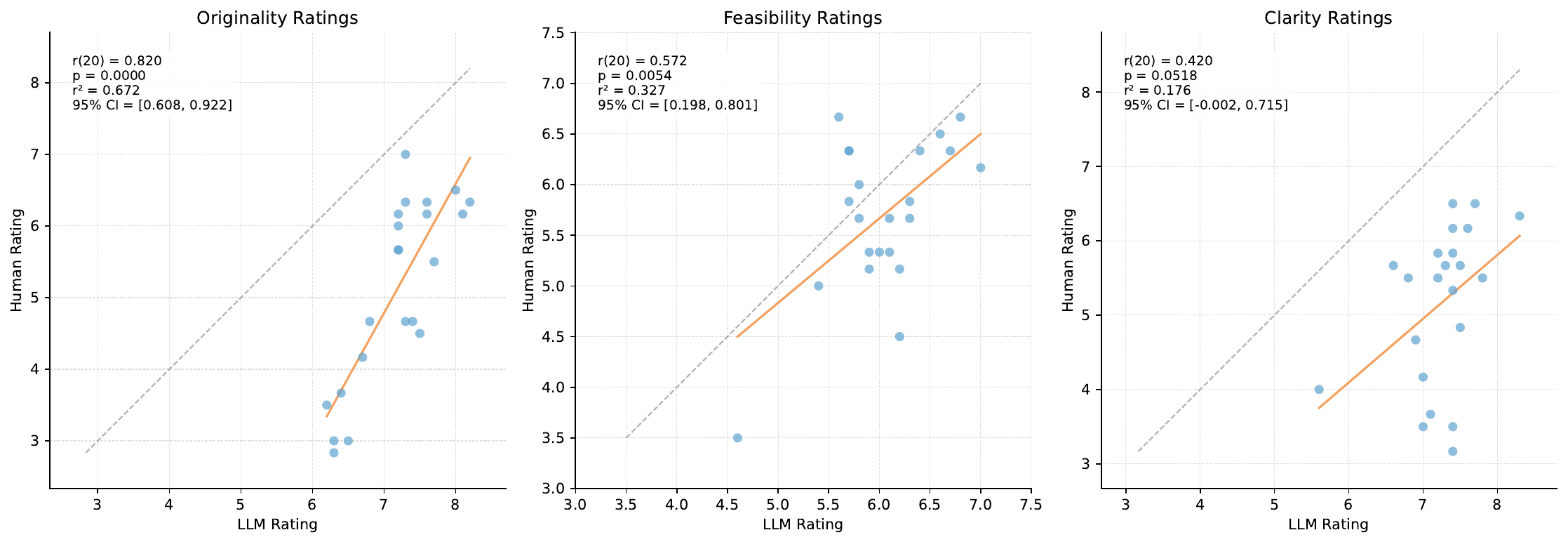}
    \caption{Scatter plot comparing Human Expert (average of 6 experts) and LLM Judge (average of 10 LLM judges) ratings for originality, feasibility, and clarity on 22 PDE-related ideas.}
    \label{fig:scatter_comparison} 
\end{figure}

The subsequent figures in this section present a comparative analysis between the average ratings from the 6 human experts and the average ratings assigned by the LLM judge panel for these 22 PDE-related ideas. Specifically for this validation analysis, to obtain a stable estimate of the LLM panel's collective assessment for comparison against the human expert average, the LLM rating for each idea represents the average score across all eligible critic models in the top-10 judges. This approach differs from the main benchmark's scoring procedure, which uses a random sample of 3 judges per idea, and was chosen here to minimize sampling noise in the validation comparison. Figure \ref{fig:scatter_comparison} visualizes the correlation between these averaged human and averaged LLM ratings for each dimension using scatter plots and reports the Pearson correlation coefficient. The analysis shows statistically significant positive correlations for all dimensions: a strong correlation for originality, a moderate correlation for feasibility, and a weaker but still significant correlation for clarity (The superscripts denote the level of statistical significance based on the p-value. The p-value represents the probability of observing the given correlation (or a stronger one) if there were actually no correlation between the ratings in the broader population. Lower p-values indicate stronger evidence against this possibility (i.e., stronger evidence that the correlation is real).

The strong human IRR for originality ($ICC = 0.823$) provides a solid foundation for the observed high human-LLM correlation, suggesting that the LLM judges effectively captured expert consensus on novelty within this specific domain. Similarly, the good human IRR for clarity ($ICC = 0.782$) supports the finding of a significant, albeit weaker, human-LLM correlation. Critically, the poor human IRR for feasibility ($ICC = 0.453$) complicates the interpretation of the moderate human-LLM correlation. The lack of consensus among human experts on feasibility implies that the average human rating for this dimension is inherently noisy and less reliable as a benchmark. Therefore, the moderate correlation observed likely reflects a combination of the LLMs' capabilities and the significant variability within the human expert judgments themselves. This underscores the challenge of reliably evaluating scientific feasibility based solely on brief idea descriptions, a limitation potentially affecting both human and LLM assessments.

Overall, the validation study indicates reasonable alignment between human expert judgment and the automated LLM evaluations, lending empirical support to the LLM-as-a-judge methodology employed in \texttt{LiveIdeaBench}, particularly for assessing originality and clarity. However, the findings regarding feasibility highlight the inherent challenges and necessitate caution when interpreting the benchmark's feasibility scores.

\clearpage

{\color{black}
\begin{longtable}{>{\scriptsize\ttfamily}r>{\scriptsize\ttfamily\RaggedRight\arraybackslash}p{0.25\linewidth}>{\scriptsize\ttfamily\RaggedRight\arraybackslash}p{0.65\linewidth}}

\caption{\textbf{LLM-generated scientific ideas on Partial Differential Equations used in the validation study.} The table lists the idea model and the 22 generated scientific idea for the related to ``partial differential equations'' evaluated in the human expert validation analysis} \label{tab:idea_details} \\

\toprule
\textbf{ID} & \textbf{Idea Model} & \textbf{Idea} \\
\midrule
\endfirsthead

\caption[]{Continued Generated Ideas} \\
\toprule
\textbf{ID} & \textbf{Idea Model} & \textbf{Idea} \\
\midrule
\endhead

\midrule
\multicolumn{3}{r}{\itshape (Continued on next page)} \\
\endfoot

\bottomrule
\endlastfoot

1 & nova-lite-v1 & **Keyword: Partial Differential Equations** **Idea:** Develop a novel numerical method for solving high-dimensional partial differential equations (PDEs) that leverages machine learning to adaptively refine mesh resolution. This approach would significantly reduce computational costs while maintaining accuracy. By integrating neural networks to predict regions requiring higher resolution, the method could efficiently handle complex, multi-dimensional problems in fluid dynamics, materials science, and quantum mechanics. This innovation promises to overcome current limitations in scalability and efficiency of traditional numerical methods. \\
2 & nova-lite-v1 & **Keyword: Partial Differential Equations** **Idea:** Develop a novel numerical method for solving high-dimensional partial differential equations (PDEs) that leverages machine learning to adaptively refine mesh resolution. This approach would significantly reduce computational costs while maintaining accuracy. By integrating neural networks to predict regions requiring higher resolution, the method would be particularly useful for complex systems in fluid dynamics and materials science. This innovative technique promises to enhance both efficiency and precision in solving PDEs. \\
3 & claude-3.5-haiku-20241022 & Here's a novel approach for solving complex partial differential equations (PDEs) using quantum-inspired neural networks: Background: Traditional numerical methods struggle with high-dimensional, nonlinear PDEs across physics and engineering domains. Idea: Develop a hybrid quantum-classical neural architecture that leverages quantum probability amplitude mappings to approximate PDE solutions. By encoding boundary conditions as quantum state constraints and utilizing quantum-enhanced feature extraction, the network can dramatically reduce computational complexity and improve convergence for systems like turbulent fluid dynamics, quantum mechanics wave equations, and climate modeling partial differential systems. The approach combines machine learning's adaptability with quantum computational principles. \\
4 & claude-3.5-haiku-20241022 & Here's a submission focused on partial differential equations (PDEs): Novel PDE Machine Learning Predictive Framework Background: Traditional PDE solving methods struggle with complex, nonlinear systems across physics and engineering. Proposed Idea: Develop a hybrid neural network architecture combining physics-informed neural networks (PINNs) with adaptive mesh refinement techniques. This approach would dynamically adjust computational resolution based on solution gradient complexity, enabling more accurate and computationally efficient numerical solutions for multiscale PDEs in fluid dynamics, quantum mechanics, and materials science. The framework would integrate residual learning, constraint enforcement, and intelligent mesh adaptation to dramatically improve predictive modeling of high-dimensional, nonlinear systems. \\
5 & claude-3.5-sonnet & Background: Partial differential equations (PDEs) are crucial in modeling physical phenomena but solving them numerically can be computationally intensive and sometimes unstable. Idea: Develop a hybrid solver that combines traditional numerical methods with machine learning to adaptively switch between different solution strategies. The system would analyze the local behavior of the PDE in real-time and automatically select the most efficient and stable method (finite elements, spectral methods, or neural network approximations) for that specific region. This would significantly reduce computational costs while maintaining accuracy, particularly for PDEs with mixed-type behavior or discontinuities. \\
6 & claude-3.5-sonnet & Background: Partial differential equations (PDEs) are crucial in modeling physical phenomena but solving them numerically can be computationally intensive, especially for complex geometries. Idea: Develop a hybrid solver that combines machine learning with traditional numerical methods by training neural networks to identify regions where high-resolution calculations are necessary versus where simplified approximations suffice. The neural network would adaptively adjust the mesh resolution and computational approach in real-time based on the solution's local complexity, significantly reducing computational costs while maintaining accuracy in critical regions where precise PDE solutions are essential. \\
7 & claude-3.7-sonnet & \# Multi-Resolution Graph Neural Operators for Heterogeneous PDEs **Background:** PDEs in heterogeneous media with multiple spatial scales remain computationally challenging despite decades of research. **Idea:** We propose adaptive graph neural operators that learn multi-resolution representations of PDE solutions. The approach automatically identifies regions requiring different resolution levels, constructing hierarchical graph representations that efficiently capture both fine-scale features and global behaviors. This enables fast, accurate predictions for complex multi-scale systems like subsurface flow and biomechanics while reducing computational costs by orders of magnitude compared to traditional solvers. The method preserves important physical invariants through specialized graph convolutions. \\
8 & claude-3.7-sonnet & \# Wavelet-Guided Adaptive Physics-Informed Neural Networks for Multiscale PDEs Physics-informed neural networks (PINNs) have emerged as powerful PDE solvers, but struggle with multiscale phenomena and sharp transitions that characterize many real-world problems. I propose Wavelet-Guided Adaptive PINNs that dynamically allocate computational resources based on local solution complexity. By integrating wavelet analysis to identify regions requiring higher resolution, the network automatically adjusts its architecture—increasing depth or width where needed. This approach significantly improves efficiency and accuracy for problems with boundary layers and multiscale features in fluid dynamics, materials science, and climate modeling, addressing a fundamental limitation in existing neural PDE solvers. \\
9 & deepseek-r1-distill-llama-70b & **Idea:** Develop a deep learning framework to solve partial differential equations (PDEs) using neural networks that eliminate the need for traditional mesh generation. This approach would train neural networks to approximate PDE solutions directly from input parameters (e.g., space, time, coefficients), reducing computational overhead and enabling real-time solutions for complex, high-dimensional problems. **Background:** PDEs are fundamental in modeling physical phenomena, but numerical solutions often require computationally expensive mesh discretization. This innovation would democratize PDE solving by enabling fast, mesh-free solutions on standard computing hardware. \\
10 & deepseek-r1-distill-llama-70b & **Idea:** Developing a hybrid framework combining physics-informed neural networks (PINNs) with adaptive mesh refinement (AMR) to solve partial differential equations (PDEs). This approach integrates the flexibility of neural networks in capturing complex solutions with AMR's ability to dynamically allocate computational resources to regions of interest. The hybrid method will improve accuracy and efficiency for multiscale PDEs, such as those in fluid dynamics or materials science, by automatically refining meshes where neural network errors are high. This innovative approach addresses the limitations of traditional finite element or finite difference methods, enabling more precise and computationally efficient simulations. \\
11 & gpt-4.5-preview & Background: Approximating solutions of nonlinear partial differential equations (PDEs) commonly relies on computationally expensive numerical methods. Current machine learning approaches, like neural PDE solvers, provide speedup but lack accuracy guarantees of traditional numerical schemes. Idea: A hybrid method integrating neural operator networks with adaptive finite element error correction is proposed. The neural component rapidly predicts solutions, while an iterative residual-based finite-element refinement ensures accuracy within specified tolerances. This architecture leverages machine learning speed with traditional PDE solver precision, significantly reducing computational cost while providing mathematical reliability in solving nonlinear PDE problems relevant across physics and engineering disciplines. \\
12 & gpt-4.5-preview & Background: Solving high-dimensional partial differential equations (PDEs), central to modeling complex systems such as financial markets, fluid dynamics, and quantum mechanics, suffers from prohibitively high computational demands (curse of dimensionality). Idea: Develop a hybrid deep-learning spectral decomposition method that approximates PDE solutions by dynamically partitioning high-dimensional spaces into data-driven low-dimensional spectral subspaces. Combining deep neural networks and adaptive spectral decomposition, this approach significantly reduces computational cost and storage complexity, enables real-time solutions to previously intractable PDE problems, and provides interpretable spectral insights into underlying physical phenomena while maintaining predictive accuracy and scalability. \\
13 & o1 & Partial differential equations (PDEs) underlie many scientific fields. IDEA: Use a microservices-based domain decomposition for PDE solvers, enabling parallel subproblem computations. Each subdomain is handled by specialized microservices that adapt mesh resolution in real time. This fosters scalable high-performance computing (HPC) performance with improved load balancing and quicker solutions. Feasible with containerization technologies, it addresses fluid dynamics, structural analysis, and beyond. \\
14 & o1 & We propose a novel partial differential equations-based model for real-time wildfire spread forecasting. The approach couples Navier-Stokes partial differential equations for wind flow with a reaction-diffusion process to capture fire propagation. By combining GPU-accelerated domain decomposition with adaptive time-stepping, it can account for heterogeneous terrain and evolving fuel conditions, enabling predictive wildfire simulations. This method could guide resource allocation and evacuation planning. \\
15 & qwen-2.5-7b-instruct & Partial differential equations (PDEs) are essential in modeling complex systems across various disciplines. My proposed work focuses on developing a novel numerical method to solve high-dimensional PDEs more efficiently. Specifically, we will leverage machine learning techniques to approximate the solution, reducing computational complexity through dimensionality reduction and adaptive sampling. This approach addresses the curse of dimensionality, a significant challenge in PDEs for high-dimensional problems, enabling more accurate and faster simulations in fields like finance, physics, and engineering. \\
16 & qwen-2.5-7b-instruct & Partial differential equations (PDEs) are fundamental in modeling complex systems in physics, engineering, and finance. My idea is to develop a novel numerical method that integrates machine learning with traditional PDE solvers. This hybrid approach aims to significantly reduce computational time and enhance accuracy in solving high-dimensional PDEs, which are currently infeasible with existing methods. This innovation addresses a critical gap in solving complex, real-world problems involving multiple interacting variables. \\
17 & qwen-2.5-coder-32b-instruct & Partial differential equations (PDEs) are fundamental in modeling complex systems across physics, engineering, and biology. They describe phenomena involving space and time dependencies but can be challenging to solve analytically. My idea is to develop a novel, adaptive machine learning algorithm tailored for solving high-dimensional PDEs with varying boundary conditions. This approach will leverage neural networks to approximate solutions efficiently, addressing the curse of dimensionality and enabling real-time simulations. This innovation would significantly enhance predictive modeling capabilities in fields like fluid dynamics and climate science. \\
18 & qwen-2.5-coder-32b-instruct & Partial differential equations (PDEs) are fundamental in modeling complex systems in physics, engineering, and biology. A novel idea is to develop a machine learning framework that automatically learns to solve high-dimensional PDEs using neural networks, addressing the "curse of dimensionality." This approach, combining deep learning with traditional numerical methods, could lead to more efficient and accurate solutions for real-world problems, such as fluid dynamics and climate modeling. The feasibility lies in recent advancements in neural network architectures, and the approach would significantly enhance computational capabilities in solving complex PDEs. \\
19 & qwq-32b & **Background and Idea:** High-dimensional PDEs hinder progress in quantum physics and finance due to the curse of dimensionality. Propose a neural-tensor hybrid framework where physics-aware neural networks parametrize solutions as low-rank tensor networks optimized via PDE-constrained loss functions. This combines neural adaptability for nonlinear dynamics with tensor compression for scalability, enabling high-accuracy solutions in \textgreater{}10 dimensions. Feasible via auto-differentiation tools and tensor libraries, this approach addresses intractable computations while ensuring physical consistency, driving breakthroughs in many-body systems and stochastic modeling. (Word count: 94) **Why It Works:** - **Originality:** Blends neural networks with low-rank tensor decompositions for PDEs, addressing a longstanding scalability issue uniquely. - **Feasibility:** Leverages existing ML frameworks (e.g., TensorFlow/PyTorch) and tensor libraries (e.g., TensorLy), requiring no exotic hardware. - **Impact:** Enables solutions to critical high-dimensional PDEs in fields like quantum chemistry or financial risk modeling, bypassing computational bottlenecks. - **Clarity:** Articulates a clear, actionable method (physics-informed networks × tensor compression). \\
20 & qwq-32b &  **Background:** Partial differential equations (PDEs) are central to modeling physical systems but struggle with computational inefficiency and scalability in high dimensions. Existing methods often trade accuracy for speed or require prohibitive resources.  **Idea:** Develop an AI-driven adaptive PDE solver integrating physics-informed neural networks (PINNs) with error-gradient-guided mesh adaptation. The solver dynamically refines computational grids in regions of high solution variability, trained in tandem with PINNs using gradient-based optimization. This hybrid approach slashes computational costs for high-dimensional problems (e.g., climate modeling, quantum chemistry) while preserving accuracy, addressing a critical bottleneck in scientific simulations.  (100 words) \\
21 & qwen2.5-dracarys2-72b & **Background**: Partial differential equations (PDEs) are fundamental in modeling physical phenomena but often pose significant computational challenges. **Idea**: Develop a hybrid method combining deep learning with traditional numerical techniques to solve high-dimensional PDEs more efficiently. This approach leverages neural networks to approximate solutions in regions where data is sparse, while using classical methods like finite differences or spectral methods where they excel. The integration ensures accuracy and stability, potentially revolutionizing fields such as fluid dynamics, quantum mechanics, and financial modeling by reducing computational time and resource requirements. \\
22 & qwen2.5-dracarys2-72b & **Background:** Partial differential equations (PDEs) are fundamental in modeling complex systems across physics, engineering, and finance. However, solving PDEs for real-world applications often requires significant computational resources and time. **Idea:** Develop a machine learning framework that integrates with existing numerical solvers to predict solutions of PDEs more efficiently. By training neural networks on a diverse dataset of PDE solutions, this hybrid approach can reduce computation time while maintaining accuracy. This method could be particularly useful for real-time simulations and scenarios where rapid solution updates are crucial, such as climate modeling and financial market analysis. \\
\end{longtable}
}


{\color{black}
\begin{longtable}{>{\small}r|>{\small\ttfamily}r>{\small\ttfamily}r>{\small\ttfamily}r|>{\small\ttfamily}r>{\small\ttfamily}r>{\small\ttfamily}r|>{\small\ttfamily}r>{\small\ttfamily}r>{\small\ttfamily}r|>{\small\ttfamily}r>{\small\ttfamily}r>{\small\ttfamily}r|>{\small\ttfamily}r>{\small\ttfamily}r>{\small\ttfamily}r|>{\small\ttfamily}r>{\small\ttfamily}r>{\small\ttfamily}r|>{\small\ttfamily}r>{\small\ttfamily}r>{\small\ttfamily}r}

\caption{\textbf{Comparison of LLM judge and human expert ratings for PDE-related scientific ideas.} Averaged ratings from the LLM judge panel and individual ratings from 6 human experts are presented for originality (O), feasibility (F), and clarity (C) scores assigned to 22 scientific ideas generated for the keyword ``partial differential equations''.} \label{tab:scores} \\

\toprule
\multicolumn{1}{c}{{\small ID}} & \multicolumn{3}{c|}{{\small LLMs (Avg.)}} & \multicolumn{3}{c|}{{\small Expert 1}} & \multicolumn{3}{c|}{{\small Expert 2}} & \multicolumn{3}{c|}{{\small Expert 3}} & \multicolumn{3}{c|}{{\small Expert 4}} & \multicolumn{3}{c|}{{\small Expert 5}} & \multicolumn{3}{c}{{\small Expert 6}} \\
\cmidrule(lr){2-4} \cmidrule(lr){5-7} \cmidrule(lr){8-10} \cmidrule(lr){11-13} \cmidrule(lr){14-16} \cmidrule(lr){17-19} \cmidrule(lr){20-22}
 & \multicolumn{1}{c}{{\small O}} & \multicolumn{1}{c}{{\small F}} & \multicolumn{1}{c}{{\small C}} & \multicolumn{1}{c}{{\small O}} & \multicolumn{1}{c}{{\small F}} & \multicolumn{1}{c}{{\small C}} & \multicolumn{1}{c}{{\small O}} & \multicolumn{1}{c}{{\small F}} & \multicolumn{1}{c}{{\small C}} & \multicolumn{1}{c}{{\small O}} & \multicolumn{1}{c}{{\small F}} & \multicolumn{1}{c}{{\small C}} & \multicolumn{1}{c}{{\small O}} & \multicolumn{1}{c}{{\small F}} & \multicolumn{1}{c}{{\small C}} & \multicolumn{1}{c}{{\small O}} & \multicolumn{1}{c}{{\small F}} & \multicolumn{1}{c}{{\small C}} & \multicolumn{1}{c}{{\small O}} & \multicolumn{1}{c}{{\small F}} & \multicolumn{1}{c}{{\small C}} \\
\midrule
\endfirsthead

\caption[]{Continued Scores} \\
\toprule
\multicolumn{1}{c}{{\small\ttfamily ID}} & \multicolumn{3}{c|}{{\small\ttfamily LLM Judge}} & \multicolumn{3}{c|}{{\small\ttfamily Expert 1}} & \multicolumn{3}{c|}{{\small\ttfamily Expert 2}} & \multicolumn{3}{c|}{{\small\ttfamily Expert 3}} & \multicolumn{3}{c|}{{\small\ttfamily Expert 4}} & \multicolumn{3}{c|}{{\small\ttfamily Expert 5}} & \multicolumn{3}{c}{{\small\ttfamily Expert 6}} \\
\cmidrule(lr){2-4} \cmidrule(lr){5-7} \cmidrule(lr){8-10} \cmidrule(lr){11-13} \cmidrule(lr){14-16} \cmidrule(lr){17-19} \cmidrule(lr){20-22}
 & \multicolumn{1}{c}{{\small\ttfamily O}} & \multicolumn{1}{c}{{\small\ttfamily F}} & \multicolumn{1}{c}{{\small\ttfamily C}} & \multicolumn{1}{c}{{\small\ttfamily O}} & \multicolumn{1}{c}{{\small\ttfamily F}} & \multicolumn{1}{c}{{\small\ttfamily C}} & \multicolumn{1}{c}{{\small\ttfamily O}} & \multicolumn{1}{c}{{\small\ttfamily F}} & \multicolumn{1}{c}{{\small\ttfamily C}} & \multicolumn{1}{c}{{\small\ttfamily O}} & \multicolumn{1}{c}{{\small\ttfamily F}} & \multicolumn{1}{c}{{\small\ttfamily C}} & \multicolumn{1}{c}{{\small\ttfamily O}} & \multicolumn{1}{c}{{\small\ttfamily F}} & \multicolumn{1}{c}{{\small\ttfamily C}} & \multicolumn{1}{c}{{\small\ttfamily O}} & \multicolumn{1}{c}{{\small\ttfamily F}} & \multicolumn{1}{c}{{\small\ttfamily C}} & \multicolumn{1}{c}{{\small\ttfamily O}} & \multicolumn{1}{c}{{\small\ttfamily F}} & \multicolumn{1}{c}{{\small\ttfamily C}} \\
\midrule
\endhead

\midrule
\multicolumn{22}{r}{\itshape (Continued on next page)} \\
\endfoot

\bottomrule
\endlastfoot

1 & 7.4 & 5.8 & 7.8 & 3 & 8 & 10 & 7 & 4 & 7 & 5 & 6 & 7 & 4 & 4 & 1 & 6 & 7 & 6 & 3 & 5 & 2 \\
2 & 7.3 & 6.1 & 7.3 & 7 & 6 & 7 & 7 & 4 & 7 & 6 & 5 & 5 & 6 & 7 & 7 & 8 & 7 & 5 & 4 & 5 & 3 \\
3 & 7.3 & 4.6 & 5.6 & 8 & 5 & 5 & 7 & 3 & 4 & 6 & 3 & 4 & 7 & 3 & 4 & 8 & 4 & 5 & 6 & 3 & 2 \\
4 & 7.2 & 6.4 & 6.6 & 9 & 7 & 8 & 7 & 9 & 7 & 5 & 5 & 6 & 4 & 5 & 4 & 6 & 7 & 5 & 3 & 5 & 4 \\
5 & 7.5 & 5.4 & 7.4 & 7 & 2 & 10 & 5 & 5 & 8 & 3 & 4 & 5 & 3 & 6 & 6 & 5 & 7 & 5 & 4 & 6 & 5 \\
6 & 7.3 & 5.6 & 7.5 & 7 & 6 & 7 & 4 & 8 & 7 & 5 & 6 & 6 & 3 & 7 & 4 & 6 & 8 & 6 & 3 & 5 & 4 \\
7 & 8.0 & 6.2 & 7.2 & 8 & 2 & 8 & 6 & 6 & 7 & 6 & 4 & 6 & 6 & 3 & 5 & 8 & 7 & 7 & 5 & 5 & 2 \\
8 & 8.1 & 6.2 & 7.6 & 9 & 3 & 8 & 7 & 4 & 4 & 6 & 6 & 7 & 5 & 5 & 6 & 7 & 8 & 7 & 3 & 5 & 5 \\
9 & 6.7 & 5.7 & 7.5 & 6 & 8 & 7 & 3 & 7 & 5 & 4 & 4 & 6 & 4 & 4 & 4 & 5 & 7 & 5 & 3 & 5 & 2 \\
10 & 7.6 & 6.6 & 8.3 & 7 & 6 & 7 & 8 & 8 & 7 & 6 & 6 & 6 & 4 & 6 & 6 & 8 & 7 & 6 & 5 & 6 & 6 \\
11 & 7.6 & 6.8 & 7.7 & 9 & 8 & 8 & 6 & 4 & 7 & 7 & 7 & 7 & 3 & 8 & 5 & 8 & 8 & 6 & 4 & 5 & 6 \\
12 & 7.7 & 5.9 & 6.9 & 7 & 7 & 7 & 4 & 4 & 3 & 6 & 4 & 5 & 6 & 4 & 4 & 6 & 7 & 5 & 4 & 6 & 4 \\
13 & 7.2 & 5.7 & 7.2 & 8 & 6 & 7 & 3 & 9 & 7 & 6 & 5 & 6 & 4 & 5 & 4 & 8 & 7 & 6 & 7 & 6 & 3 \\
14 & 7.2 & 5.8 & 7.4 & 8 & 6 & 6 & 4 & 8 & 8 & 7 & 6 & 7 & 7 & 7 & 5 & 6 & 6 & 7 & 5 & 3 & 4 \\
15 & 6.5 & 6.3 & 7.1 & 2 & 7 & 5 & 4 & 8 & 3 & 3 & 4 & 3 & 3 & 3 & 3 & 3 & 8 & 6 & 3 & 5 & 2 \\
16 & 6.3 & 6.1 & 7.4 & 3 & 7 & 8 & 4 & 5 & 3 & 4 & 4 & 3 & 1 & 6 & 3 & 3 & 6 & 2 & 3 & 4 & 2 \\
17 & 6.2 & 5.9 & 7.0 & 2 & 7 & 5 & 4 & 4 & 4 & 3 & 3 & 4 & 3 & 6 & 3 & 6 & 7 & 3 & 3 & 4 & 2 \\
18 & 6.3 & 6.3 & 7.4 & 1 & 8 & 4 & 3 & 4 & 5 & 3 & 4 & 3 & 4 & 5 & 1 & 3 & 8 & 4 & 3 & 5 & 2 \\
19 & 8.2 & 7.0 & 7.4 & 8 & 6 & 7 & 9 & 8 & 7 & 7 & 7 & 7 & 4 & 6 & 4 & 5 & 4 & 3 & 5 & 6 & 7 \\
20 & 7.2 & 5.7 & 7.4 & 8 & 7 & 7 & 7 & 8 & 7 & 5 & 6 & 6 & 5 & 5 & 4 & 5 & 7 & 5 & 4 & 5 & 3 \\
21 & 6.8 & 6.0 & 6.8 & 7 & 4 & 7 & 4 & 8 & 8 & 5 & 4 & 6 & 3 & 6 & 5 & 6 & 5 & 5 & 3 & 5 & 2 \\
22 & 6.4 & 6.7 & 7.0 & 2 & 7 & 4 & 8 & 7 & 7 & 3 & 4 & 3 & 3 & 7 & 3 & 3 & 8 & 5 & 3 & 5 & 3 \\
\end{longtable}
}


\clearpage

\section{Reasoning Length vs. Idea Quality Metrics}
\label{sec:reasoning_length}

The main text established minimal correlation between the length of the final generated idea and its quality. However, some models utilize explicit reasoning steps (e.g., ``\texttt{thinking}'' variants), resulting in longer full responses (thoughts and final idea). This section examines whether this full response length correlates with the final extracted idea's quality for these models.

The following hexbin plots visualize this relationship for selected reasoning-centric models. Each plot displays the density of ideas based on the full response length versus the idea quality score.

\begin{figure}[h!]
    \centering
    \includegraphics[width=0.6\textwidth]{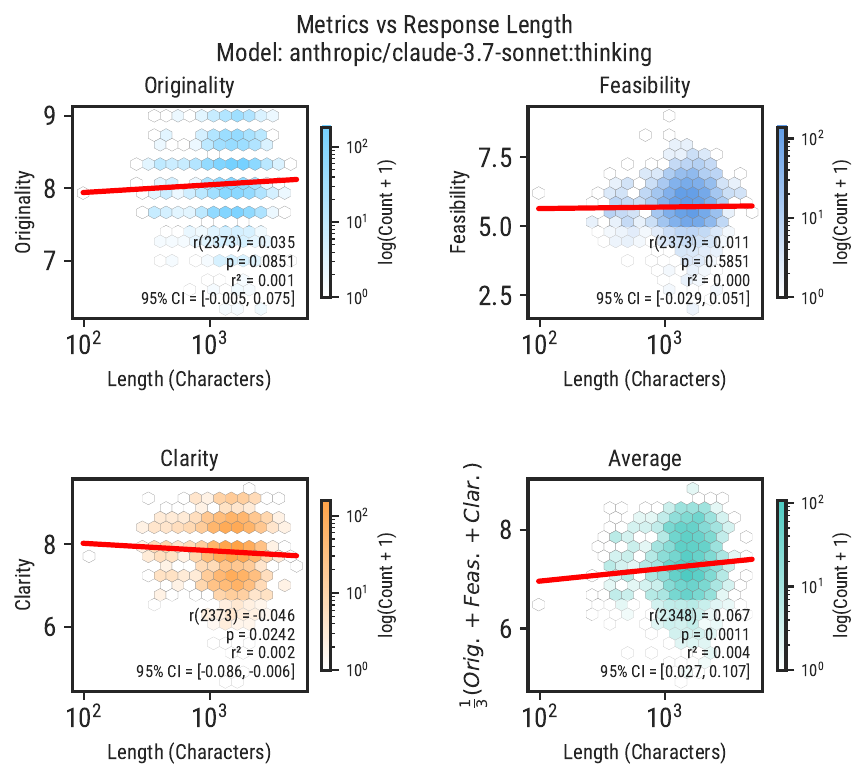}
    \caption{\textbf{Relationship between Full Response Length and Idea Quality for \texttt{claude-3.7-sonnet:thinking}}. Hexbin plot showing the density distribution of generated ideas based on the full response length (characters, x-axis) versus the idea quality score (y-axis). The color intensity represents the number of ideas in each bin.}
    \label{fig:hexbin_claude_thinking}
\end{figure}

\begin{figure}[h!]
    \centering
    \includegraphics[width=0.6\textwidth]{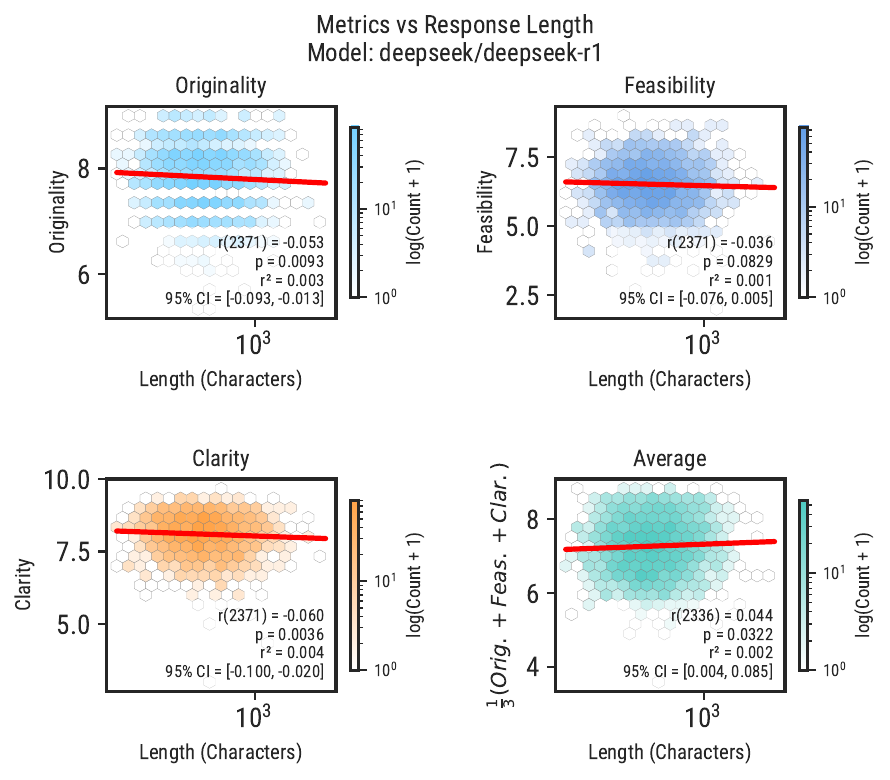}
    \caption{\textbf{Relationship between Full Response Length and Idea Quality for \texttt{deepseek-r1}}. Hexbin plot showing the density distribution of generated ideas based on the full response length (characters, x-axis) versus the idea quality score (y-axis). The color intensity represents the number of ideas in each bin.}
    \label{fig:hexbin_deepseek_r1}
\end{figure}

\begin{figure}[h!]
    \centering
    \includegraphics[width=0.6\textwidth]{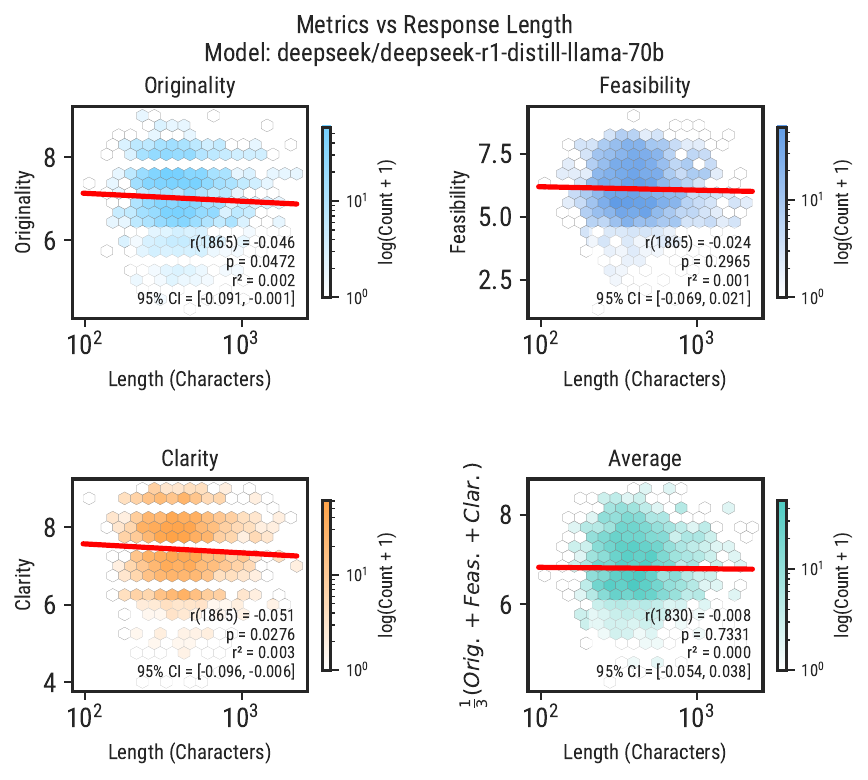}
    \caption{\textbf{Relationship between Full Response Length and Idea Quality for \texttt{deepseek-r1-distill-llama-70b}}. Hexbin plot showing the density distribution of generated ideas based on the full response length (characters, x-axis) versus the idea quality score (y-axis). The color intensity represents the number of ideas in each bin.}
    \label{fig:hexbin_deepseek_distill_llama}
\end{figure}

\begin{figure}[h!]
    \centering
    \includegraphics[width=0.6\textwidth]{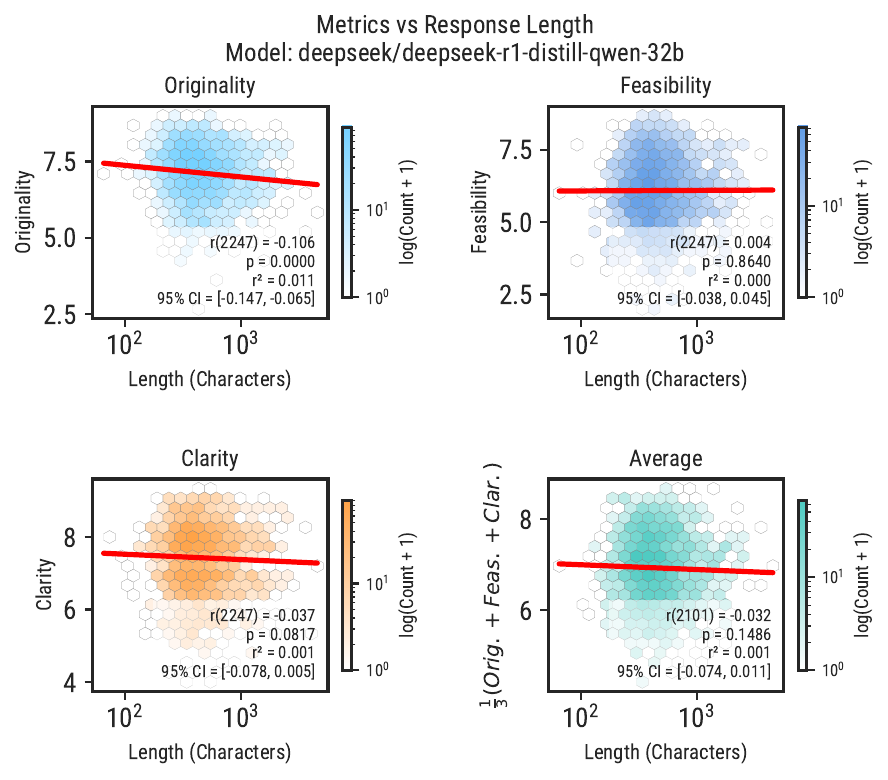}
    \caption{\textbf{Relationship between Full Response Length and Idea Quality for \texttt{deepseek-r1-distill-qwen-32b}}. Hexbin plot showing the density distribution of generated ideas based on the full response length (characters, x-axis) versus the idea quality score (y-axis). The color intensity represents the number of ideas in each bin.}
    \label{fig:hexbin_deepseek_distill_qwen}
\end{figure}

\begin{figure}[h!]
    \centering
    \includegraphics[width=0.6\textwidth]{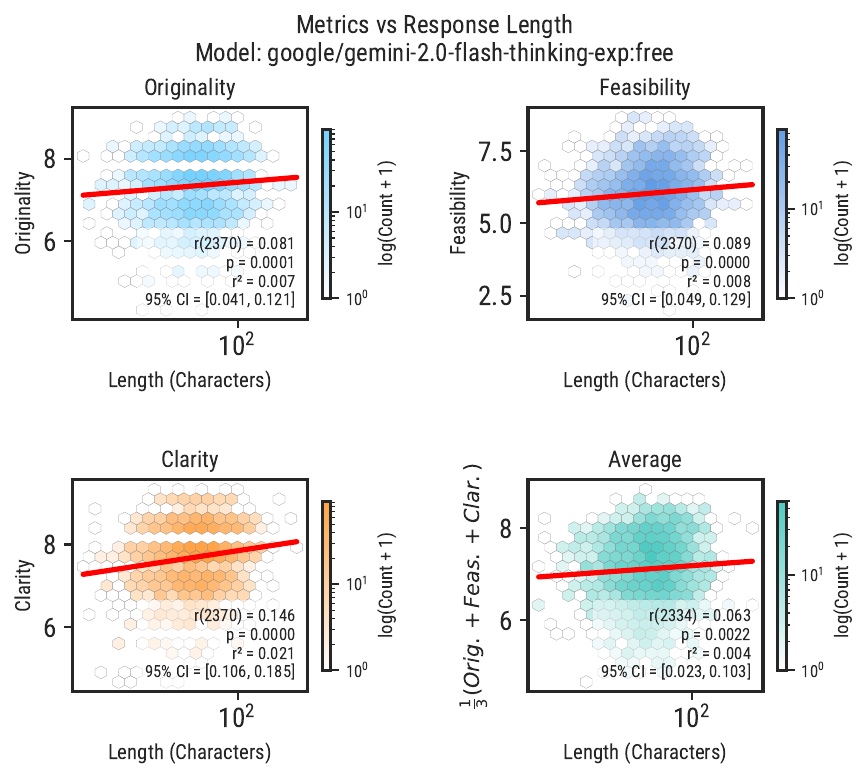}
    \caption{\textbf{Relationship between Full Response Length and Idea Quality for \texttt{gemini-2.0-flash-thinking-exp}}. Hexbin plot showing the density distribution of generated ideas based on the full response length (characters, x-axis) versus the idea quality score (y-axis). The color intensity represents the number of ideas in each bin.}
    \label{fig:hexbin_gemini_flash_thinking}
\end{figure}

\begin{figure}[h!]
    \centering
    \includegraphics[width=0.6\textwidth]{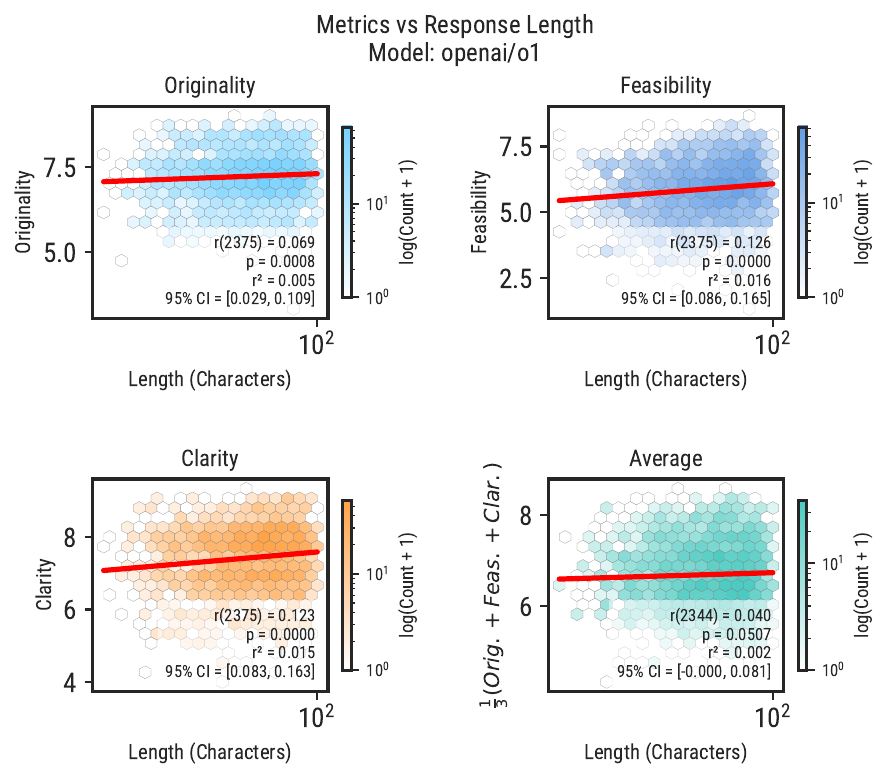}
    \caption{\textbf{Relationship between Full Response Length and Idea Quality for \texttt{o1}}. Hexbin plot showing the density distribution of generated ideas based on the full response length (characters, x-axis) versus the idea quality score (y-axis). The color intensity represents the number of ideas in each bin.}
    \label{fig:hexbin_o1}
\end{figure}

\begin{figure}[h!]
    \centering
    \includegraphics[width=0.6\textwidth]{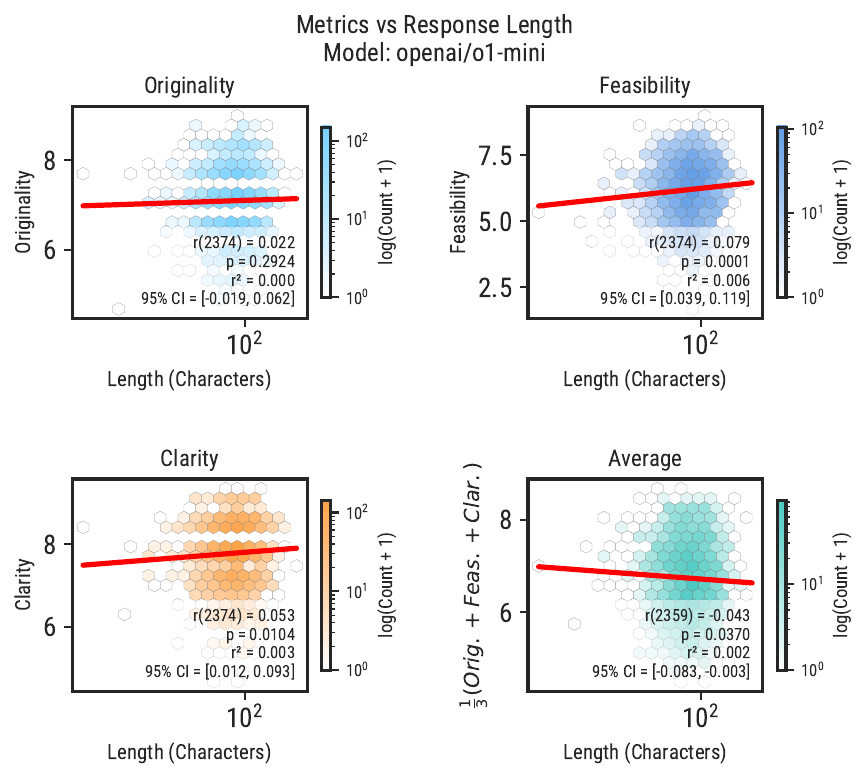}
    \caption{\textbf{Relationship between Full Response Length and Idea Quality for \texttt{o1-mini}}. Hexbin plot showing the density distribution of generated ideas based on the full response length (characters, x-axis) versus the idea quality score (y-axis). The color intensity represents the number of ideas in each bin.}
    \label{fig:hexbin_o1_mini}
\end{figure}

\begin{figure}[h!]
    \centering
    \includegraphics[width=0.6\textwidth]{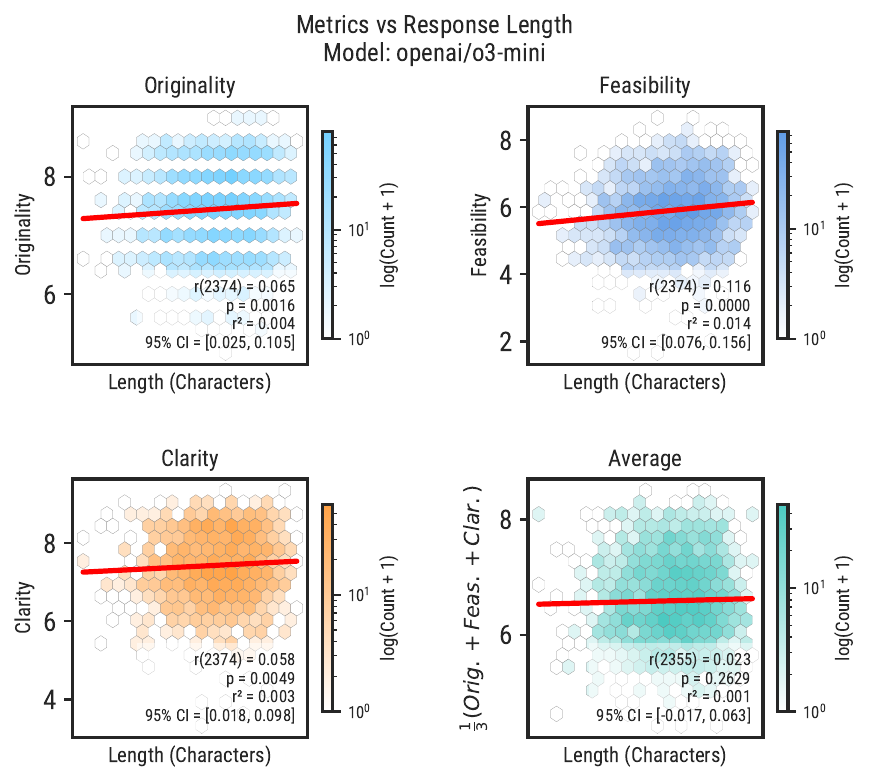}
    \caption{\textbf{Relationship between Full Response Length and Idea Quality for \texttt{o3-mini}}. Hexbin plot showing the density distribution of generated ideas based on the full response length (characters, x-axis) versus the idea quality score (y-axis). The color intensity represents the number of ideas in each bin.}
    \label{fig:hexbin_o3_mini}
\end{figure}

\begin{figure}[h!]
    \centering
    \includegraphics[width=0.6\textwidth]{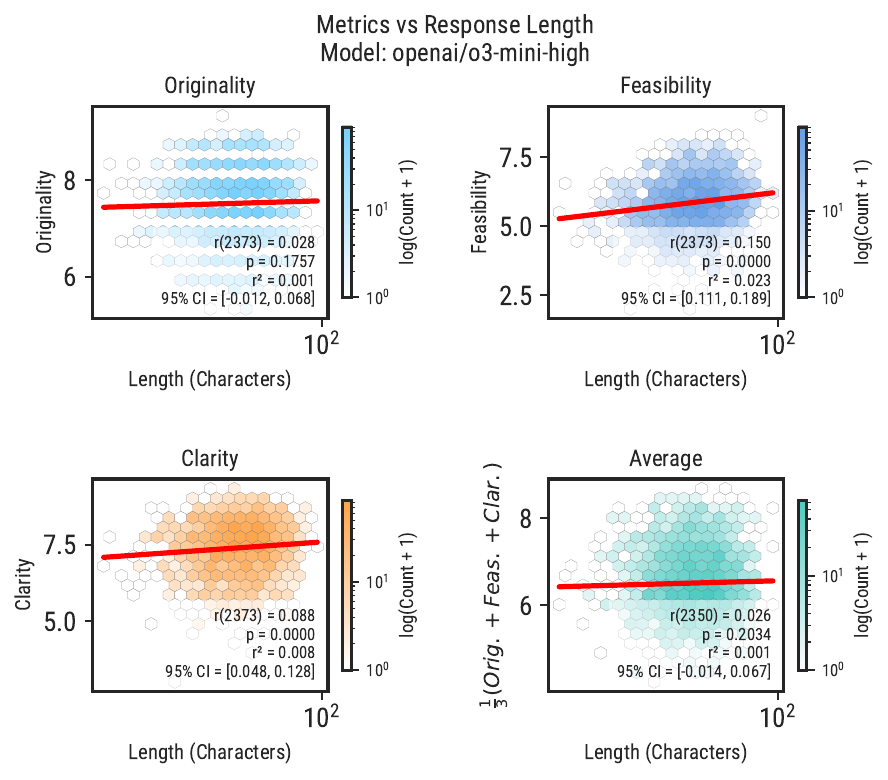}
    \caption{\textbf{Relationship between Full Response Length and Idea Quality for \texttt{o3-mini-high}}. Hexbin plot showing the density distribution of generated ideas based on the full response length (characters, x-axis) versus the idea quality score (y-axis). The color intensity represents the number of ideas in each bin.}
    \label{fig:hexbin_o3_mini_high}
\end{figure}

\begin{figure}[h!]
    \centering
    \includegraphics[width=0.6\textwidth]{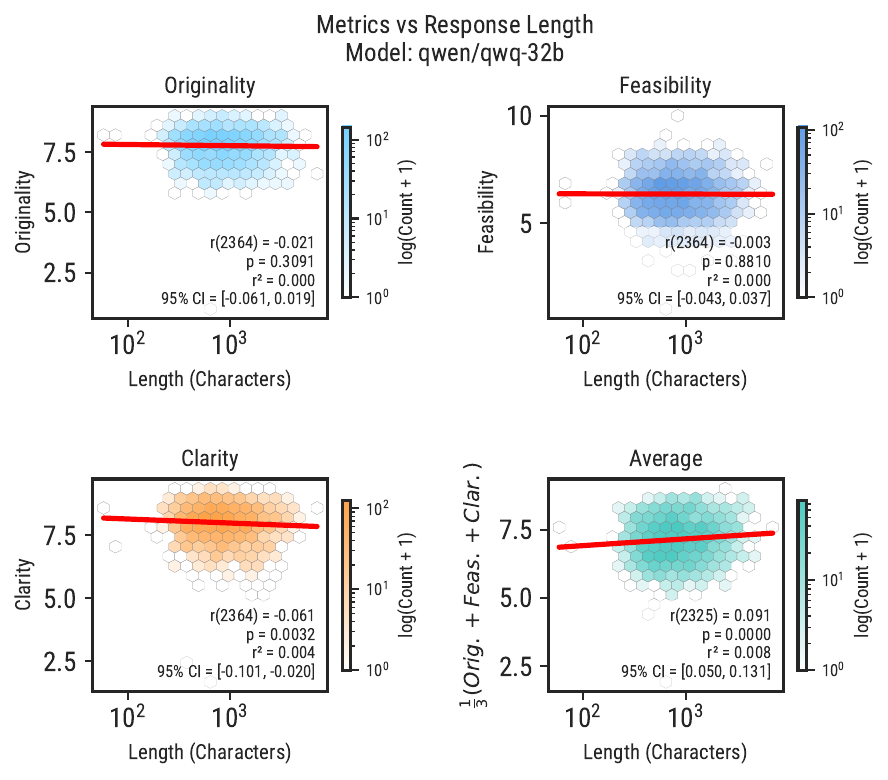}
    \caption{\textbf{Relationship between Full Response Length and Idea Quality for \texttt{qwq-32b}}. Hexbin plot showing the density distribution of generated ideas based on the full response length (characters, x-axis) versus the idea quality score (y-axis). The color intensity represents the number of ideas in each bin.}
    \label{fig:hexbin_qwq_32b}
\end{figure}

\begin{figure}[h!]
    \centering
    \includegraphics[width=0.6\textwidth]{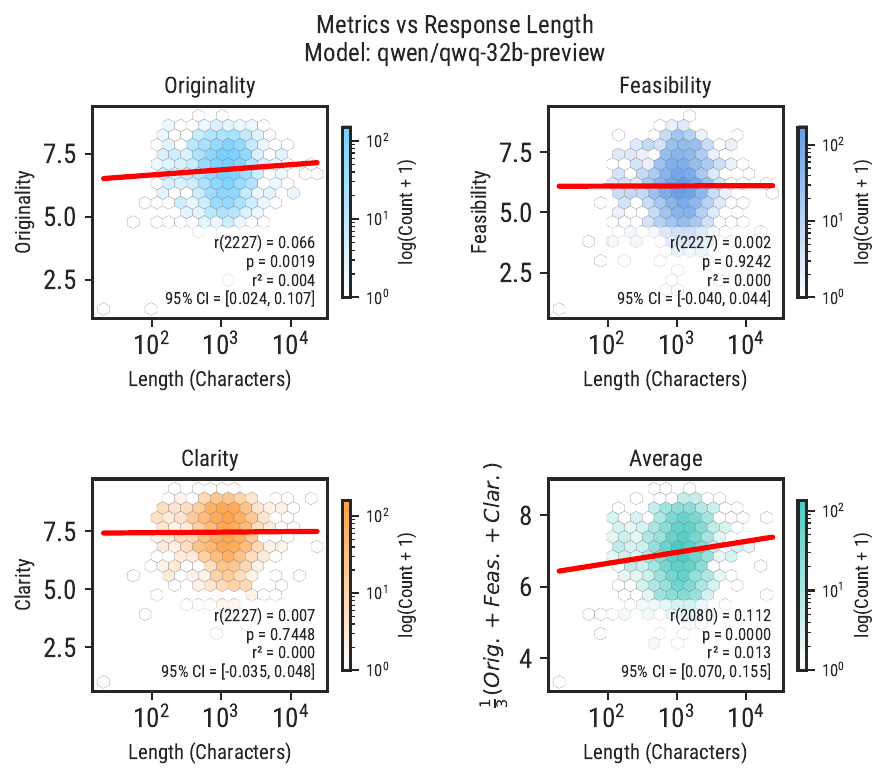}
    \caption{\textbf{Relationship between Full Response Length and Idea Quality for \texttt{qwq-32b-preview}}. Hexbin plot showing the density distribution of generated ideas based on the full response length (characters, x-axis) versus the idea quality score (y-axis). The color intensity represents the number of ideas in each bin.}
    \label{fig:hexbin_qwq_32b_preview}
\end{figure}

As the Figures \ref{fig:hexbin_claude_thinking}--\ref{fig:hexbin_qwq_32b_preview} show, even for these models, there is no strong correlation between the length of the full reasoning output and the quality of the final idea. This finding reinforces the conclusion that idea quality is not merely a function of reasoning length or efforts, distinguishing idea generation capability from some aspects of general reasoning performance.


\clearpage

\section{Statistical Correlation Metrics}
\label{sec:statistical_metrics}

In this study, we employ two standard statistical measures to evaluate the relationships between different evaluation metrics: the Pearson correlation coefficient \((r)\) and Spearman's rank correlation coefficient \((\rho)\). The choice of metric depends on the distribution of the underlying data, which was assessed using the Shapiro-Wilk test for normality.

\paragraph{Pearson Correlation Coefficient \((r)\)}
Pearson's \(r\) is a parametric measure of the linear correlation between two sets of continuous data. It is utilized when the data distributions do not significantly deviate from normality (for instance, when evaluating the correlation between general intelligence scores and scientific ideation capabilities). For two variables \(X\) and \(Y\) with \(n\) paired observations \((x_i, y_i)\), the coefficient is calculated as the covariance of the two variables divided by the product of their standard deviations:

\[
r = \frac{\sum_{i=1}^{n} (x_i - \bar{x})(y_i - \bar{y})}{\sqrt{\sum_{i=1}^{n} (x_i - \bar{x})^2} \sqrt{\sum_{i=1}^{n} (y_i - \bar{y})^2}}
\]
\vspace{6pt}

\noindent where \(\bar{x}\) and \(\bar{y}\) represent the sample means of \(X\) and \(Y\), respectively. The value of \(r\) ranges from \(-1\) to \(1\), where \(1\) indicates a perfect positive linear correlation, \(0\) indicates no linear correlation, and \(-1\) indicates a perfect negative linear correlation.

\paragraph{Spearman's Rank Correlation Coefficient \((\rho)\)}
Spearman's \(\rho\) is a non-parametric measure of rank correlation. It assesses how well the relationship between two variables can be described using a monotonic function, making it robust to outliers and suitable for non-normally distributed data. 

To calculate \(\rho\), the raw scores \(X_i\) and \(Y_i\) are first converted to ranks \(\text{rg}(X_i)\) and \(\text{rg}(Y_i)\). The coefficient is then computed as the Pearson correlation between these rank variables. In cases where all \(n\) ranks are distinct integers (no tied ranks), the formula simplifies to:
\[
\rho = 1 - \frac{6 \sum_{i=1}^{n} d_i^2}{n(n^2 - 1)}
\]
where \(d_i = \text{rg}(X_i) - \text{rg}(Y_i)\) is the difference between the two ranks of each observation, and \(n\) is the total number of observations. Similar to Pearson's \(r\), Spearman's \(\rho\) yields a value between \(-1\) and \(1\).


\begin{thebibliography}{10}
\expandafter\ifx\csname url\endcsname\relax
  \def\url#1{\texttt{#1}}\fi
\expandafter\ifx\csname urlprefix\endcsname\relax\def\urlprefix{URL }\fi
\providecommand{\bibinfo}[2]{#2}
\providecommand{\eprint}[2][]{\url{#2}}

\bibitem{reviewer5_Nature_wang2023scientific}
\bibinfo{author}{Wang, H.} \emph{et~al.}
\newblock \bibinfo{title}{Scientific discovery in the age of artificial intelligence}.
\newblock \emph{\bibinfo{journal}{Nature}} \textbf{\bibinfo{volume}{620}}, \bibinfo{pages}{47--60} (\bibinfo{year}{2023}).

\bibitem{reviewer5_NC_shi2023surprising}
\bibinfo{author}{Shi, F.} \& \bibinfo{author}{Evans, J.}
\newblock \bibinfo{title}{Surprising combinations of research contents and contexts are related to impact and emerge with scientific outsiders from distant disciplines}.
\newblock \emph{\bibinfo{journal}{Nature Communications}} \textbf{\bibinfo{volume}{14}}, \bibinfo{pages}{1641} (\bibinfo{year}{2023}).

\bibitem{reviewer5_arxiv_ai4science2023impact}
\bibinfo{author}{AI4Science, M.~R.} \& \bibinfo{author}{Quantum, M.~A.}
\newblock \bibinfo{title}{The impact of large language models on scientific discovery: a preliminary study using {GPT-4}}.
\newblock \emph{\bibinfo{journal}{arXiv preprint arXiv:2311.07361}}  (\bibinfo{year}{2023}).

\bibitem{reviewer5_NMI_krenn2023forecasting}
\bibinfo{author}{Krenn, M.} \emph{et~al.}
\newblock \bibinfo{title}{Forecasting the future of artificial intelligence with machine learning-based link prediction in an exponentially growing knowledge network}.
\newblock \emph{\bibinfo{journal}{Nature Machine Intelligence}} \textbf{\bibinfo{volume}{5}}, \bibinfo{pages}{1326--1335} (\bibinfo{year}{2023}).

\bibitem{reviewer3_reviewer5_NHB_rafner2023creativity_Rafner_Janet_Beaty_Roger}
\bibinfo{author}{Rafner, J.}, \bibinfo{author}{Beaty, R.~E.}, \bibinfo{author}{Kaufman, J.~C.}, \bibinfo{author}{Lubart, T.} \& \bibinfo{author}{Sherson, J.}
\newblock \bibinfo{title}{Creativity in the age of generative {AI}}.
\newblock \emph{\bibinfo{journal}{Nature Human Behaviour}} \textbf{\bibinfo{volume}{7}}, \bibinfo{pages}{1836--1838} (\bibinfo{year}{2023}).

\bibitem{AlphaGeometry}
\bibinfo{author}{Trinh, T.~H.}, \bibinfo{author}{Wu, Y.}, \bibinfo{author}{Le, Q.~V.}, \bibinfo{author}{He, H.} \& \bibinfo{author}{Luong, T.}
\newblock \bibinfo{title}{Solving olympiad geometry without human demonstrations}.
\newblock \emph{\bibinfo{journal}{Nature}} \textbf{\bibinfo{volume}{625}}, \bibinfo{pages}{476--482} (\bibinfo{year}{2024}).

\bibitem{getzels1962creativity}
\bibinfo{author}{Getzels, J.~W.} \& \bibinfo{author}{Jackson, P.~W.}
\newblock \emph{\bibinfo{title}{Creativity and Intelligence: Exploration with Gifted Students}} (\bibinfo{publisher}{John Wiley \& Sons}, \bibinfo{address}{New York}, \bibinfo{year}{1962}).

\bibitem{guilford1967nature}
\bibinfo{author}{Guilford, J.~P.}
\newblock \emph{\bibinfo{title}{The nature of human intelligence}} (\bibinfo{publisher}{McGraw-Hill}, \bibinfo{year}{1967}).

\bibitem{JAUK2013212_Mathias_Benedek}
\bibinfo{author}{Jauk, E.}, \bibinfo{author}{Benedek, M.}, \bibinfo{author}{Dunst, B.} \& \bibinfo{author}{Neubauer, A.~C.}
\newblock \bibinfo{title}{The relationship between intelligence and creativity: New support for the threshold hypothesis by means of empirical breakpoint detection}.
\newblock \emph{\bibinfo{journal}{Intelligence}} \textbf{\bibinfo{volume}{41}}, \bibinfo{pages}{212--221} (\bibinfo{year}{2013}).
\newblock \urlprefix\url{https://www.sciencedirect.com/science/article/pii/S016028961300024X}.

\bibitem{reviewer5_JSGE_kim2005can}
\bibinfo{author}{Kim, K.~H.}
\newblock \bibinfo{title}{Can only intelligent people be creative? {A} meta-analysis}.
\newblock \emph{\bibinfo{journal}{Journal of Secondary Gifted Education}} \textbf{\bibinfo{volume}{16}}, \bibinfo{pages}{57--66} (\bibinfo{year}{2005}).

\bibitem{reviewer5_PID_preckel2006relationship}
\bibinfo{author}{Preckel, F.}, \bibinfo{author}{Holling, H.} \& \bibinfo{author}{Wiese, M.}
\newblock \bibinfo{title}{Relationship of intelligence and creativity in gifted and non-gifted students: An investigation of threshold theory}.
\newblock \emph{\bibinfo{journal}{Personality and individual differences}} \textbf{\bibinfo{volume}{40}}, \bibinfo{pages}{159--170} (\bibinfo{year}{2006}).

\bibitem{reviewer5_NER_gralewski2012intelligence}
\bibinfo{author}{Gralewski, J.}, \bibinfo{author}{Weremczuk, E.} \& \bibinfo{author}{Karwowski, M.}
\newblock \bibinfo{title}{Intelligence and creativity of polish middle-school students: Looking for the threshold hypothesis}.
\newblock \emph{\bibinfo{journal}{The New Educational Review}} \textbf{\bibinfo{volume}{29}}, \bibinfo{pages}{328--338} (\bibinfo{year}{2012}).

\bibitem{reviewer5_PS_park2008ability}
\bibinfo{author}{Park, G.}, \bibinfo{author}{Lubinski, D.} \& \bibinfo{author}{Benbow, C.~P.}
\newblock \bibinfo{title}{Ability differences among people who have commensurate degrees matter for scientific creativity}.
\newblock \emph{\bibinfo{journal}{Psychological science}} \textbf{\bibinfo{volume}{19}}, \bibinfo{pages}{957--961} (\bibinfo{year}{2008}).

\bibitem{reviewer5_CCAQ_runco1986threshold}
\bibinfo{author}{Runco, M.~A.} \& \bibinfo{author}{Albert, R.~S.}
\newblock \bibinfo{title}{The threshold theory regarding creativity and intelligence: An empirical test with gifted and nongifted children}.
\newblock \emph{\bibinfo{journal}{Creative Child and Adult Quarterly}} \textbf{\bibinfo{volume}{11}}, \bibinfo{pages}{212--218} (\bibinfo{year}{1986}).

\bibitem{reviewer5_LID_benedek2024relationship_Mathias_Benedek}
\bibinfo{author}{Benedek, M.}
\newblock \bibinfo{title}{On the relationship between creative potential and creative achievement: Challenges and future directions}.
\newblock \emph{\bibinfo{journal}{Learning and Individual Differences}} \textbf{\bibinfo{volume}{110}}, \bibinfo{pages}{102424} (\bibinfo{year}{2024}).

\bibitem{reviewer5_JI_weiss2020reappraisal}
\bibinfo{author}{Weiss, S.}, \bibinfo{author}{Steger, D.}, \bibinfo{author}{Schroeders, U.} \& \bibinfo{author}{Wilhelm, O.}
\newblock \bibinfo{title}{A reappraisal of the threshold hypothesis of creativity and intelligence}.
\newblock \emph{\bibinfo{journal}{Journal of Intelligence}} \textbf{\bibinfo{volume}{8}}, \bibinfo{pages}{38} (\bibinfo{year}{2020}).

\bibitem{guilford1950creativity}
\bibinfo{author}{Guilford, J.~P.}
\newblock \bibinfo{title}{Creativity}.
\newblock \emph{\bibinfo{journal}{{American Psychologist}}} \textbf{\bibinfo{volume}{5}}, \bibinfo{pages}{444--454} (\bibinfo{year}{1950}).

\bibitem{reviewer2_boden2004creative}
\bibinfo{author}{Boden, M.~A.}
\newblock \emph{\bibinfo{title}{The creative mind: Myths and mechanisms}} (\bibinfo{publisher}{Routledge}, \bibinfo{year}{2004}).

\bibitem{reviewer2_amabile1983social}
\bibinfo{author}{Amabile, T.~M.}
\newblock \bibinfo{title}{The social psychology of creativity: A componential conceptualization}.
\newblock \emph{\bibinfo{journal}{Journal of Personality and Social Psychology}}, \bibinfo{volume}{45}(\bibinfo{number}{2}), \bibinfo{pages}{357} (\bibinfo{year}{1983}).

\bibitem{reviewer2_rhodes1961analysis}
\bibinfo{author}{Rhodes, M.}
\newblock \bibinfo{title}{An analysis of creativity}.
\newblock \emph{\bibinfo{journal}{The Phi Delta Kappan}} \textbf{\bibinfo{volume}{42}}, \bibinfo{pages}{305--310} (\bibinfo{year}{1961}).

\bibitem{reviewer3_COBS_CORTES201990_Adam_Green}
\bibinfo{author}{Cortes, R.~A.}, \bibinfo{author}{Weinberger, A.~B.}, \bibinfo{author}{Daker, R.~J.} \& \bibinfo{author}{Green, A.~E.}
\newblock \bibinfo{title}{Re-examining prominent measures of divergent and convergent creativity}.
\newblock \emph{\bibinfo{journal}{Current Opinion in Behavioral Sciences}} \textbf{\bibinfo{volume}{27}}, \bibinfo{pages}{90--93} (\bibinfo{year}{2019}).
\newblock \urlprefix\url{https://www.sciencedirect.com/science/article/pii/S2352154618301694}.
\newblock \bibinfo{note}{Creativity}.

\bibitem{rafner2023towards}
\bibinfo{author}{Rafner, J.} \emph{et~al.}
\newblock \bibinfo{title}{Towards game-based assessment of creative thinking}.
\newblock \emph{\bibinfo{journal}{Creativity Research Journal}}, \bibinfo{volume}{35}(\bibinfo{number}{4}), \bibinfo{pages}{763--782} (\bibinfo{year}{2023}).

\bibitem{wei2022chain}
\bibinfo{author}{Wei, J.} \emph{et~al.}
\newblock \bibinfo{title}{Chain-of-thought prompting elicits reasoning in large language models}.
\newblock \emph{\bibinfo{journal}{{Advances in Neural Information Processing Systems}}} \textbf{\bibinfo{volume}{35}}, \bibinfo{pages}{24824--24837} (\bibinfo{year}{2022}).

\bibitem{10.5555/3600270.3601883}
\bibinfo{author}{Kojima, T.}, \bibinfo{author}{Gu, S.~S.}, \bibinfo{author}{Reid, M.}, \bibinfo{author}{Matsuo, Y.} \& \bibinfo{author}{Iwasawa, Y.}
\newblock \bibinfo{title}{Large language models are zero-shot reasoners}.
\newblock In \emph{\bibinfo{booktitle}{Proceedings of the 36th International Conference on Neural Information Processing Systems}} (\bibinfo{address}{Red Hook, NY, USA}, \bibinfo{year}{2024}).

\bibitem{10.5555/3666122.3666639}
\bibinfo{author}{Yao, S.} \emph{et~al.}
\newblock \bibinfo{title}{Tree of thoughts: deliberate problem solving with large language models}.
\newblock In \emph{\bibinfo{booktitle}{Proceedings of the 37th International Conference on Neural Information Processing Systems}} (\bibinfo{address}{Red Hook, NY, USA}, \bibinfo{year}{2024}).

\bibitem{li2023think}
\bibinfo{author}{Li, X.-Y.}, \bibinfo{author}{Xue, J.-T.}, \bibinfo{author}{Xie, Z.} \& \bibinfo{author}{Li, M.}
\newblock \bibinfo{title}{Think outside the code: Brainstorming boosts large language models in code generation}.
\newblock \emph{\bibinfo{journal}{arXiv preprint arXiv:2305.10679}}  (\bibinfo{year}{2023}).

\bibitem{lu2024aiscientistfullyautomated}
\bibinfo{author}{Lu, C.} \emph{et~al.}
\newblock \bibinfo{title}{The {AI Scientist}: Towards fully automated open-ended scientific discovery}.
\newblock \emph{\bibinfo{journal}{arXiv preprint arXiv:2408.06292}}  (\bibinfo{year}{2024}).

\bibitem{hu2024novaiterativeplanningsearch}
\bibinfo{author}{Hu, X.} \emph{et~al.}
\newblock \bibinfo{title}{Nova: An iterative planning and search approach to enhance novelty and diversity of {LLM} generated ideas}.
\newblock \emph{\bibinfo{journal}{arXiv preprint arXiv:2410.14255}}  (\bibinfo{year}{2024}).

\bibitem{baek2024researchagentiterativeresearchidea}
\bibinfo{author}{Baek, J.}, \bibinfo{author}{Jauhar, S.~K.}, \bibinfo{author}{Cucerzan, S.} \& \bibinfo{author}{Hwang, S.~J.}
\newblock \bibinfo{title}{{ResearchAgent}: Iterative research idea generation over scientific literature with large language models}.
\newblock \emph{\bibinfo{journal}{arXiv preprint arXiv:2404.07738}}  (\bibinfo{year}{2024}).

\bibitem{radensky2024scideatorhumanllmscientificidea}
\bibinfo{author}{Radensky, M.} \emph{et~al.}
\newblock \bibinfo{title}{Scideator: {Human-LLM} scientific idea generation grounded in research-paper facet recombination}.
\newblock \emph{\bibinfo{journal}{arXiv preprint arXiv:2409.14634}}  (\bibinfo{year}{2024}).

\bibitem{gottweis2025aicoscientist}
\bibinfo{author}{Gottweis, J.} \emph{et~al.}
\newblock \bibinfo{title}{Towards an {AI} co-scientist}.
\newblock \emph{\bibinfo{journal}{arXiv preprint arXiv:2502.18864}}  (\bibinfo{year}{2025}).
\newblock \urlprefix\url{https://arxiv.org/abs/2502.18864}.
\newblock \eprint{2502.18864}.

\bibitem{reviewer5_ACL_wang2024scimon}
\bibinfo{author}{Wang, Q.}, \bibinfo{author}{Downey, D.}, \bibinfo{author}{Ji, H.} \& \bibinfo{author}{Hope, T.}
\newblock \bibinfo{title}{Scimon: Scientific inspiration machines optimized for novelty}.
\newblock In \emph{\bibinfo{booktitle}{Proceedings of the 62nd Annual Meeting of the Association for Computational Linguistics (Volume 1: Long Papers)}}, \bibinfo{pages}{279--299} (\bibinfo{year}{2024}).

\bibitem{reviewer5_arxiv_gu2024interesting}
\bibinfo{author}{Gu, X.} \& \bibinfo{author}{Krenn, M.}
\newblock \bibinfo{title}{Interesting scientific idea generation using knowledge graphs and {LLMs}: Evaluations with 100 research group leaders}.
\newblock \emph{\bibinfo{journal}{arXiv preprint arXiv:2405.17044}}  (\bibinfo{year}{2024}).

\bibitem{reviewer5_arxiv_gu2024forecasting}
\bibinfo{author}{Gu, X.} \& \bibinfo{author}{Krenn, M.}
\newblock \bibinfo{title}{Forecasting high-impact research topics via machine learning on evolving knowledge graphs}.
\newblock \emph{\bibinfo{journal}{arXiv preprint arXiv:2402.08640}}  (\bibinfo{year}{2024}).

\bibitem{wang2024scipipllmbasedscientificpaper}
\bibinfo{author}{Wang, W.} \emph{et~al.}
\newblock \bibinfo{title}{{SciPIP}: An {LLM}-based scientific paper idea proposer}.
\newblock \emph{\bibinfo{journal}{arXiv preprint arXiv:2410.23166}}  (\bibinfo{year}{2024}).

\bibitem{pu2024ideasynthiterativeresearchidea}
\bibinfo{author}{Pu, K.} \emph{et~al.}
\newblock \bibinfo{title}{{IdeaSynth}: Iterative research idea development through evolving and composing idea facets with literature-grounded feedback}.
\newblock \emph{\bibinfo{journal}{arXiv preprint arXiv:2410.04025}}  (\bibinfo{year}{2024}).

\bibitem{meincke2024using}
\bibinfo{author}{Meincke, L.}, \bibinfo{author}{Girotra, K.}, \bibinfo{author}{Nave, G.}, \bibinfo{author}{Terwiesch, C.} \& \bibinfo{author}{Ulrich, K.~T.}
\newblock \bibinfo{title}{Using large language models for idea generation in innovation}.
\newblock \emph{\bibinfo{journal}{The Wharton School Research Paper Forthcoming}}  (\bibinfo{year}{2024}).
\newblock \urlprefix\url{https://ssrn.com/abstract=4526071}.
\newblock \eprint{4526071}.

\bibitem{si2024llmsgeneratenovelresearch}
\bibinfo{author}{Si, C.}, \bibinfo{author}{Yang, D.} \& \bibinfo{author}{Hashimoto, T.}
\newblock \bibinfo{title}{Can {LLMs} generate novel research ideas? a large-scale human study with 100+ {NLP} researchers}.
\newblock \emph{\bibinfo{journal}{arXiv preprint arXiv:2409.04109}}  (\bibinfo{year}{2024}).
\newblock \urlprefix\url{https://arxiv.org/abs/2409.04109}.
\newblock \eprint{2409.04109}.

\bibitem{reviewer5_NHB_lee2024empirical}
\bibinfo{author}{Lee, B.~C.} \& \bibinfo{author}{Chung, J.}
\newblock \bibinfo{title}{An empirical investigation of the impact of {ChatGPT} on creativity}.
\newblock \emph{\bibinfo{journal}{Nature Human Behaviour}} \textbf{\bibinfo{volume}{8}}, \bibinfo{pages}{1906--1914} (\bibinfo{year}{2024}).

\bibitem{reviewer3_LL_cropley2023artificial}
\bibinfo{author}{Cropley, D.}
\newblock \bibinfo{title}{Is artificial intelligence more creative than humans?: {ChatGPT} and the divergent association task}.
\newblock \emph{\bibinfo{journal}{Learning Letters}} \textbf{\bibinfo{volume}{2}}, \bibinfo{pages}{13--13} (\bibinfo{year}{2023}).

\bibitem{reviewer3_CRJ_marrone2024does}
\bibinfo{author}{Marrone, R.}, \bibinfo{author}{Cropley, D.} \& \bibinfo{author}{Medeiros, K.}
\newblock \bibinfo{title}{How does narrow {AI} impact human creativity?}
\newblock \emph{\bibinfo{journal}{Creativity Research Journal}} \bibinfo{pages}{1--11} (\bibinfo{year}{2024}).

\bibitem{reviewer3_arxiv_wenger2025we_Yoed_Kenett}
\bibinfo{author}{Wenger, E.} \& \bibinfo{author}{Kenett, Y.}
\newblock \bibinfo{title}{We're different, we're the same: Creative homogeneity across {LLMs}}.
\newblock \emph{\bibinfo{journal}{arXiv preprint arXiv:2501.19361}}  (\bibinfo{year}{2025}).

\bibitem{ScienceAdvances_Doshi2023GenerativeAE}
\bibinfo{author}{Doshi, A.~R.} \& \bibinfo{author}{Hauser, O.~P.}
\newblock \bibinfo{title}{Generative {AI} enhances individual creativity but reduces the collective diversity of novel content}.
\newblock \emph{\bibinfo{journal}{Science Advances}} \textbf{\bibinfo{volume}{10}} (\bibinfo{year}{2023}).
\newblock \urlprefix\url{https://api.semanticscholar.org/CorpusID:271119565}.

\bibitem{lu2024benchmarkinglanguagemodelcreativity}
\bibinfo{author}{Lu, Y.}, \bibinfo{author}{Wang, D.}, \bibinfo{author}{Li, T.}, \bibinfo{author}{Jiang, D.} \& \bibinfo{author}{Khashabi, D.}
\newblock \bibinfo{title}{Benchmarking language model creativity: A case study on code generation}.
\newblock \emph{\bibinfo{journal}{arXiv preprint arXiv:2407.09007}}  (\bibinfo{year}{2024}).

\bibitem{zhao2024assessing}
\bibinfo{author}{Zhao, Y.} \emph{et~al.}
\newblock \bibinfo{title}{Assessing and understanding creativity in large language models}.
\newblock \emph{\bibinfo{journal}{arXiv preprint arXiv:2401.12491}}  (\bibinfo{year}{2024}).

\bibitem{dubois2024lengthcontrolledalpacaevalsimpleway}
\bibinfo{author}{Dubois, Y.}, \bibinfo{author}{Galambosi, B.}, \bibinfo{author}{Liang, P.} \& \bibinfo{author}{Hashimoto, T.~B.}
\newblock \bibinfo{title}{Length-controlled {AlpacaEval}: A simple way to debias automatic evaluators}.
\newblock \emph{\bibinfo{journal}{arXiv preprint arXiv:2404.04475}}  (\bibinfo{year}{2024}).

\bibitem{MT-Bench}
\bibinfo{author}{Zheng, L.} \emph{et~al.}
\newblock \bibinfo{title}{Judging {LLM-as-a-judge} with {MT-Bench} and {Chatbot Arena}}.
\newblock \emph{\bibinfo{journal}{{Advances in Neural Information Processing Systems}}} \textbf{\bibinfo{volume}{36}}, \bibinfo{pages}{46595--46623} (\bibinfo{year}{2023}).

\bibitem{li2024crowdsourced}
\bibinfo{author}{Li, T.} \emph{et~al.}
\newblock \bibinfo{title}{From crowdsourced data to high-quality benchmarks: {Arena-Hard} and {BenchBuilder} pipeline}.
\newblock \emph{\bibinfo{journal}{arXiv preprint arXiv:2406.11939}}  (\bibinfo{year}{2024}).

\bibitem{reviewer5_ACL_chiang-lee-2023-large}
\bibinfo{author}{Chiang, C.-H.} \& \bibinfo{author}{Lee, H.-y.}
\newblock \bibinfo{title}{Can large language models be an alternative to human evaluations?}
\newblock In \emph{\bibinfo{booktitle}{Proceedings of the 61st Annual Meeting of the Association for Computational Linguistics (Volume 1: Long Papers)}}, \bibinfo{pages}{15607--15631} (\bibinfo{publisher}{Association for Computational Linguistics}, \bibinfo{address}{Toronto, Canada}, \bibinfo{year}{2023}).
\newblock \urlprefix\url{https://aclanthology.org/2023.acl-long.870}.

\bibitem{reviewer5_ACL_chen-etal-2023-exploring-use}
\bibinfo{author}{Chen, Y.}, \bibinfo{author}{Wang, R.}, \bibinfo{author}{Jiang, H.}, \bibinfo{author}{Shi, S.} \& \bibinfo{author}{Xu, R.}
\newblock \bibinfo{title}{Exploring the use of large language models for reference-free text quality evaluation: An empirical study}.
\newblock In \bibinfo{editor}{Park, J.~C.} \emph{et~al.} (eds.) \emph{\bibinfo{booktitle}{Findings of the Association for Computational Linguistics: IJCNLP-AACL 2023 (Findings)}}, \bibinfo{pages}{361--374} (\bibinfo{publisher}{Association for Computational Linguistics}, \bibinfo{address}{Nusa Dua, Bali}, \bibinfo{year}{2023}).
\newblock \urlprefix\url{https://aclanthology.org/2023.findings-ijcnlp.32/}.

\bibitem{reviewer5_LREC_huang-etal-2024-chatgpt}
\bibinfo{author}{Huang, F.}, \bibinfo{author}{Kwak, H.}, \bibinfo{author}{Park, K.} \& \bibinfo{author}{An, J.}
\newblock \bibinfo{title}{{C}hat{GPT} rates natural language explanation quality like humans: But on which scales?}
\newblock In \bibinfo{editor}{Calzolari, N.} \emph{et~al.} (eds.) \emph{\bibinfo{booktitle}{Proceedings of the 2024 Joint International Conference on Computational Linguistics, Language Resources and Evaluation (LREC-COLING 2024)}}, \bibinfo{pages}{3111--3132} (\bibinfo{publisher}{ELRA and ICCL}, \bibinfo{address}{Torino, Italia}, \bibinfo{year}{2024}).
\newblock \urlprefix\url{https://aclanthology.org/2024.lrec-main.277/}.

\bibitem{reviewer5_NAC_gilardi2023chatgpt}
\bibinfo{author}{Gilardi, F.}, \bibinfo{author}{Alizadeh, M.} \& \bibinfo{author}{Kubli, M.}
\newblock \bibinfo{title}{{ChatGPT} outperforms crowd workers for text-annotation tasks}.
\newblock \emph{\bibinfo{journal}{Proceedings of the National Academy of Sciences}} \textbf{\bibinfo{volume}{120}}, \bibinfo{pages}{e2305016120} (\bibinfo{year}{2023}).

\bibitem{badshah2024referenceguidedverdictllmsasjudgesautomatic}
\bibinfo{author}{Badshah, S.} \& \bibinfo{author}{Sajjad, H.}
\newblock \bibinfo{title}{Reference-Guided Verdict: {LLMs-as-Judges} in Automatic Evaluation of Free-Form Text}.
\newblock \emph{\bibinfo{journal}{arXiv preprint arXiv:2408.09235}} (\bibinfo{year}{2024}).

\bibitem{verga2024replacingjudgesjuriesevaluating}
\bibinfo{author}{Verga, P.} \emph{et~al.}
\newblock \bibinfo{title}{Replacing judges with juries: Evaluating {LLM} generations with a panel of diverse models}.
\newblock \emph{\bibinfo{journal}{arXiv preprint arXiv:2404.18796}}  (\bibinfo{year}{2024}).

\bibitem{reviewer5_arxiv_kim2024prometheus}
\bibinfo{author}{Kim, S.} \emph{et~al.}
\newblock \bibinfo{title}{Prometheus 2: An open source language model specialized in evaluating other language models}.
\newblock \emph{\bibinfo{journal}{arXiv preprint arXiv:2405.01535}}  (\bibinfo{year}{2024}).

\bibitem{reviewer3_Bonus_1st_author_rafner2022mapping}
\bibinfo{author}{Rafner, J.} \emph{et~al.}
\newblock \bibinfo{title}{Mapping citizen science through the lens of human-centered {AI}}.
\newblock \emph{\bibinfo{journal}{Human Computation}} \textbf{\bibinfo{volume}{9}}, \bibinfo{pages}{66--95} (\bibinfo{year}{2022}).

\bibitem{reviewer5_Computer_akata2020research}
\bibinfo{author}{Akata, Z.} \emph{et~al.}
\newblock \bibinfo{title}{A research agenda for hybrid intelligence: augmenting human intellect with collaborative, adaptive, responsible, and explainable artificial intelligence}.
\newblock \emph{\bibinfo{journal}{Computer}} \textbf{\bibinfo{volume}{53}}, \bibinfo{pages}{18--28} (\bibinfo{year}{2020}).

\bibitem{reviewer5_BISE_dellermann2019hybrid}
\bibinfo{author}{Dellermann, D.}, \bibinfo{author}{Ebel, P.}, \bibinfo{author}{S{\"o}llner, M.} \& \bibinfo{author}{Leimeister, J.~M.}
\newblock \bibinfo{title}{Hybrid intelligence}.
\newblock \emph{\bibinfo{journal}{Business \& Information Systems Engineering}} \textbf{\bibinfo{volume}{61}}, \bibinfo{pages}{637--643} (\bibinfo{year}{2019}).

\bibitem{livebench}
\bibinfo{author}{White, C.} \emph{et~al.}
\newblock \bibinfo{title}{{LiveBench}: A challenging, contamination-limited {LLM} benchmark}.
\newblock In \emph{\bibinfo{booktitle}{The Thirteenth International Conference on Learning Representations}} (\bibinfo{year}{2025}).
\newblock \urlprefix\url{https://openreview.net/forum?id=sKYHBTAxVa}.

\bibitem{jain2024livecodebench}
\bibinfo{author}{Jain, N.} \emph{et~al.}
\newblock \bibinfo{title}{{LiveCodeBench}: Holistic and contamination free evaluation of large language models for code}.
\newblock \emph{\bibinfo{journal}{arXiv preprint arXiv:2403.07974}}  (\bibinfo{year}{2024}).

\bibitem{wei2024simplesyntheticdatareduces}
\bibinfo{author}{Wei, J.}, \bibinfo{author}{Huang, D.}, \bibinfo{author}{Lu, Y.}, \bibinfo{author}{Zhou, D.} \& \bibinfo{author}{Le, Q.~V.}
\newblock \bibinfo{title}{Simple synthetic data reduces sycophancy in large language models}.
\newblock \emph{\bibinfo{journal}{arXiv preprint arXiv:2308.03958}}  (\bibinfo{year}{2024}).
\newblock \urlprefix\url{https://arxiv.org/abs/2308.03958}.
\newblock \eprint{2308.03958}.

\bibitem{gu2025surveyllmasajudge}
\bibinfo{author}{Gu, J.} \emph{et~al.}
\newblock \bibinfo{title}{A survey on {LLM-as-a-Judge}}.
\newblock \emph{\bibinfo{journal}{arXiv preprint arXiv:2411.15594}}  (\bibinfo{year}{2025}).
\newblock \urlprefix\url{https://arxiv.org/abs/2411.15594}.
\newblock \eprint{2411.15594}.

\bibitem{fanous2025sycevalevaluatingllmsycophancy}
\bibinfo{author}{Fanous, A.} \emph{et~al.}
\newblock \bibinfo{title}{{SycEval}: Evaluating {LLM} sycophancy}.
\newblock \emph{\bibinfo{journal}{arXiv preprint arXiv:2502.08177}}  (\bibinfo{year}{2025}).
\newblock \urlprefix\url{https://arxiv.org/abs/2502.08177}.
\newblock \eprint{2502.08177}.

\bibitem{reviewer3_TIS_huang2025survey}
\bibinfo{author}{Huang, L.} \emph{et~al.}
\newblock \bibinfo{title}{A survey on hallucination in large language models: Principles, taxonomy, challenges, and open questions}.
\newblock \emph{\bibinfo{journal}{ACM Transactions on Information Systems}} \textbf{\bibinfo{volume}{43}}, \bibinfo{pages}{1--55} (\bibinfo{year}{2025}).

\bibitem{final_cite_reviewer3_reviewer5_CHI_falk2025hackathons}
\bibinfo{author}{Falk, J.} \emph{et~al.}
\newblock \bibinfo{title}{How do hackathons foster creativity? towards ai collaborative evaluation of creativity at scale}.
\newblock In \emph{\bibinfo{booktitle}{CHI'25: Proceedings of the 2025 CHI Conference on Human Factors in Computing Systems}} (\bibinfo{organization}{Association for Computing Machinery (ACM)}, \bibinfo{year}{2025}).

\bibitem{final_cite_reviewer5_distefano2025evaluating}
\bibinfo{author}{DiStefano, P.~V.} \emph{et~al.}
\newblock \bibinfo{title}{Evaluating ai’s ideas: The role of individual creativity and expertise in human-ai co-creativity}  (\bibinfo{year}{2025}).

\bibitem{reviewer5_NHB_sourati2023accelerating}
\bibinfo{author}{Sourati, J.} \& \bibinfo{author}{Evans, J.~A.}
\newblock \bibinfo{title}{Accelerating science with human-aware artificial intelligence}.
\newblock \emph{\bibinfo{journal}{Nature Human Behaviour}} \textbf{\bibinfo{volume}{7}}, \bibinfo{pages}{1682--1696} (\bibinfo{year}{2023}).

\bibitem{reviewer5_arxiv_lacoste2019quantifying}
\bibinfo{author}{Lacoste, A.}, \bibinfo{author}{Luccioni, A.}, \bibinfo{author}{Schmidt, V.} \& \bibinfo{author}{Dandres, T.}
\newblock \bibinfo{title}{Quantifying the carbon emissions of machine learning}.
\newblock \emph{\bibinfo{journal}{arXiv preprint arXiv:1910.09700}}  (\bibinfo{year}{2019}).

\bibitem{ecologits-calculator}
\bibinfo{author}{Rince, S.} \& \bibinfo{author}{Banse, A.}
\newblock \bibinfo{title}{{EcoLogits}: Evaluate the environmental impacts of generative {AI}} (\bibinfo{year}{2025}).
\newblock \urlprefix\url{https://doi.org/10.5281/zenodo.15132099}.

\bibitem{anthropic2024claude3}
\bibinfo{author}{{AI Anthropic}}.
\newblock \bibinfo{title}{The {Claude} 3 model family: {Opus}, {Sonnet}, {Haiku}}.
\newblock \bibinfo{howpublished}{Claude-3 Model Card} (\bibinfo{year}{2024}).

\bibitem{achiam2023gpt}
\bibinfo{author}{Achiam, J.} \emph{et~al.}
\newblock \bibinfo{title}{{GPT-4} technical report}.
\newblock \emph{\bibinfo{journal}{arXiv preprint arXiv:2303.08774}}  (\bibinfo{year}{2023}).

\bibitem{team2024gemini}
\bibinfo{author}{Team, G.} \emph{et~al.}
\newblock \bibinfo{title}{Gemini 1.5: Unlocking multimodal understanding across millions of tokens of context}.
\newblock \emph{\bibinfo{journal}{arXiv preprint arXiv:2403.05530}}  (\bibinfo{year}{2024}).

\bibitem{bai2023qwen}
\bibinfo{author}{Bai, J.} \emph{et~al.}
\newblock \bibinfo{title}{Qwen technical report}.
\newblock \emph{\bibinfo{journal}{arXiv preprint arXiv:2309.16609}}  (\bibinfo{year}{2023}).

\bibitem{qwen2025qwen25technicalreport}
\bibinfo{author}{Qwen:} \emph{et~al.}
\newblock \bibinfo{title}{Qwen2.5 technical report}.
\newblock \emph{\bibinfo{journal}{arXiv preprint arXiv:2412.15115}}  (\bibinfo{year}{2025}).
\newblock \urlprefix\url{https://arxiv.org/abs/2412.15115}.
\newblock \eprint{2412.15115}.

\bibitem{liu2024deepseek}
\bibinfo{author}{Liu, A.} \emph{et~al.}
\newblock \bibinfo{title}{Deepseek-v2: A strong, economical, and efficient mixture-of-experts language model}.
\newblock \emph{\bibinfo{journal}{arXiv preprint arXiv:2405.04434}}  (\bibinfo{year}{2024}).

\bibitem{deepseekai2025deepseekv3technicalreport}
\bibinfo{author}{DeepSeek-AI}.
\newblock \bibinfo{title}{{DeepSeek-V3} technical report}.
\newblock \emph{\bibinfo{journal}{arXiv preprint arXiv:2412.19437}}  (\bibinfo{year}{2025}).
\newblock \eprint{2412.19437}.

\bibitem{deepseekai2025deepseekr1incentivizingreasoningcapability}
\bibinfo{author}{DeepSeek-AI}.
\newblock \bibinfo{title}{{DeepSeek-R1}: Incentivizing reasoning capability in {LLMs} via reinforcement learning}.
\newblock \emph{\bibinfo{journal}{arXiv preprint arXiv:2501.12948}}  (\bibinfo{year}{2025}).
\newblock \eprint{2501.12948}.

\bibitem{dubey2024llama}
\bibinfo{author}{Dubey, A.} \emph{et~al.}
\newblock \bibinfo{title}{The {Llama} 3 herd of models}.
\newblock \emph{\bibinfo{journal}{arXiv preprint arXiv:2407.21783}}  (\bibinfo{year}{2024}).

\bibitem{jiang2023mistral}
\bibinfo{author}{Jiang, A.~Q.} \emph{et~al.}
\newblock \bibinfo{title}{Mistral {7B}}.
\newblock \emph{\bibinfo{journal}{arXiv preprint arXiv:2310.06825}}  (\bibinfo{year}{2023}).

\bibitem{intelligence2024amazon}
\bibinfo{author}{Intelligence, A. A.~G.}
\newblock \bibinfo{title}{The {Amazon Nova} family of models: Technical report and model card}  (\bibinfo{year}{2024}).

\bibitem{abdin2024phi4technicalreport}
\bibinfo{author}{Abdin, M.} \emph{et~al.}
\newblock \bibinfo{title}{Phi-4 technical report}.
\newblock \emph{\bibinfo{journal}{arXiv preprint arXiv:2412.08905}}  (\bibinfo{year}{2024}).
\newblock \eprint{2412.08905}.

\bibitem{beltagy-etal-2019-scibert}
\bibinfo{author}{Beltagy, I.}, \bibinfo{author}{Lo, K.} \& \bibinfo{author}{Cohan, A.}
\newblock \bibinfo{title}{{S}ci{BERT}: A pretrained language model for scientific text}.
\newblock In \bibinfo{editor}{Inui, K.}, \bibinfo{editor}{Jiang, J.}, \bibinfo{editor}{Ng, V.} \& \bibinfo{editor}{Wan, X.} (eds.) \emph{\bibinfo{booktitle}{Proceedings of the 2019 Conference on Empirical Methods in Natural Language Processing and the 9th International Joint Conference on Natural Language Processing (EMNLP-IJCNLP)}}, \bibinfo{pages}{3615--3620} (\bibinfo{address}{Hong Kong, China}, \bibinfo{year}{2019}).
\newblock \urlprefix\url{https://aclanthology.org/D19-1371}.

\bibitem{cohen2021boundary}
\bibinfo{author}{Cohen, E.}
\newblock \bibinfo{title}{The boundary lens: {T}heorising academic activity}.
\newblock In \emph{\bibinfo{booktitle}{The University and its Boundaries: Thriving or Surviving in the 21st Century}}, \bibinfo{pages}{14--41} (\bibinfo{publisher}{Routledge}, \bibinfo{address}{New York, New York}, \bibinfo{year}{2021}), \bibinfo{edition}{1} edn.

\end{thebibliography}
\end{document}